\pdfoutput=1
\documentclass{article}
\usepackage{iclr2027_conference,times}

\usepackage{amsmath,amsfonts,bm}

\def\eqref#1{equation~\ref{#1}}

\def\1{\bm{1}}

\DeclareMathAlphabet{\mathsfit}{\encodingdefault}{\sfdefault}{m}{sl}
\SetMathAlphabet{\mathsfit}{bold}{\encodingdefault}{\sfdefault}{bx}{n}

\usepackage{hyperref}
\usepackage{url}

\usepackage{subcaption}
\usepackage{graphicx}
\usepackage{mathtools}
\usepackage{amssymb}
\usepackage{booktabs}
\usepackage{multirow}

\definecolor{myblue}{rgb}{0.141, 0.514, 0.596}

\title{Deep Latents, Long Memory: \\ Latent persistence explains temporal correlations in transformer activations}

\author{%
Francesco Cagnetta\thanks{Correspondence to: \texttt{fr.cagnetta@gmail.com}.}\\
Simplex
\And
Santiago Acevedo\\
SISSA
}

\iclrfinalcopy
\begin{document}

\maketitle
\lhead{Preprint.}

\begin{abstract}
The hidden layers of Large Language Models (LLMs) are believed to encode representations of the data whose abstraction increases with depth. However, in the absence of an operational notion of abstraction, this statement is hard to test quantitatively. Here, we propose an alternative view, building on the principle that the correlations of a sequence reflect the structure of the process that generated it. We define the autocorrelation function of a layer as the average dot product between its activations at two context positions separated by a lag. Latent variables encoded persistently along the context create slowly decaying tails in the autocorrelation, and we find these tails to grow across layers in both synthetic and natural data.
In synthetic data with controlled hierarchical latent structure, this growth stems from the encoding of abstract latent variables, which govern long stretches of the input.
Indeed, the persistence of latent codes predicts the qualitative structure of autocorrelations and accounts for most of the empirical measures.
In real text, we take sparse-autoencoder features as proxies for latent variables, and find that their persistence correlates strongly with the magnitude of autocorrelations across datasets and LLMs. In particular, the persistence time increases with depth, as do autocorrelations, implying that deeper latent features persist for longer context spans. We complement this analysis by explicitly ablating the persistent sparse-autoencoder features and observing a consequent decrease in the autocorrelations.
\end{abstract}
\begin{figure}[h!]
    \centering
    \includegraphics[width=\linewidth]{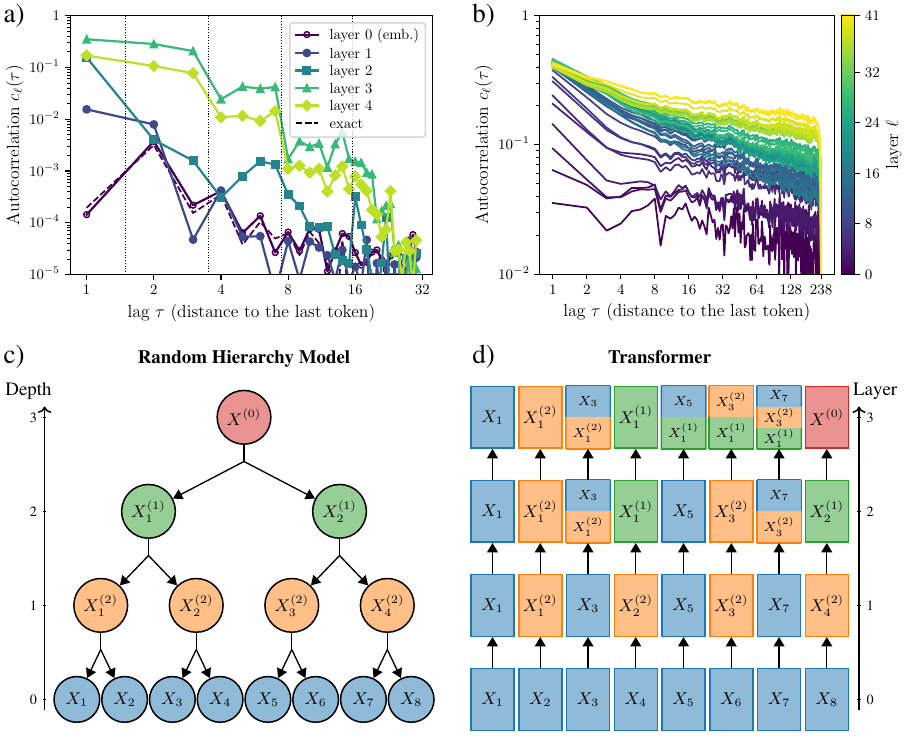}
    \caption{
        \textbf{Neural autocorrelations $c_\ell(\tau)$ (Eq.~\ref{eq:autocorrelations}) develop with depth due to persistently encoded abstract latent features.}
        \textbf{a) 5-block transformer trained on the Random Hierarchy Model (RHM)} and evaluated on RHM sentences (dashed: exact token-level curve; dotted: span ends of the latents governing the next token, cf. panel d).
        \textbf{b) Pre-trained Gemma 2 9B}~\citep{gemmateam2024gemma2} evaluated on The Pile.
        In both a) and b), deep layers show a slower decay with the lag. 
        \textbf{c), d)} illustrate the relationship between long-range autocorrelations and the persistent encoding of abstract latent features.
        \textbf{c)} shows our idealised representation of the latent hierarchical structure of natural language data: pairs of adjacent input tokens (blue) are generated by a single, latent parent variable (orange); pairs of adjacent parent variables are generated by a single latent grandparent (green), and so on.
        Deep latent variables correspond to abstract concepts and influence large portions of the input.
        \textbf{d)} Deep learning methods reconstruct the data structure layer by layer~\citep {cagnetta2024random,cagnetta2024towards}, here illustrated schematically in a deep autoregressive transformer.
        Each row represents the variables encoded by the activations in a given layer, assuming that layer $\ell$ can only compute latent variables of depth $\ell$---see the main text for a detailed breakdown.
        Deeper latent variables emerge in later layers, and persist for longer across context positions: E.g., depth-$1$ (orange) variables persist for two slots in layers $2$ and $3$, whereas depth-$2$ variables (green) persist for $4$ slots, but only in layer $3$.
        This scheme implies a specific pattern of decay of $c_\ell(\tau)$, highlighted with dotted lines in panel a) and approximately respected by the measured autocorrelations.
    }
    \label{fig:main}
\end{figure}

\section{Introduction}

Language is structured across many temporal scales: local constructions constrain nearby tokens, while abstract concepts such as topic and discourse states can remain relevant over much longer spans.
Large Language Models (LLMs) adapt to this structure over training, developing representations of syntactic~\citep{peters2018dissecting, tenney2019bert, manning2020emergent,diego2024polar} and semantic features~\citep{engels2024not, gurnee2025when,acevedo2026differential}.
Understanding such representations, including their content and the mechanism of their formation, remains a central challenge for a theory of LLMs and deep learning at large~\citep{michaud2026purpose}.

On the empirical side, a growing collection of methods probes the content of LLMs' representations, as well as their organisation across layers.
Sparse autoencoder (SAE) features~\citep{huben2024sparse} change or emerge with depth~\citet{balcells2024evolution}, shifting from positional and structural signals in early layers to context-binding, semantic features in intermediate layers, and then to output-oriented signals in the final layers \citep{sun2025dense}.
Features shared across languages are likewise most common in the middle layers~\citep{lindsey2025biology}.
Temporal extensions of SAEs~\citep{lubana2026priors,bhalla2026temporal,luo2026persistent} find that semantic features vary slowly along the context, whereas rapidly changing ones capture lexical and syntactic patterns.
Geometric measures, both local~\citep{hosseini2023straighten,king2026curvature} and global~\citep{cheng2025emergence}, also indicate that representations become more abstract with depth, as the superficial properties captured by the early layers gradually give way to high-level semantic information.
Abstract representations can be identified via similarity measures between points in semantic correspondence, such as image-caption pairs~\citep{platonic} and translations~\citep{acevedo2026semantic}.
However, we still lack a framework that unifies these observations and characterises the layerwise information processing in LLMs without relying on prior assumptions about the represented concepts.

On the theoretical side, training LLMs on synthetic data with controlled latent structure~\citep{cagnetta2024random, shai2024transformers, allenzhu} enables quantitative studies of how representations are acquired and of the sample cost of recovering abstract variables.
In particular, these analyses show how deeper latent structures and longer-range statistical dependencies emerge progressively with data, training, and network depth~\citep{cagnetta2024towards,garnier-brun2025how,zhou2026how,schulz2026unraveling}. However, the quantitative application of these results to real data, where the true latent structure is unknown, remains an open problem.

\subsection{Our contribution}
\label{sec:our-contribution}
Bridging theoretical and empirical approaches requires translating insights obtained from synthetic datasets into observables that can also describe the behaviour of pretrained language models.
Here, \textbf{we propose the autocorrelation function of neural activations across context positions as one such bridge}.
Given a set of LLM activations $\phi_{t}^{\scriptscriptstyle(\ell)}$ (layer $\ell$, token position $t=1,\dots,T$) standardised so that $\mathbb{E}[\phi]\,{=}\,0$, $\mathbb{E}[\| \phi\|^2]\,{=}\,1$, we define\footnote{Eq.~\ref{eq:autocorrelations} uses the last token $T$ as an ``anchor'' or reference token,  but the autocorrelation function can be measured between any two tokens separated at distance $\tau$.} an autocorrelation function of neural activations as
\begin{align}
c_\ell (\tau) 
\coloneqq 
\mathbb{E} \left[\phi_{T-\tau}^{(\ell)} \cdot \phi_{T}^{(\ell)}\right ].
\label{eq:autocorrelations}
\end{align}
Eq.~\ref{eq:autocorrelations} \textbf{generalises token-coincidence probabilities to the continuous inner representations of LLMs}, as described in Sec.~\ref{sec:autocorr-explained}. 
Under the linear representation hypothesis \citep{park2024linear}, a latent variable contributes to autocorrelations when it is encoded at both positions in aligned directions.
We measure Eq.~\ref{eq:autocorrelations} in synthetic data, where scope and identity of the underlying latent variables are known, as well as natural data, where SAE features offer an imperfect but tractable proxy for latent content.

Our main results are summarised graphically in Fig.~\ref{fig:main}:
~\ref{fig:main}a) and~\ref{fig:main}b) show that autocorrelations grow with depth in both synthetic (Random Hierarchy Model, RHM) and natural language (see Appendix~\ref{app:autocorrelations-other-models} for an extension to different models).
We propose this is due to the latent hierarchical structure of language data (\ref{fig:main}c, d): if more abstract latents govern a longer stretch of text, and their code persists over that stretch, we expect them to produce a longer autocorrelation tail.
We support this hypothesis by showing that:
\begin{itemize}
\item \textbf{Persistent encoding explains autocorrelations when the latents are known}, as in the synthetic data represented in Fig.~\ref{fig:main}c, where the construction of Fig.~\ref{fig:main}d explains a significant fraction of the measured correlations (Fig.~\ref{fig:rhm-trace-composition});
\item \textbf{The development of autocorrelations during training follows the learning of the latents:} In synthetic data, each branch of the autocorrelation function emerges over training precisely when the model discovers the corresponding latent variable (Fig.~\ref{fig:rhm-dynamics});
\item \textbf{The relationship between correlations and persistence holds in pretrained LLMs}, where the median and mean persistence length of SAE features correlate strongly with the integrated autocorrelation $\chi_\ell = \sum_\tau c_\ell (\tau)$ (Fig.~\ref{fig:layer-dependences-NLP});
\item \textbf{Persistent latent features carry the autocorrelations}, as removing them from the activations decreases autocorrelations significantly, whereas removing a non-persistent variable does not (Table~\ref{tab:causal-ablation}).
\end{itemize}

\subsection{Further related work}
\label{sec:related-work}

\paragraph{Correlations at the token level.}
Two-point statistics of the tokens identify the underlying generative processes.
In written texts, long-range correlations have been traced to the hierarchical structure of paragraphs and chapters~\citep{alvarezlacalle2006hierarchical}.
Power-law decay is indeed typical of hierarchical structures such as context-free grammars~\citep{lin2017critical}, whereas a Markovian structure would imply exponential decay.
The same statistics govern learning: the decay of pairwise token correlations with the separation, combined with the decay of the next-token conditional entropy with the context length, predicts the exponents of data-limited neural scaling laws~\citep{cagnetta2026deriving}.
All these works characterise input statistics.
Eq.~\ref{eq:autocorrelations} extends the same observable to representations, with the token-level function as its layer-$0$ limit.

\paragraph{Temporal structure of LLM representations.}
Several of the works cited in the introduction also track representations along context and layers, with different statistics.
The curvature of~\citet{hosseini2023straighten,king2026curvature} can be thought of as a local autocorrelation, confined to time lag $1$ and averaged across end positions.
Our measure, instead, captures local as well as long-range structure.
\citet{skrill2023windows} use token perturbations to quantify effective integration windows, or receptive fields, finding a crossover from short-range, exponential behaviour in the early layers to long-range, power-law behaviour in the deeper layers.
\citet{gurnee2026workspace} measure the autocorrelation of the dominant concept read out by a Jacobian lens, and find it largest in the intermediate layers.
This observation supports the emergence of a global workspace, carrying abstract content that persists across the context.
Our approach is complementary, as it focuses on raw activations, independently of how they affect the model's output.

\paragraph{Representations of models trained on data with controlled latent structure.}
Our analysis of synthetic data belongs to a line of work that trains networks on data with known latent variables and asks how these are represented.
\citet{allenzhu} train GPT-2 models on synthetic context-free grammars and show that the hidden states linearly encode the latent variables at the boundaries of their spans.
\citet{shai2024transformers} show that transformers trained on hidden Markov models linearly represent, in the residual stream, the belief state over the hidden variables of the generator, reproducing its geometry within the probability simplex.
In this language, the persistence of a code over the span of a latent variable is a belief about that variable that stays fixed until the span closes.
On hierarchical data, attention maps and linear probes show that latent variables are resolved sequentially over training and depth~\citep{garnier-brun2025how,cagnetta2025scaling,zhou2026how}.
These works establish that latent variables are linearly readable from the activations.
The regression of Eq.~\ref{eq:linear-representation-hypothesis} is the analogue of their probes, which we use to ask how long latent variables remain encoded in the activations. Then, we turn the answer into an observable that transfers to natural language.

\section{Autocorrelations of neural activity}
\label{sec:autocorr-explained}

In this section, we explain how the autocorrelation function of Eq.~\ref{eq:autocorrelations} generalises token-coincidence probabilities.

Let $(X_t)_{t=1\dots T}$ be discrete tokens with vocabulary size $v$, and denote with $e_t \in \mathbb{R}^v$ the one-hot encodings,
\begin{align}
(e_t)_i = \begin{cases}
1\quad\text{if}\quad X_t=i,\\ 
0 \quad\text{otherwise}.
\end{cases} 
\end{align}
The mean one-hot vector over the data distribution coincides with the marginal probability, while the mean scalar product between two one-hot vectors is the coincidence probability,
\begin{align} \mathbb{E}\left[(e_t)_i\right] = P(X_t = i),\quad \mathbb{E}\left[ e_T \cdot e_{T-\tau}\right] = P(X_T = X_{T-\tau}).
\end{align}
Centering by subtracting $\mathbb{E}[e_t]$ removes the coincidences expected from the marginal token frequency, i.e., the coincidence probability of two i.i.d. tokens.
Assuming translational invariance of the marginal probabilities, we get the following (non-standardised) token autocorrelation
\begin{align}\label{eq:token-autocorr-unnorm}
\mathbb{E}\left[ (e_{T-\tau}- \mathbb{E}\left[e_{T-\tau}\right])\cdot(e_T- \mathbb{E}\left[e_T\right])\right] = \mathbb{P}\left( X_{T-\tau}=X_T \right)-\sum_i  \mathbb{P}\left( X_t=i\right)^2.
\end{align}

After the embedding, layer-0 (non-standardised) activations read $\phi_t = W e_t+p_t$, where $W\in \mathbb{R}^{d\times v}$ is the embedding matrix and $p_t\in \mathbb{R}^d$ the positional encoding.
Centering removes $p_t$, thus 
\begin{align}
\mathbb{E}\left[ (\phi_{T-\tau}- \mathbb{E}\left[\phi_{T-\tau}\right])\cdot(\phi_T- \mathbb{E}\left[\phi_T\right])\right] &= \mathbb{E}\left[(e_{T-\tau}- \mathbb{E}\left[e_{T-\tau}\right])^\top W^\top W  (e_T- \mathbb{E}\left[e_T\right])\right].
\end{align}
The only difference from Eq.~\ref{eq:token-autocorr-unnorm} is the matrix $W^\top W$ between the two centered encodings, with entries $(W^\top W)_{ij}\,{=}\,w_i\cdot w_j$ equal to the dot-product between the embedding vectors of tokens $i$ and $j$.
If the embedding vectors were orthogonal with equal norm, $W^\top W$ would be a multiple of the identity.
Standardisation would then cancel this
multiple exactly, so that the autocorrelations of the embeddings and one-hot encodings coincide.
For random embeddings, this coincidence is approximate: random vectors in high dimension are close to orthogonal, with cosines
of order $d^{\scriptscriptstyle-1/2}$.
Appendix~\ref{app:rhm-embeddings} shows that random embeddings indeed leave the autocorrelation function's shape unchanged on the logarithmic scale.

The linear representation hypothesis extends this relationship from tokens to deeper layers:
\begin{align}\label{eq:linear-representation-hypothesis}
    \phi^{(\ell)}_t = A^{(\ell)}_t\, e_\mathbf{Z} + \epsilon^{(\ell)}_t ,
\end{align}
where the one-hot encodings of the tokens are replaced by those of the latent features $\mathbf{Z}$, the embedding vectors by feature directions, i.e. the columns of $A^{\scriptscriptstyle(\ell)}_t$,
and $\epsilon^{\scriptscriptstyle(\ell)}_t$ represents a residual component uncorrelated from the latent features $\mathbf{Z}$.
Where the two positions share feature directions, the autocorrelation counts how often the latent features they encode
coincide beyond chance, as it does for tokens.
Substituting Eq.~\ref{eq:linear-representation-hypothesis} into Eq.~\ref{eq:autocorrelations} splits the autocorrelation into linear codes and residuals,
\begin{align}\label{eq:rhm-trace-split}
    c_\ell(\tau) = c^{\rm code}_\ell(\tau) + c^{\rm res}_\ell(\tau),\quad\begin{cases}
    c^{\rm code}_\ell(\tau) &\propto \mathrm{Tr}\!\left[A^{(\ell)}_{T}\,\mathrm{Cov}(e_\mathbf{Z})\,A^{(\ell)\top}_{T-\tau}\right],\\
    c^{\rm res}_\ell(\tau) &\propto \mathbb{E}\!\left[\epsilon^{(\ell)}_{T-\tau}\cdot \epsilon^{(\ell)}_{T}\right].\end{cases}
\end{align}
Expanding $\mathrm{Cov}(e_\mathbf{Z})$ over pairs of nodes splits $c^{\rm code}_\ell$ further into \emph{same-node} terms, with one variable encoded at both positions along aligned directions, and \emph{cross-node} terms, arising in the presence of correlations between latent variables. Same-node terms represent persistence.

\section{Autocorrelations with controlled hierarchical structure}
\label{sec:rhm}

In this section, we look at a GPT-2-style model trained on synthetic data, and describe how neural autocorrelations across layers reflect the latent hierarchical structure of the data.

\paragraph{Random Hierarchy Model (RHM).}
The RHM is a family of data-generating processes with controlled hierarchical structure, introduced in~\citep{cagnetta2024random} as a testbed for studying deep representation learning.
A sentence of tokens is generated top-down along a binary tree of fixed depth (Fig.~\ref{fig:main}c):
Starting from the root, every symbol is replaced by a pair of symbols of the next level according to fixed rules.
This hierarchy of hidden variables constitutes the latent hierarchical structure of the data.
Since the rules are fixed, the entropy of the distribution of generated sentences is known.
Hence, we can track precisely the progress of any learning model trained on RHM data by comparing the cross-entropy loss to such entropy.
Here, we consider a deep GPT-2-style LLM with enough layers to represent the latent hierarchical structure of the RHM, and given just enough data, parameters and training time to reach the optimal cross-entropy loss (details in Appendix~\ref{app:rhm-setup}).
Specifically, we use vocabulary size $v\,{=}\,32$, $m\,{=}\,8$ rules per symbol, and depth $5$ for both transformer and data-generating process.
All sequences have length $T\,{=}\,2^5$; The model reads a beginning-of-sequence token [BOS] followed by $X_1,\dots,X_{31}$ and predicts $X_1,\dots,X_{32}$.

\paragraph{Content of the trained network.}
We now compare the activations of the trained LLM to the sketch in Fig.~\ref{fig:main}, d.
The ground truth latent variables of every sentence are the hidden nodes of the tree (Fig.~\ref{fig:main}, c).
Hence, we can ask directly if any activation encodes them linearly.
More specifically, following the notation of Eq.~\ref{eq:linear-representation-hypothesis}, we: \emph{i)} Collect the one-hot encodings of all $2^{6}-1\,{=}\,63$ nodes of the tree, including the $32$ tokens, in a single vector $e_\mathbf{Z}$;
\emph{ii)} Fit the matrices $A^{\scriptscriptstyle(\ell)}_t$ via ordinary least squares on the activations.
Note that $A_t^{\scriptscriptstyle(\ell)}$ is a $d$-by-$(v\times63)$ matrix, assigning a different $d$-vector to each possible value (out of $v$) of any of the $63$ variables $Z$.
We refer to each block of $v$ $d$-vectors as the \emph{code} of $Z$.
In addition, the residuals $\epsilon^{\scriptscriptstyle(\ell)}_t$ are uncorrelated with every node by construction.
Hence, Eq.~\ref{eq:linear-representation-hypothesis} is the best approximation of the activation as a sum of single-variable codes, with the residual collecting non-additive contributions.
Appendix~\ref{app:probing} details the regression method, its distinction from linear probes, and complete encoding maps with the share of activation's variance explained by each latent variable.
Here we summarise the results and compare them to the construction of Fig.~\ref{fig:main}d), which assumes that each layer resolves one level of parent-child dependencies.
\begin{itemize}
 \item  Layer $0$ (embedding layer) follows the sketch exactly: each activation can only represent the corresponding input token (blue in the figure). In fact, because of causal masking, the activations in position $1$ depend only on $X_1$ \emph{at all layers}. Hence, position-$1$ activations encode $X_1$ independently of the layer.
 \item In layer $1$, the encoding of position $2$ changes to the parent latent variable $X^{\scriptscriptstyle(L-1)}_1$ (orange) as predicted. The parent perfectly summarises the effect of the input pair $(X_1,X_2)$ on future tokens: its encoding is favourable for both representational and sample efficiency. The token-parent pattern of positions $1$ and $2$ repeats periodically across the whole context in layer $1$ (sketch in Fig.~\ref{fig:main}d) and measured codes in Fig.~\ref{fig:rhm-tree-encodings-layer1}). This is compatible with the assumption that each LLM layer resolves one RHM layer. For example, the correlation between $(X_3,X_4)$ and $(X_1,X_2)$ is not visible in layer $1$, because it stems from the depth-$2$ dependency between the parents $X^{\scriptscriptstyle(L-1)}_1$ and $X^{\scriptscriptstyle(L-1)}_2$.
 \item Layer $2$ resolves the second RHM layer as predicted, and the encodings of positions $3$ and $4$ change accordingly. Position $3$ acquires an encoding of the last parent $X^{\scriptscriptstyle(L-1)}_1$ (illustrated as a $2$-slot box in Fig.~\ref{fig:main}d)), whose influence on $(X_3,X_4)$ is now visible. Position $4$ summarises the length-$4$ input $X_{1:4}$ with the corresponding grandparent variable $X^{\scriptscriptstyle(L-2)}_1$ (green). The resulting length-$4$ pattern repeats periodically across the whole context (sketch in Fig.~\ref{fig:main}d), reproduced by the measured codes in Fig.~\ref{fig:rhm-tree-encodings-layer2}).
 \item Layer $3$ follows this construction: position $8$ encodes the great-grandparent $X^{\scriptscriptstyle(L-3)}_1$ (red), and positions $5$-to-$7$ acquire an encoding of the last grandparent $X^{\scriptscriptstyle(L-2)}_1$. However, the measured codes also reveal structures expected to be found in layer $4$: a partial encoding of the depth-$4$ variable $X^{\scriptscriptstyle(L-4)}_1$ in the first half of the context, and persistence of the encoding of the great-grandparents $X^{\scriptscriptstyle(L-3)}_i$ beyond their first emergence, albeit with decreased strength (e.g., $X^{\scriptscriptstyle(2)}_3$ in Fig.~\ref{fig:rhm-tree-encodings-layer3}). In simple terms, the one-to-one correspondence between LLM and RHM layers is not as clean for layers $3$ and $4$ as it was for layers $1$ and $2$.
\end{itemize}

\begin{figure}[h!]
  \centering
  \begin{subfigure}[t]{0.5\textwidth}
    \vspace{0pt}
    \centering
    \includegraphics[width=\linewidth]{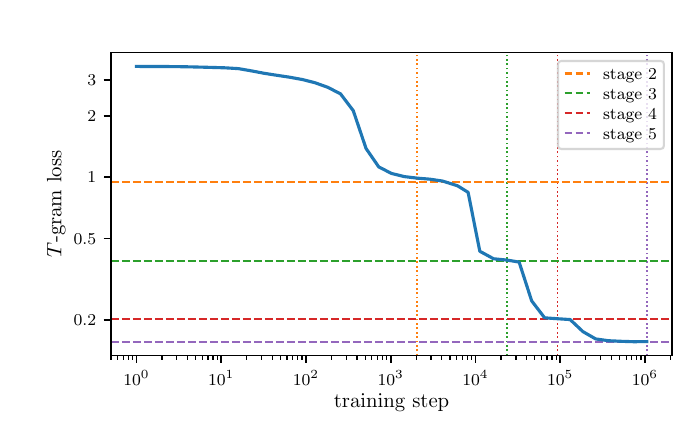}
    \label{fig:rhm-dynamics-loss}
  \end{subfigure}%
  \begin{subfigure}[t]{0.5\textwidth}
    \vspace{0pt}
    \centering
    \includegraphics[width=\linewidth]{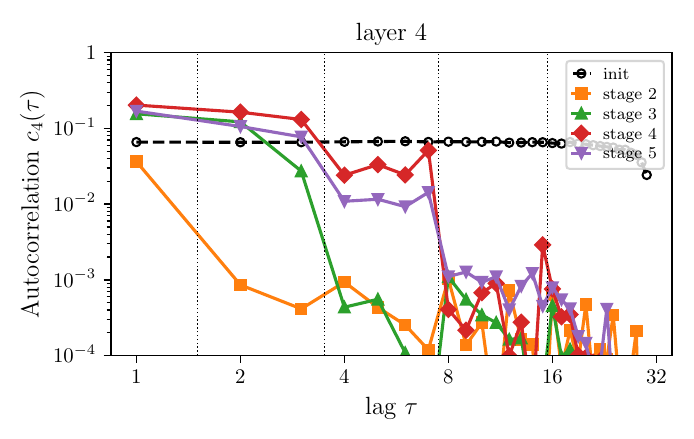}
    \label{fig:rhm-dynamics-trace}
  \end{subfigure}%
  \vspace{-20pt}
  \caption{
    \textbf{RHM: the autocorrelations develop stage by stage.}
    \textbf{Left:} Validation loss of the last token, $X_{32}$ given $X_1,\dots,X_{31}$, against the training step.
    Dashed lines: the plateau values of the stages, i.e. the conditional entropy of depth-$h$ subtrees.
    Dotted lines: the checkpoints of panel b, coloured by stage.
    \textbf{Right:} Absolute value of the autocorrelation of layer $4$ at initialisation (dashed) and at the four checkpoints; dotted vertical lines at lags $1.5$, $3.5$, $7.5$ and $15.5$ as in Fig.~\ref{fig:main}a, corresponding to the predicted locations of the steps in the autocorrelations.
  }
  \label{fig:rhm-dynamics}
\end{figure}

\paragraph{Composition of the autocorrelations.}
Persistence alone predicts the shape of the curves in Fig.~\ref{fig:main}a, as explained by Fig.~\ref{fig:rhm-trace-composition} for layer $4$ (see Appendix~\ref{app:rhm-composition} for additional details and the other layers).
The reference activation is the previous-to-last position $t\,{=}\,31$, since $X_{32}$ is never an input.
Due to the tree structure, the latent variables sitting on the tree diagonal (left panel) perfectly summarise the influence of their descendants on the probability of $X_{32}$~\citep{cagnetta2024towards}: $X_{31}$, then $X^{\scriptscriptstyle(4)}_{15}$, parent of $X_{29:30}$,~\footnote{$X_{i:j}$ represents the span of tokens $X_i,X_{i+1},\dots,X_j$.} then $X^{\scriptscriptstyle(3)}_{7}$, grandparent of $X_{25:28}$,  and $X^{\scriptscriptstyle(2)}_{3}$, great-grandparent of $X_{17:24}$ (superscripts count levels from the root).
The right panel compares the measured autocorrelation with the same-node terms of the three relevant latent variables (colour code in the key).
Each term is non-zero only over the descendants of its latent variable, following the sketch of Fig.~\ref{fig:main}d, and vanishing as soon as the target position leaves the corresponding subtree: at lag $2$ for $X^{\scriptscriptstyle(4)}_{15}$, lag $4$ for $X^{\scriptscriptstyle(3)}_{7}$ and lag $8$ for $X^{\scriptscriptstyle(2)}_{3}$.
Fig.~\ref{fig:rhm-trace-shares} in Appendix~\ref{app:rhm-composition} stacks the shares of the measured autocorrelation: the predicted codes account for $55$--$95\%$ of the measurement, the rest stemming mostly from the non-additive residual and confined to the first plateau.
\begin{figure}[h!]
  \centering
  \includegraphics[width=\linewidth]{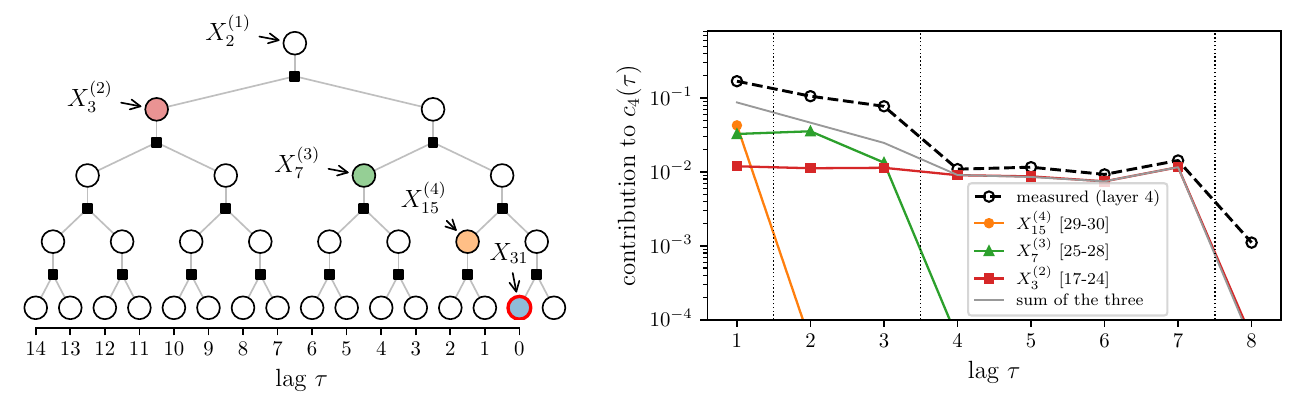}
  \vspace{-20pt}
  \caption{
    \textbf{RHM: composition of the autocorrelations of layer $4$.}
    \textbf{Left:} the relevant latents $X^{\scriptscriptstyle(4)}_{15}$, $X^{\scriptscriptstyle(3)}_{7}$ and $X^{\scriptscriptstyle(2)}_{3}$ in the subtree of the anchor $X_{31}$ (red ring), with the lag axis below the leaves.
    \textbf{Right:} measured autocorrelation $c_4(\tau)$ (dashed) against the same-node terms of Eq.~\ref{eq:rhm-trace-split} of the three relevant latents (spans in brackets) and their sum (grey).
    Vertical dashed lines: the span ends of $X^{\scriptscriptstyle(4)}_{15}$ (between lags 1 and 2), $X^{\scriptscriptstyle(3)}_{7}$ (between lags $3$ and $4$), and $X^{\scriptscriptstyle(2)}_{3}$ (between $7$ and $8$).
  }
  \label{fig:rhm-trace-composition}
\end{figure}
\paragraph{Development over training.}
Persistence also predicts how the autocorrelation function changes during training.
For deep transformers trained on RHM data, the learning curves proceed in stages, each corresponding to the resolution of one more level of the latent hierarchy~\citep{cagnetta2024towards, cagnetta2025scaling}.
Fig.~\ref{fig:rhm-dynamics}a shows this for the validation loss of the last token, $X_{32}$, given the preceding ones, which decreases through a sequence of plateaus (the full autoregressive loss has identical stages; see Appendix~\ref{app:rhm-training}).
The plateaus sit at the theoretical entropies of depth-$h$ subtrees (dashed lines), compatibly with a model that has learned only a portion of the latent hierarchical structure, hence treats blocks of $2^h$ tokens as independent. 
Like the optimal loss, these plateau values are exact functions of the rules (Appendix~\ref{app:rhm-setup}).
We measure the autocorrelations at initialisation and at one checkpoint on each plateau (vertical dotted lines).
Fig.~\ref{fig:rhm-dynamics}b shows the result for layer $4$: the staircase is built outward, one plateau per stage.
At initialisation, the autocorrelation is flat and independent of the lag, as attention is nearly uniform over the context (Appendix~\ref{app:rhm-training}).
Training removes this uniform component, and the staircase grows one plateau at a time.
At stage $2$, where depth-$2$ subtrees are learned, only lag $1$ stands out (note the logarithmic scale).
At stage $3$, the plateau at lags $1$--$3$ is in place; at stage $4$, the plateau at lags $4$--$7$ appears, then it settles at stage $5$.
The other layers follow the same stagewise behaviour, with their own reach (Appendix~\ref{app:rhm-training}).

\section{Autocorrelations in natural language}
\label{sec:nlp}

In this section, we verify the relationship between autocorrelations, persistence, and depth of latent features in open LLMs trained on natural language.
We used five LLMs for which open source trained SAE features are available: Gemma 2 2B and Gemma 2 9B~\citep{gemmateam2024gemma2} with the residual-stream JumpReLU SAEs of Gemma Scope~\citep{lieberum2024gemmascope}, Pythia-160M~\citep{biderman2023pythia} and Llama-3-8B~\citep{grattafiori2024llama3} with the top-$k$ SAEs~\citep{gao2024scaling} released by EleutherAI~\citep{eleutherai2024sparsify}, and Qwen3-8B~\citep{yang2025qwen3} with the top-$k$ SAEs of Qwen-Scope~\citep{deng2026qwenscope}.
For datasets we used WikiText~\citep{merity2017pointer} and The Pile~\citep{gao2020pile}, standard open source corpora in English. 

\paragraph{Integrated autocorrelation.}
By summing over time lags $\tau$ we get a scalar value quantifying the amount of autocorrelation of each layer, the \emph{integrated} autocorrelation
\begin{equation}
    \chi_\ell = \sum_\tau c_\ell (\tau).
    \label{eq:susceptibility}
\end{equation}
$\chi$ is called \textit{susceptibility} in equilibrium statistical mechanics, where integrated correlations are related to linear response to perturbations via the fluctuation–dissipation relations~\cite{goldenfeld2018lectures}.  
For systems in which correlations change signs, Eq.~\ref{eq:susceptibility} could average to zero even for finite values of $c_\ell(\tau)$, but this is not observed in our setup. 

\paragraph{SAE and Persistence.}
SAE features are human-interpretable components extracted from the dense, entangled activation spaces of LLMs. SAEs do not represent the entire activation space: we report the fraction of the activation's variance explained by the SAE reconstructions in Appendix~\ref{sec:SAE-explained-variance}. An SAE feature is \emph{active} at a token position $t$ when its code is non-zero. The persistence of a firing episode of an SAE feature is the number of consecutive tokens on which it is active: $\psi\,{=}\,e\,{-}\,s+1$ for starting position $s$ and ending position $e$ (see Fig.~\ref{fig:SAE-ablation-scheme}, upper row).
We report persistencies averaged per feature: record $\psi$ for each firing episode of a given feature and average over firing episodes.

\subsection{Layer dependencies}

\begin{figure}[t]
    \centering
    \includegraphics[width=0.668\linewidth]{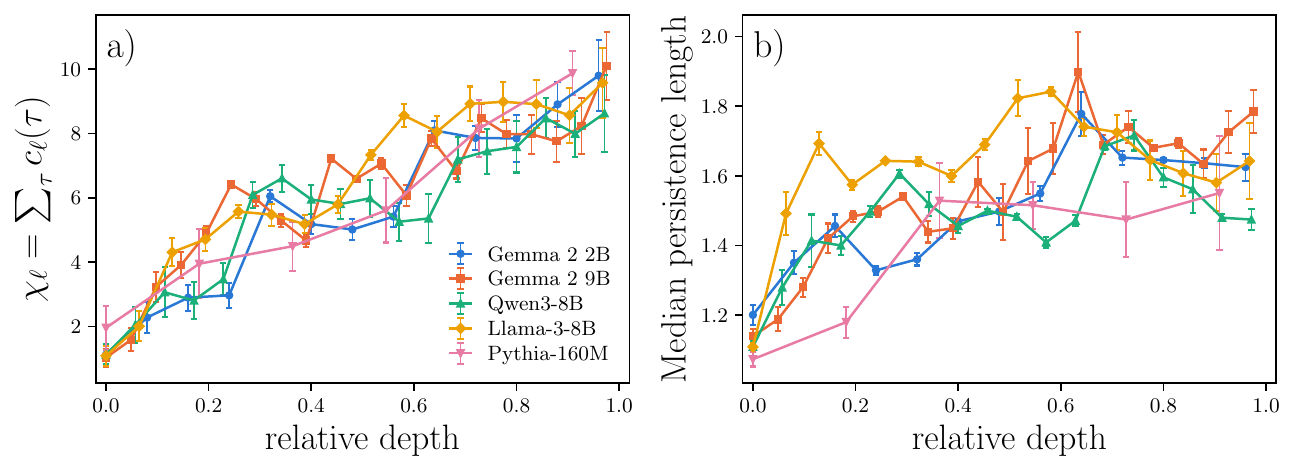}\hfill
    \includegraphics[width=0.328\linewidth]{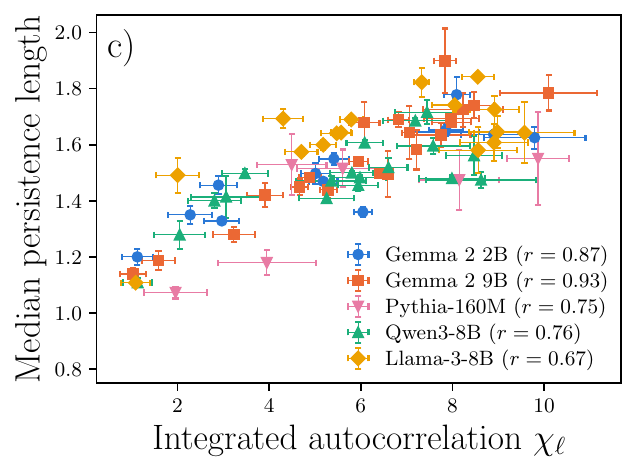}
    \caption{
      \textbf{ Panel a)}: Layer dependence of the integrated autocorrelation for different models (see Eq.~\ref{eq:susceptibility}). 
      \textbf{ Panel b)}: Layer dependence of the median of the distribution of per-latent persistencies $\psi$ (see Sec.~\ref{sec:nlp}).
      \textbf{ Panel c)}: Median persistence against the integrated autocorrelation $\chi_\ell$, one point per layer; $r$ is the per-model Pearson correlation.
    In all panels, error bars show one standard deviation computed from results on the Wikitext and The Pile datasets. 
    }
    \label{fig:layer-dependences-NLP}
\end{figure}
Fig.~\ref{fig:layer-dependences-NLP}a and Fig.~\ref{fig:layer-dependences-NLP}b show the layer dependence of the integrated autocorrelations (Eq.~\ref{eq:susceptibility}) and the median persistence time of SAE features, respectively, across models and datasets.
Both quantities increase roughly monotonically with depth, although the median persistence tends to plateau or slightly decrease in the last quarter of the networks (see Appendix~\ref{app:SAE-layer-dependencies} for the mean, showing a similar trend).
Accordingly, they are strongly correlated across layers, as shown in Fig.~\ref{fig:layer-dependences-NLP}c for all models.

This correlation could, however, originate from the activation frequency of the features rather than from their persistence.
On the one hand, a frequently active feature has longer runs by chance.
On the other hand, at large lags the autocorrelations of Fig.~\ref{fig:main}b approach a plateau that grows with depth: a sequence-wide component, given by the mean representation of each sequence, which depends on how often each feature is active in that sequence and not on when.
Appendix~\ref{app:subtracting-mean-chi} separates the two origins by randomly shuffling the token positions of each sequence, which preserves the activation frequencies and destroys the temporal order.
After shuffling, the growth of persistence with depth largely disappears, whereas the autocorrelation reduces to the sequence-wide component (Fig.~\ref{fig:frequency-control}).
Hence, frequency does not explain the growth of persistence, and its contribution to the autocorrelations can be removed by subtracting from the activations the mean of their sequence.
The resulting integrated autocorrelation $\tilde{\chi}_\ell$ remains strongly correlated with persistence, in most cases more than $\chi_\ell$ (see Table~\ref{tab:docmean-correlations}).
Furthermore, $\tilde{\chi}_\ell$ has non-monotonic layer dependence with an approximate plateau in the second half of the network (see Fig.~\ref{fig:mean-token--susceptibility}).
Interestingly, this layer profile is compatible with recent results in which measures of representational abstraction peak or saturate in intermediate layers~\cite{cheng2025emergence,acevedo2026semantic,acevedo2026differential,gurnee2026workspace}.

\subsection{Intervening on activations: ablation of SAE features}\label{sec:SAE-ablations}

In this section, we verify that projecting out persistent features from the activations significantly decreases autocorrelations.
Given the neural activity of an LLM layer $\phi_t$~\footnote{We omit the layer superscript in this section to ease notation, and restrict the analysis to the deepest layer for which there is an available SAE online.} and an SAE feature $i$, ablation is performed by removing the projection of $\phi_t$ along the decoding vector $W_{\rm dec}(i) $ of the SAE decoder:
\begin{equation}
    \tilde{\phi}_t = \phi_t - \frac{(\phi_t,W_{\rm dec}(i))}{||W_{\rm dec}(i)||^2} W_{\rm dec}(i).
    \label{eq:ablation-SAE}
\end{equation}
We ablate token positions $t=s,s+1,...,e-1$, such that the reference token $e$ is not simultaneously ablated, then report the mean change in autocorrelations 
\begin{equation}
\Delta = \frac{1}{e\,{-}\,s} \sum_{t=s}^{e-1} \mathbb{E} \left[(\tilde\phi_{t}-\phi_{t}) \cdot \phi_{e}\right]
\label{eq:Delta-ablation}
\end{equation}
$\Delta$ is further averaged over the 500 longest firing episodes of the layer's 400 most persistent latents, with at most 2 episodes per latent, and expressed as a percentage of the average clean value~\footnote{Eq.~\ref{eq:ablation-SAE} avoids overestimating the magnitude of a single feature ablation caused by the imperfect reconstruction of the SAE.}.

\begin{figure}
    \centering
    \includegraphics[width=\linewidth]{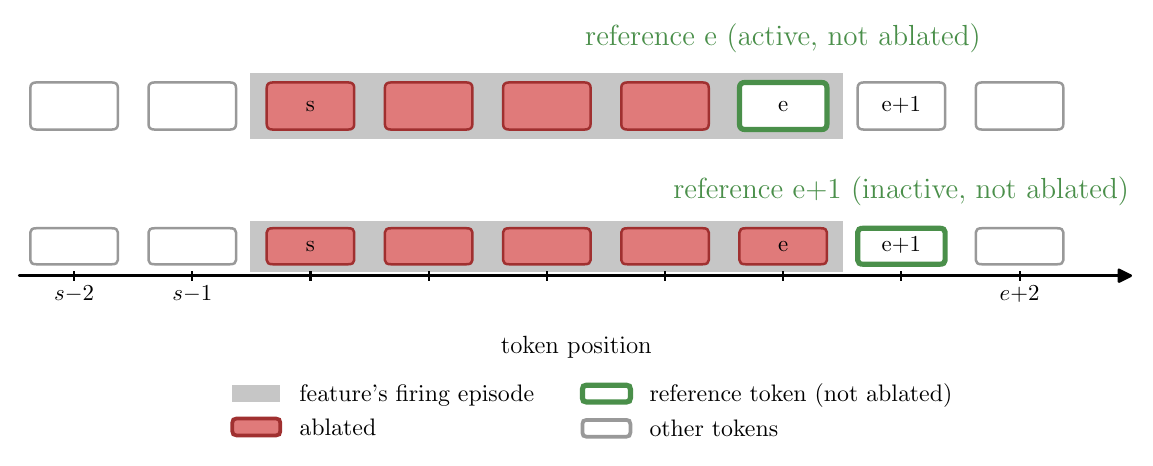}
    \caption{Scheme of the intervention experiment: 
    In the gray areas, a given SAE-feature of interest is firing. $s$ and $e$ denote the starting and ending token positions of the firing episode. 
    In the upper row, we measure the mean autocorrelations inside the feature's firing episode, taking as reference the last token of the firing episode, signaled with a green box. 
    Then, we ablate SAE-features of the tokens colored in red using different protocols, and recompute the mean autocorrelation, to quantify how the ablation of SAE-features influences the autocorrelation. 
    Results are reported in Table~\ref{tab:causal-ablation}.
    In the bottom row, we show an example of a control experiment in which the reference token is outside the firing episode, for which the decrease in autocorrelations generated by ablating SAE-features is smaller, since the feature is not active in it. 
    }
    \label{fig:SAE-ablation-scheme}
\end{figure}

The third and fourth columns of Table~\ref{tab:causal-ablation} show the largest value of $\Delta$ obtained by removing one and three SAE features, as depicted in Fig.~\ref{fig:SAE-ablation-scheme}. 
The fifth column (a-control) shows the value of a control experiment in which we remove an SAE feature that was \textit{active} at position s, but \textit{not persistent} inside the window, for which we expect a much smaller $\Delta$.  
The ablation of all active latents (sixth column) using the SAE intermediate representation does not remove all the baseline autocorrelations (i.e., $|\Delta| < 100$), as expected from imperfect SAE reconstructions (reported in Appendix~\ref{sec:SAE-explained-variance}).
The last three columns of Table~\ref{tab:causal-ablation} show the magnitude of $\Delta$ for the case in which the reference token is outside the measured firing episode of the latent ($e+1,e+2,e+3$, see Fig.~\ref{fig:SAE-ablation-scheme} bottom row). As expected, the effect of the ablation is not negligible, but smaller than that measured using position $e$ as reference (third column), and moreover it decays with the distance to the $e$ token, for all models except Pythia. 

\begin{table}[h!]
\centering
\small
\begin{tabular}{llrrrrrrr}
\hline
Model ($\Delta$, \%) & corpus & 1 latent & 3 latents & all latents & a-control & $e{+}1$ & $e{+}2$ & $e{+}3$ \\
\hline
Gemma 2 2B & Pile & $-3.9$ & $-7.5$ & $-58.8$ & $-0.0$ & $-0.4$ & $+0.0$ & $+0.2$ \\
Gemma 2 2B & WikiText & $-3.2$ & $-5.8$ & $-68.3$ & $+0.0$ & $-0.5$ & $-0.2$ & $+0.0$ \\
Gemma 2 9B & Pile & $-3.2$ & $-6.2$ & $-53.0$ & $-0.0$ & $-1.2$ & $-1.3$ & $-1.2$ \\
Gemma 2 9B & WikiText & $-3.0$ & $-4.6$ & $-53.1$ & $+0.0$ & $-1.3$ & $-1.0$ & $-1.1$ \\
Pythia-160M & Pile & $-12.2$ & $-18.0$ & $-43.7$ & $-0.7$ & $-8.4$ & $-9.4$ & $-7.0$ \\
Pythia-160M & WikiText & $-13.3$ & $-24.3$ & $-60.4$ & $-0.7$ & $-8.1$ & $-8.6$ & $-9.0$ \\
Llama-3-8B & Pile & $-4.4$ & $-5.6$ & $-68.1$ & $-0.1$ & $-2.1$ & $-1.7$ & $-1.5$ \\
Llama-3-8B & WikiText & $-4.2$ & $-4.4$ & $-64.3$ & $+0.0$ & $-1.8$ & $-1.4$ & $-1.2$ \\
Qwen3-8B & Pile & $-6.8$ & $-20.3$ & $-65.8$ & $-0.2$ & $-4.2$ & $-4.6$ & $-3.5$ \\
Qwen3-8B & WikiText & $-7.3$ & $-10.3$ & $-67.6$ & $-0.1$ & $-3.8$ & $-3.7$ & $-3.5$ \\
\hline
\end{tabular}
\caption{
Change $\Delta$ (Eq.~\ref{eq:Delta-ablation}) of the mean integrated autocorrelations generated by the ablation of SAE-features, across models and two different corpora, and expressed as a percentage of the value of the baseline correlations in each model. See Fig.~\ref{fig:SAE-ablation-scheme} for a scheme of the ablation setup, and Sec.~\ref{sec:SAE-ablations} for the full discussion about the experiment.
Columns: the chosen persistent latent; it plus its two most persistent co-active latents; every active feature latent using the SAE's intermediate representation; an active but non-persistent latent (a-control); and the latent itself with the reference moved to $e{+}1$, $e{+}2$ and $e{+}3$ token positions, where the latent of interest was not found to be active.
}
\label{tab:causal-ablation}
\end{table}

\section{Discussion and perspectives}

In this work, we explored the connection between the autocorrelations of neural activations, Eq.~\ref{eq:autocorrelations}, and the encoding of latent variables.
At the token level, with one-hot representations, Eq.~\ref{eq:autocorrelations} reduces to the centered probability of token coincidence.
Under the linear representation hypothesis, it extends the coincidence to latent codes, i.e., the persistence of variables encoded along aligned directions at both positions.
Therefore, as at the token level, the functional form of autocorrelations reveals the structure of the underlying process.
Note that autocorrelations are read from the raw activations, without downstream tasks or predefined features: the organisation of the encoded variables is probed without even identifying their content.
Thus, our results highlight the value of geometric observables for studying high-dimensional neural representations, taking a step toward connecting the geometry of LLM representations to the information they encode~\citep{shai2024transformers}.

Eq.~\ref{eq:autocorrelations} is invariant under a common change of orthonormal basis: a latent variable contributes only if it is encoded along aligned directions at both positions.
While this property makes Eq.~\ref{eq:autocorrelations} a measure of persistence, nothing forces a model to write a variable along the same direction at different positions.
Measures that include such an effect would rely on the full cross-covariance matrix, $\mathbb{E}\big[\phi^{\scriptscriptstyle(\ell)}_{T-\tau}\,\phi^{\scriptscriptstyle(\ell)\top}_{T}\big] \in \mathbb{R}^{D\times D}$.
Comparing these measures with Eq.~\ref{eq:autocorrelations}, which corresponds to the trace of the matrix, could establish how much of the persistence of a layer is carried by fixed directions, as in our reading of Eq.~\ref{eq:autocorrelations}, and how much by codes that change orientation across positions.

As a perspective for incoming research, we would like to connect correlations to the response to perturbations.
In equilibrium statistical mechanics, fluctuation--dissipation relations link the correlations of the unperturbed system to its linear response to weak external perturbations.
Transformer representations do not constitute an equilibrium dynamical system, so such a relation is not expected to hold a priori.
Nevertheless, first-order response matrices, such as the Jacobian lens~\cite{gurnee2026workspace}, have the potential to help us better understand how transformers process information internally.
In this context, our results motivate asking whether the temporal structure of $c_\ell(\tau)$ predicts the lag-resolved response to small perturbations of earlier tokens or activations---for instance, whether layers with longer-ranged correlations also propagate perturbations over longer context ranges.
Establishing such a fluctuation--response relation, or characterising its violations, would connect the passive geometrical observable studied here to causal measures of information propagation.

\subsection*{Acknowledgements}

The authors would like to thank Matthieu Wyart, Konstantin Nikolaou and Jasmina Urdshals for useful discussions and feedback on the manuscript. For part of this project, the work of FC was supported by the European Union's Horizon Europe program under the Marie Sk\l{}odowska-Curie grant agreement No.~101154584.

\bibliography{iclr2027_conference}
\bibliographystyle{iclr2027_conference}
\newpage

\appendix
\section{Random Hierarchy Model: data, learning model and training}
\label{app:rhm-setup}

\paragraph{Generative process.}
The RHM~\citep{cagnetta2024random} generates sentences of $T\,{=}\,2^L$ tokens along a binary tree of depth $L$, all of whose variables, latent and visible, take values in a set of $v$ symbols.
At each level $l\,{=}\,0,\dots,L\,{-}\,1$ below the root, each symbol$a\,{=}\,1,\dots,v$ is assigned $m$ production rules $a\to(b,c)$, replacing it with a pair of symbols of level $l+1$. Rules are \emph{unambiguous}: The $vm$ right-hand sides are sampled uniformly without replacement among the $v^2$ possible pairs, so that no pair is produced by two different symbols.
Sentence generation proceeds as follows.
First, the root is drawn uniformly.
Secondly, the root symbol is expanded by one of its $m$ rules, chosen uniformly and independently of everything else.
Each of the resulting symbols is itself expanded by one of its $m$ rules, chosen uniformly and independently of everything else.
The iterative expansion ends after $L$ levels.
The $k$-th variable at level $\ell$ is denoted with $X^{\scriptscriptstyle(\ell)}_k$, having depth (i.e. distance from the input tokens) $h\,{=}\,L-\ell$. $X^{\scriptscriptstyle(\ell)}_k$ spans the $2^h$ tokens $X_{\scriptscriptstyle (k-1)2^h+1},\dots,X_{\scriptscriptstyle k2^h}$.
The root spans the whole sentence, while depth-$1$ variables span pairs of adjacent input tokens.
We use $L\,{=}\,5$, $v\,{=}\,32$ and $m\,{=}\,8$: $2^5\,{=}\,32$ input tokens, $31$ hidden nodes and $vm\,{=}\,256$ grammatical pairs per level out of $v^2\,{=}\,1024$.
Unambiguity makes the map from a sentence to its tree of latent variables deterministic, and we recover the latent variables of every sentence by parsing it bottom-up.

\paragraph{Optimal loss.}
Since the generative model is a tree, the next-token probability $\mathbb{P}(X_{t+1} \mid X_{1:t})$ is computed exactly by belief propagation, and its conditional entropy $v$, averaged over contexts, is the optimal cross-entropy loss at position $t$.
The average over all positions,
\begin{equation}
    \frac{1}{T} \sum_{t=0}^{T-1} H(X_{t+1} \mid X_{1:t}) =  \frac{1}{T}H(X_{1:T}) 
\end{equation}
is the optimal autoregressive loss.

\paragraph{Model.}
We consider a decoder-only GPT-2-style transformer: learned token and absolute positional embeddings of dimension $d\,{=}\,768$, $5$ pre-LayerNorm blocks with $12$ attention heads and an MLP of width $4d$, a final LayerNorm and a linear readout; no dropout, no weight decay.
Weights are initialised as standard Gaussians with the $1/\sqrt{\text{fan-in}}$ factors applied in the forward pass, and the readout is scaled by $1/d$, so that the initial predictions are close to uniform.
The input is a [BOS] (beginning-of-sequence) token followed by the first $31$ tokens of a sentence, and the model predicts $X_1,\dots,X_{32}$ autoregressively.
Layer $\ell\,{=}\,1,\dots,5$ of the main text is the residual stream after block $\ell$, and layer $0$ is the sum of the token and positional embeddings.

\paragraph{Training.}
The model is trained on the standard cross-entropy loss using Adam, with learning rate $3\times 10^{-3}$, batch size $1024$, linear warm-up over $8$ steps and cosine decay to learning rate $3\times 10^{-4}$ over $2^{19}$ steps. The training set consists of $2^{28}$ sentences, and the model is trained for $2^{20}$ steps (four passes over the training set).
The validation loss of the final checkpoint is within $0.033$ nats per sentence of the optimal loss, and within $0.007$ nats of the optimal $H_t$ at every position.
All results are for a single grammar and a single training run.

\section{Token and activation statistics in the RHM}
\label{app:rhm-statistics}

This appendix details the measurement of activation statistics for models trained on the RHM data of Appendix~\ref{app:rhm-setup}, including comparisons with analytical predictions of the autocorrelation function based on the construction of Fig.~\ref{fig:main}d.

\paragraph{Activation statistics.}
We compute all activation statistics, including autocorrelations, from $P\,{=}\,2^{25}$ sentences sampled from the grammar independently of the LLM's training set.
For each layer, including tokens and embeddings, we compute the mean and the second moments of the activations at all positions, from which standardised activations are formed:
\begin{align}
\tilde\phi_t \to \phi_t \coloneqq \frac{\tilde\phi_t-\mathbb{E}[\tilde\phi_t]}{\sqrt{\mathrm{Tr}(\Sigma_t)}},\quad\text{with}\quad\Sigma_t = \mathbb{E}\left[ \phi_t \phi_t^\top\right]\Rightarrow \mathrm{Tr}(\Sigma_t) = \mathbb{E}\| \phi_t \|^2.
\end{align}
In terms of the original, non-standardised activations, the autocorrelation function of Eq.~\ref{eq:autocorrelations} reads (layer index omitted for clarity)
\begin{align}
c(\tau) = \frac{\mathbb{E}\left[\left(\tilde\phi_{T-\tau}-\mathbb{E}[\tilde\phi_{T-\tau}]\right)\cdot\left(\tilde\phi_T-\mathbb{E}[\tilde\phi_T]\right)\right]}{\sqrt{\mathrm{Tr}(\Sigma_{T-\tau})\mathrm{Tr}(\Sigma_{T})}}.
\end{align}
For two independent centered activations, the sampling variance of their average dot product is $\mathrm{Tr}(\Sigma_{T} \Sigma_{T-\tau})/P$, so the sampling variance on $c(\tau)$ is $1/\sqrt{rP}$ with the effective rank $r\,{=}\,\mathrm{Tr}\Sigma_{T}\,\mathrm{Tr}\Sigma_{T-\tau}/\mathrm{Tr}(\Sigma_{T}\Sigma_{T-\tau})$. This noise floor lies below $10^{-4}$ at every lag and layer, and the curves are stable in $P$ wherever they lie above it.

\subsection{Token- and embedding-level autocorrelations}
\label{app:rhm-embeddings}

This appendix supports the discussion of Sec.~\ref{sec:autocorr-explained} for RHM data.
We first compute the token-level autocorrelation of Eq.~\ref{eq:token-autocorr-unnorm} analytically from the RHM's production rules, then show that embedding the tokens, with either random or trained vectors, leaves it unchanged. We repeat both steps for the idealised first-layer construction of Fig.~\ref{fig:main}d in Sec.~\ref{app:rhm-layer1}.

\begin{figure}[h]
    \centering
    \includegraphics[width=\linewidth]{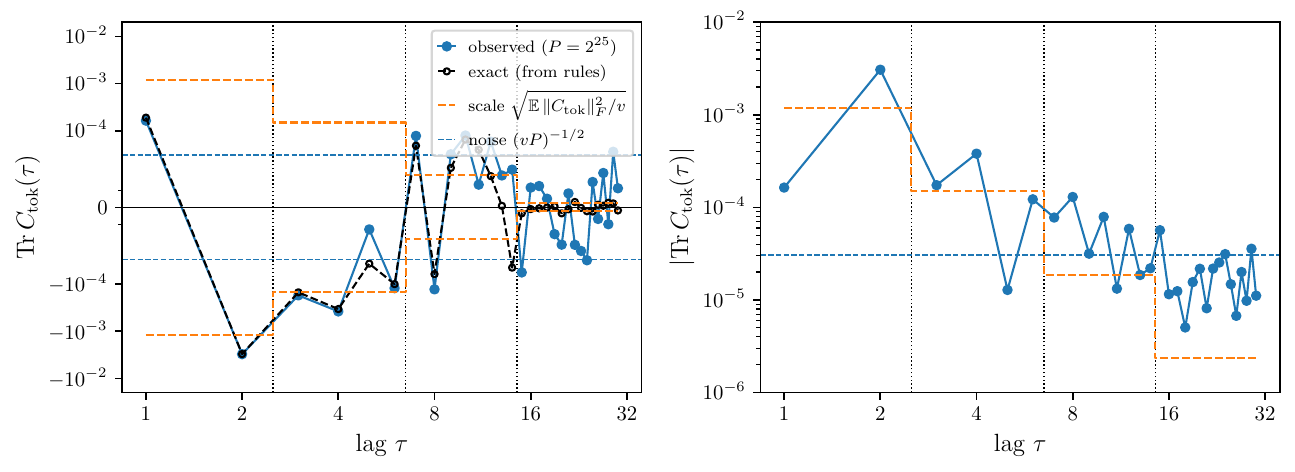}
    \caption{Token-level trace $\mathrm{Tr}\,C_{\mathrm{tok}}(\tau)$ between the one-hot encodings of $X_{31}$ and $X_{31-\tau}$: measured on $2^{25}$ sentences (blue), exact from the rules (Eq.~\ref{eq:two-leaf-joint}, black), the scale $\pm\sqrt{\mathbb{E}\,\|C\|_F^2/v}$ of Eq.~\ref{eq:trace-scale} (orange) and the sampling noise $\pm1/\sqrt{vP}$ (horizontal lines). Left: signed values on a symmetric-log axis; right: absolute values. Dotted lines at the lags where the lowest common ancestor of the two tokens moves up one level.}
    \label{fig:token-trace}
\end{figure}

We define the token-level correlation matrix as the $v\times v$ cross-covariance of the one-hot encodings,
\begin{align}
C_{\mathrm{tok}}(\tau)\,{\coloneqq}\,\mathbb{E}\big[(e_{T-\tau}-\mathbb{E}[e_{T-\tau}])(e_{T}-\mathbb{E}[e_{T}])^\top\big].
\end{align}
Eq.~\ref{eq:token-autocorr-unnorm} is the trace of $C_{\mathrm{tok}}(\tau)$. In the RHM, two tokens are coupled only through the rule drawn at their lowest common ancestor.
Let $Z$ be this ancestor, $p_Z$ its marginal, and $R(z)$ the set of $m$ pairs that the symbol $z$ can produce.
Call $\Pi_{c\to a}$ the probability that a node with value $c$ has the descendant of value $a$ at position $T$ is the product of one-step transition matrices along the path, and $\Pi'$ the same along the path to position $T-\tau$.
Then
\begin{align}\label{eq:two-leaf-joint}
    \mathbb{P}\left(X_{T-\tau}=b,\,X_{T}=a\right) = \sum_z p_Z(z)\,\frac{1}{m}\sum_{(c',c)\in R(z)} \Pi'_{c'\to b}\,\Pi_{c\to a},
\end{align}
and Eq.~\ref{eq:token-autocorr-unnorm} follows by summing the diagonal $a\,{=}\,b$ and subtracting the product of the marginals.
The marginal $p_Z$, the set of pairs $R(z)$ and the transition probabilities $\Pi$ and $\Pi'$ are all given in the RHM.
Hence, Eq.~\ref{eq:token-autocorr-unnorm} is available analytically.
Fig.~\ref{fig:token-trace}, left compares the result (black dashed) with the empirical token-level autocorrelation (blue): the two agree within $2$ standard deviations of the sampling noise, $1/\sqrt{vP}\,{=}\,3\times10^{-5}$(Appendix~\ref{app:rhm-setup}, blue dashed line in the figure), at every lag.

\paragraph{The scale of autocorrelations.}
While the sign and specific value of each point depend on the specific instance of the grammar, the overall scale of the autocorrelations is fixed by the structure of the tree and the parameters $v$ (vocabulary size of input and latent variables) and $m$ (number of rules per symbol).
Specifically, $\mathrm{Tr}\,C_{\mathrm{tok}}(\tau)$ is a random variable with zero mean over the random draw of the rules, while the standard deviation gives its typical size.
We estimate its standard deviation by treating the $v$ diagonal entries of $C_{\mathrm{tok}}$ as entries of typical size and random sign, which gives $\sqrt{\mathbb{E}\,\|C_{\mathrm{tok}}\|_F^2/v}$.
The Frobenius norm follows from~\citet{cagnetta2024towards} in the large $v$ limit with $m\propto v$: the two children of a uniform parent give $\|C\|_F^2\,{=}\,(1-m/v)/(vm)$ exactly, and each further step down the tree divides it by about $m$, since the centred rows of the transition matrices are nearly orthogonal with squared length $1/m$~\citep{parley2026deep}.
For two nodes $h$ and $h'$ levels below their lowest common ancestor,
\begin{align}\label{eq:trace-scale}
    \mathbb{E}\,\|C\|_F^2 \simeq \frac{1-m/v}{v\,m^{h+h'-1}}.
\end{align}
For two tokens, $h\,{=}\,h'$, this is a staircase with one step per level of the ancestor (orange dashed in Fig.~\ref{fig:token-trace}).
The exact trace scatters around the scale with an order-one ratio, $0.9$--$2.3$ in root-mean-square over each level.
From lag $7$ on, the scale is below the sampling noise.

\begin{figure}[h]
    \centering
    \includegraphics[width=\linewidth]{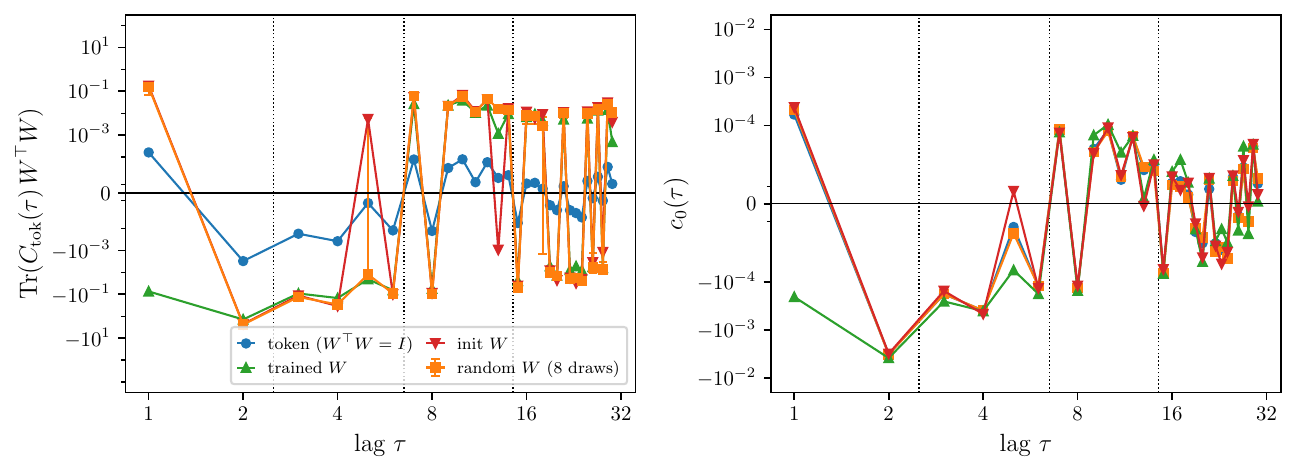}
    \caption{Token-level trace pushed through an embedding, $\mathrm{Tr}(C_{\mathrm{tok}}\,W^\top W)$ (left), and the normalised autocorrelation $c_0(\tau)$ (right), for the one-hot tokens ($W^\top W\,{=}\,\mathbb{I}$), the trained and initial embedding matrices of the transformer, and random Gaussian $W$ with $d\,{=}\,768$ (mean and spread over $8$ draws). Symmetric-log axes; dotted lines as in Fig.~\ref{fig:token-trace}.}
    \label{fig:token-v-emb}
\end{figure}

\paragraph{Embedding-level autocorrelations.}
The embedding layer maps $e_t$ to the $d$-dimensional activation $\phi^{\scriptscriptstyle(0)}_t =W e_t + p_t$, resulting in the (non-standardised) cross-covariance matrix
\begin{align}
C_0(\tau) = \mathbb{E}\big[(\phi^{\scriptscriptstyle(0)}_{T-\tau}-\mathbb{E}[\phi^{\scriptscriptstyle(0)}_{T-\tau}])(\phi^{\scriptscriptstyle(0)}_{T}-\mathbb{E}[\phi^{\scriptscriptstyle(0)}_{T}])^\top\big] = W C_{\mathrm{tok}}(\tau) W^\top.
\end{align}
Its trace is $\mathrm{Tr}(W C_{\mathrm{tok}} W^\top)\,{=}\,\mathrm{Tr}(C_{\mathrm{tok}}\,G)$, with $G\,{=}\,W^\top W$.
Splitting $G\,{=}\,g\,\mathbb{I} + (G-g\,\mathbb{I})$ with $g\,{=}\,\|W\|_F^2/v$ separates a multiple of the token trace from an anisotropic part, in which the overlap of the embedding vectors of two different tokens couples them across the two positions.
For $W$ with i.i.d.\ Gaussian entries of variance $\sigma^2$, $\mathbb{E}[\mathrm{Tr}(C_{\mathrm{tok}}G)]\,{=}\,\sigma^2 d\,\mathrm{Tr}\,C_{\mathrm{tok}}$ and the anisotropic part has standard deviation $\sigma^2\sqrt{d(\|C_{\mathrm{tok}}\|_F^2+\mathrm{Tr}\,C_{\mathrm{tok}}^2)}$, a fraction of order $\sqrt{2/d}\,\|C_{\mathrm{tok}}\|_F$ of $g$: smaller than the scale $\|C_{\mathrm{tok}}\|_F/\sqrt{v}$ by $\sqrt{2v/d}$.
Therefore, after standardisation, which divides Eq.~\ref{eq:autocorrelations} by $\sqrt{\mathrm{Tr}(\Sigma_{31}G)\,\mathrm{Tr}(\Sigma_{31-\tau}G)}$, the autocorrelation of a random embedding follows the token-level one within a fraction of the scale.
Fig.~\ref{fig:token-v-emb} shows that this holds for eight random draws, for the embedding at initialisation and also for the trained embedding, where no argument of this kind applies a priori: from lag $2$ on, the four normalised curves coincide on the logarithmic scale.
The one exception is lag $1$, where the token-level value is six times smaller than its typical scale, and the trained embedding flips its sign.

\subsection{Latent-level layer-$1$ autocorrelations}
\label{app:rhm-layer1}

\begin{figure}[h]
    \centering
    \includegraphics[width=\linewidth]{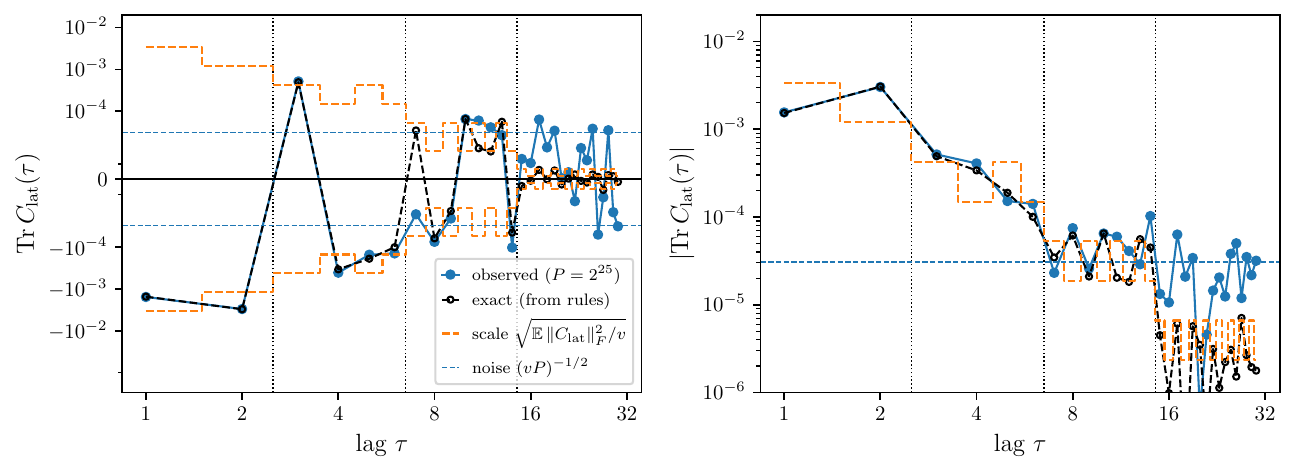}
    \caption{As Fig.~\ref{fig:token-trace}, for the idealised layer-$1$ sequence $Z_t$ (current token at odd $t$, parent of the last completed pair at even $t$): trace $\mathrm{Tr}\,C_{\rm lat}(\tau)$ between the one-hot encodings of $Z_{31}$ and $Z_{31-\tau}$, measured on $2^{25}$ sampled trees in blue, exact from the rules in black dashed, with the scale $\pm\sqrt{\mathbb{E}\,\|C_{\rm lat}\|_F^2/v}$ of Eq.~\ref{eq:trace-scale} in orange dashed and the sampling noise as a blue dashed line.}
    \label{fig:latent1-trace}
\end{figure}

We can repeat the analysis of Sec.~\ref{app:rhm-embeddings} for layer-1 activations, using the idealised construction of Fig.~\ref{fig:main}d as ground truth for the latent content of activations.
In particular, the sketch of Fig.~\ref{fig:main}d assigns to layer $1$ the following sequence of latent variables
\begin{align}
    Z_t = \begin{cases} X_t & t \text{ odd},\\ X^{(4)}_{t/2} & t \text{ even},\end{cases}\qquad t=1,\dots,31,
\end{align}
i.e.\ $X_1, X^{(4)}_1, X_3, X^{(4)}_2,\dots, X^{(4)}_{15}, X_{31}$: the current token at odd positions and the parent of the pair just completed at even positions.
We call the covariance matrix of the one-hot encodings of this latent sequence $C_{\mathrm{lat}}(\tau)$.
Since tokens and parents take values in the same set of $v$ symbols, the exact covariances can be computed via Eq.~\ref{eq:two-leaf-joint}, having a shorter path on the parent side.
However, the matching of labels is arbitrary, as relabelling the parent symbols by a random permutation is a symmetry of the RHM.
Therefore, the absolute value of the trace at odd lags has no intrinsic meaning.
The scale, instead, is given by Eq.~\ref{eq:trace-scale}, with $h'\,{=}\,h-1$ at odd lags, where the target is a parent, while $h'\,{=}\,h-1$ at even lags, where the target is a token.
As a result, the scale curve coincides with the token-level one at even lags; and is $\sqrt{m}$ times larger at odd lags.
Fig.~\ref{fig:latent1-trace} shows the exact curve, the trace measured on $2^{25}$ sampled trees, which agrees with it within $2$ noise units, and the expected scale.

\begin{figure}[h]
    \centering
    \includegraphics[width=\linewidth]{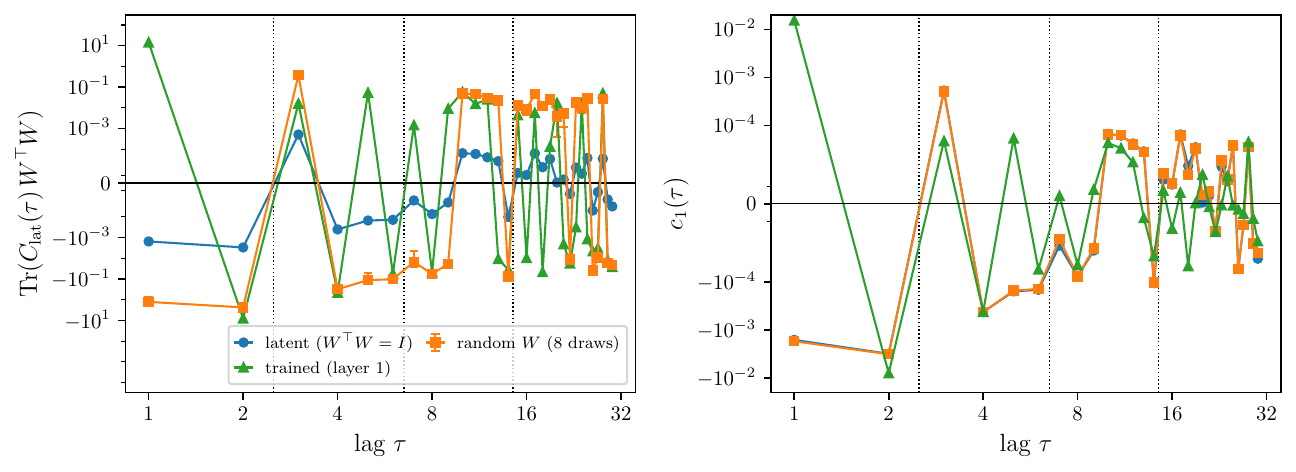}
    \caption{Same as Fig.~\ref{fig:token-v-emb}, one layer up: trace of the idealised layer-$1$ sequence pushed through an embedding, $\mathrm{Tr}(C_{\rm lat}\,W^\top W)$ (left), and the normalised autocorrelation $c_1(\tau)$ (right), for the one-hot sequence ($W^\top W\,{=}\,\mathbb{I}$), random Gaussian $W$, and the measured layer $1$ of the trained transformer (for which the left panel shows the trace of the cross-covariance of the activations).}
    \label{fig:latent1-v-layer1}
\end{figure}

\begin{figure}[h]
    \centering
    \includegraphics[width=\linewidth]{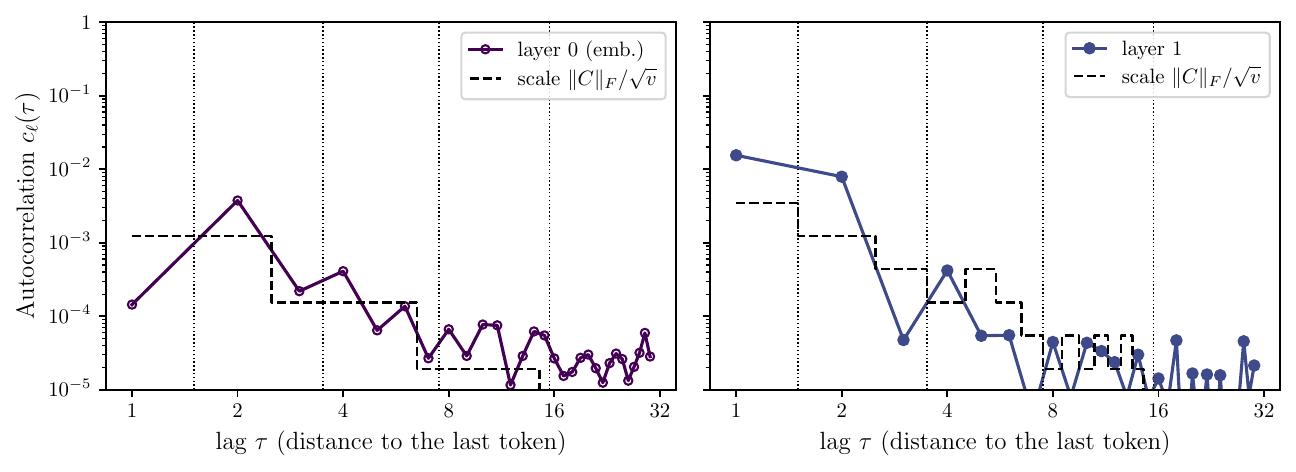}
    \caption{Measured autocorrelations $|c_0(\tau)|$ (left) and $|c_1(\tau)|$ (right) of the trained transformer, as in Fig.~\ref{fig:main}a, against the scale $\sqrt{\mathbb{E}\,\|C\|_F^2/v}$ of Eq.~\ref{eq:trace-scale} (dashed steps), for the token sequence (left) and for the idealised layer-$1$ sequence (right), normalised by the one-hot variance $1-1/v$. Dotted lines at lags $1.5$, $3.5$, $7.5$ and $15.5$ as in Fig.~\ref{fig:main}a; the steps of the scale sit instead at the levels of the lowest common ancestor of the two positions ($2.5$, $6.5$, $14.5$ for tokens).}
    \label{fig:layers01-scale}
\end{figure}

\paragraph{Layer-$1$ autocorrelations.}
Fig.~\ref{fig:latent1-v-layer1} compares the measured autocorrelation $c_1(\tau)$ of layer $1$ (in green) with those of the idealised latent sequence $Z_t$ (blue) and its random embeddings ($d\,{=}\,768$-by-$v$ random $W$ with i.i.d. Gaussian entries, orange).
Random embeddings follow the one-hot curve as for the token- and embedding-level autocorrelations of Sec.~\ref{app:rhm-embeddings}, with standardisation removing the extra scale factor.
For the trained curve, the sign does not always match, but this is expected due to the symmetry for relabelling of the parent symbols.
Indeed, as shown in Fig.~\ref{fig:layers01-scale}, the absolute value of $c_1(\tau)$ (right panel) lies within a small factor of the scale of the corresponding one-hot sequence, as
does $c_0(\tau)$ (left panel).
Exceptions can be understood within the departures of the layer-$1$ encodings from the sketch of Fig.~\ref{fig:main}d (cf. Appendix~\ref{app:probing}).
For instance, the activation in position $31$ of layer $1$ retains a small copy of the previous parent $X^{\scriptscriptstyle(4)}_{15}$ ($3.7\%$ of its variance), which is partially aligned with the code of $X^{\scriptscriptstyle(4)}_{15}$ at position $30$.
This is a same-node term, producing stronger correlations than a cross-node term.
Hence, the resulting contribution competes with the $X^{\scriptscriptstyle(4)}_{15}-X_{31}$, despite the attenuation due to the low fraction of explained variance.

\section{Linear decomposition of the activations into latent codes}
\label{app:probing}

This appendix details the encoding regression of Eq.~\ref{eq:linear-representation-hypothesis}, its relation to linear probes, and the encoding maps of the five layers that the item list of Sec.~\ref{sec:rhm} summarises. Throughout, $\phi\,{\coloneq}\,\phi^{(\ell)}_t$ is the activation of a fixed layer and position, $\Sigma_\phi\,{:=}\,\mathrm{Cov}(\phi)$ its covariance, and every node $Z$ of the tree, latent or visible, is represented by its centered one-hot encoding $e_Z\in\mathbb{R}^v$, $(e_Z)_a\,{=}\,\delta_{Z,a}-\mathbb{Z}(X\,{=}\,a)$, with covariance $\Sigma_Z$; $\mathbf{Z}\in\mathbb{R}^{63v}$ stacks the $63$ encodings.

\paragraph{The joint regression.}
A linear function of the one-hot encoding of a discrete variable $Z$ assigns a distinct arbitrary number to all possible values of $Z$.
This is an arbitrary function of the symbol $Z$.
Therefore, the linear functions of $\mathbf{Z}$ are exactly the sums $\sum_Z g_Z(Z)$ of single-node functions.
Morevoer, the least-squares solution of Eq.~\ref{eq:linear-representation-hypothesis}, $A\,{=}\,\mathrm{Cov}(\phi,e_\mathbf{Z})\,\mathrm{Cov}(e_\mathbf{Z})^{+}$, is the best approximation of the activation by such a sum.
The residual $\epsilon$, uncorrelated with every node by construction, collects what no sum of single-node functions can capture, e.g. functions of several nodes at once.
The only linear dependencies among the encodings are the sum-to-zero constraints, one per node, because no node is a sum of single-node functions of the others: a parent is determined by the pair of its children, but not by a function of one child plus a function of the other.
$\mathrm{Cov}(e_\mathbf{Z})$ has therefore rank exactly $63(v-1)$.

\paragraph{Partialling out.}
The coefficients of a joint regression obey the Frisch--Waugh--Lovell theorem~\citep{frisch1933partial,lovell1963seasonal}: the block $A_Z$ multiplying $\mathbf{e}_Z$ equals the regression of $\phi$ on $\tilde{e}_Z$, the residual of $e_Z$ after its own regression on the encodings of all the other nodes.
Importantly, this property \textbf{distinguishes Eq.~\ref{eq:linear-representation-hypothesis} from a linear probe}.
Consider, for instance,  an activation that encodes only the token $X_1$, $\phi\,{=}\,g(X_1)$.
$X_1$ and its parent $X^{\scriptscriptstyle(4)}_1$ are correlated through the rules, so $\phi$ is correlated with $e_{\scriptscriptstyle X^{\scriptscriptstyle(4)}_1}$, and a linear readout of the parent from $\phi$ will give a positive signal.
In the joint regression, instead, the initial regression of $e_{\scriptscriptstyle X^{\scriptscriptstyle(4)}_1}$ on all the other nodes, including $X_1$, leaves a residual orthogonal to $\phi$.
Thus, $A_{\scriptscriptstyle X^{\scriptscriptstyle(4)}_1}\,{=}\,0$ exactly.
In simple terms, $A_Z\,{\neq}\,0$ requires a component of the activation that correlates with the value of $Z$ beyond what any single other node implies.

\paragraph{Latent codes.}
The code of $Z$ is the block $A_Z$ of $v$ vectors in $\mathbb{R}^d$, one per value of $Z$, and its strength $S_Z\,{=}\,\mathrm{Tr}(A_Z\Sigma_Z A_Z^\top)/\mathrm{Tr}\,\Sigma_\phi$ is the fraction of the total activation variance explained by the variance of $A_Z e_Z$.
The total explained fraction $\mathrm{Tr}(A\,\mathrm{Cov}(e_\mathbf{Z})A^\top)/\mathrm{Tr}\,\Sigma_\phi$ is the variance of the sum of all codes, and its complement is the fraction of variance explained by the residual.

\paragraph{Relation to linear probes.}
A linear probe for $Z$ is the regression in the opposite direction, from $\phi$ to $e_Z$.
The least-squares solution is $W_Z\,{=}\,C_Z^\top\Sigma_\phi^{+}$, with $C_Z\,{:=}\,\mathbb{E}[\phi\,e_Z^\top]$.
The quality of the probe is measured as the fraction of the variance of $Z$ that it recovers,
\begin{align}\label{eq:probe-R2}
    R^2_Z = \frac{\mathrm{Tr}\left(C_Z^\top \Sigma_\phi^{+} C_Z\right)}{\mathrm{Tr}\,\Sigma_Z}.
\end{align}
This is the infinite-data, unregularised limit of the probes trained by gradient descent in the works of Sec.~\ref{sec:related-work}.
Tree correlations between the latent variables inflate $R^2_Z$: a parent, for instance, reads at $11$--$23\%$ wherever its children are encoded.
$S_Z$ measures how much of the activation is a code of $Z$, without preventing readability.
For instance, a code of a few per cent can be read perfectly if it occupies directions of its own, as happens for the depth-$3$ variable $X^{\scriptscriptstyle(2)}_1$ in layer $3$: it can be read perfectly ($R^2\,{\simeq}\,1$) while only explaining $\lesssim 10 \%$ of the activation variance.
There is one more caveat with probes.
At positions $1$ and $2$, there are only $v\,{=}\,32$ and $mv\,{=}\,256$ possible prefixes, respectively.
Both are smaller than the embedding dimension $d$, so any injective map, including a randomly initialised layer, makes every function of the prefix linearly readable.
Even at later positions, a random layer is a random-feature map of the prefix, and a probe can still recover parts of any variable without implying a particular structure in the activations.
For these reasons we report encoding strengths throughout and use $R^2$ only for cross-checks such as the one above.

\paragraph{Statistical controls.}
Regression is performed using the moments of $2^{18}$ sentences and evaluated against those of $2^{16}$, both sets drawn independently from the training set and the set used for measuring autocorrelations.
A node whose code is pure noise picks up a fraction $(v-1)/P$ of the residual variance, about $10^{-4}$ at $P\,{=}\,2^{18}$.
This spurious part scales as $1/P$, whereas the true code does not.
Every strength we report is required to change by less than a factor $2$ between $P\,{=}\,2^{16}$ and $2^{18}$, where pure noise changes by a factor $4$.

\begin{figure}[p]
    \centering
    \includegraphics[width=\linewidth]{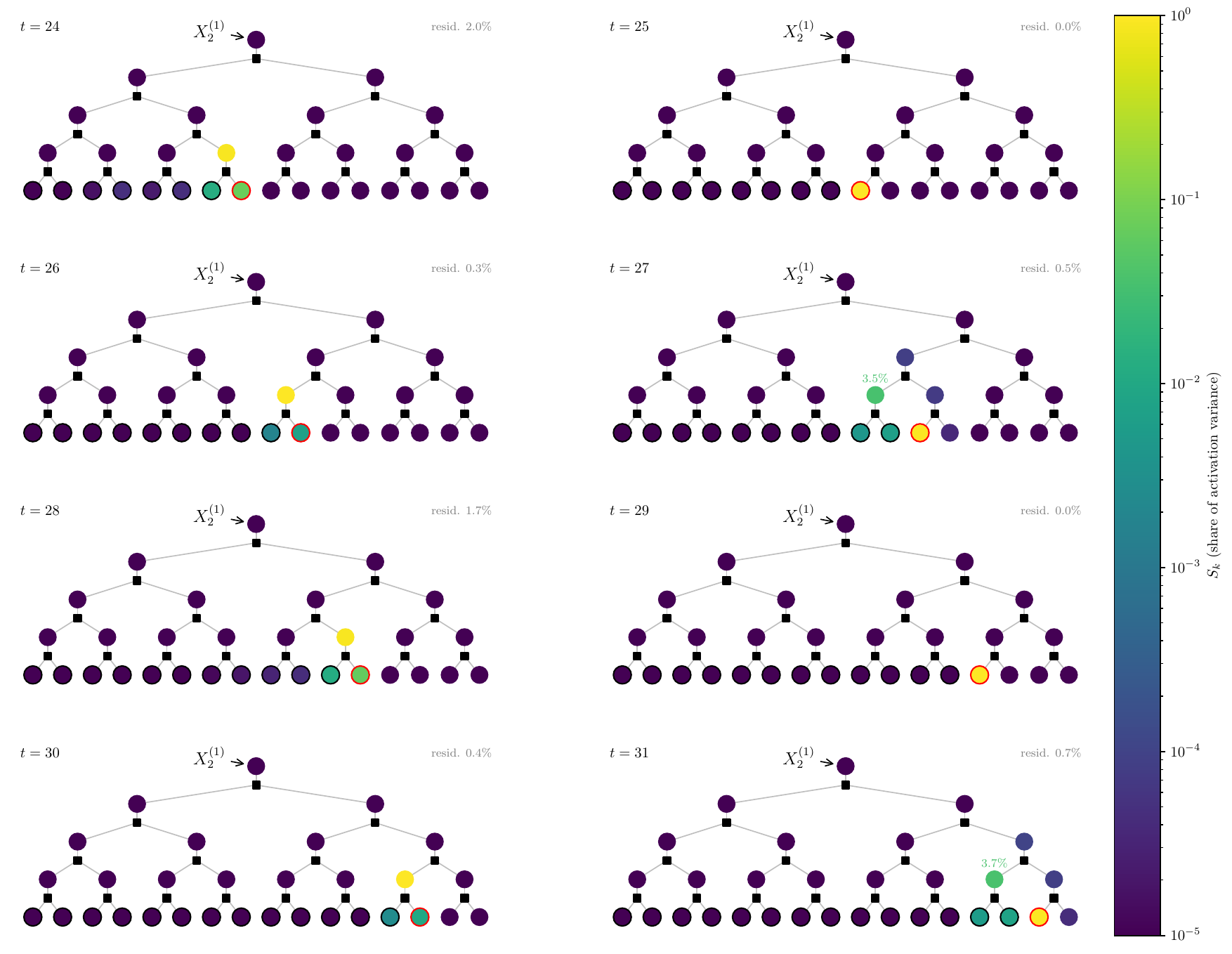}
    \caption{
        \textbf{Codes of layer $1$ on the data tree}, positions $t\,{=}\,24$--$31$ (one panel each, $t$ in the top-left corner).
        Each panel shows the depth-$4$ subtree of $X^{\scriptscriptstyle(1)}_2$ (root in the figures), whose leaves are the last $16$ tokens $X_{17},\dots,X_{32}$.
        Squares mark the production rules.
        The colour of each node is the strength $S_Z$ of its code in the activation at position $t$ (log scale, colourbar on the right), with printed percentages for selected nodes.
        The share of the non-additive residual is in the top-right corner.
        The red ring marks the current token $X_t$, while the tokens already read are surrounded by black rings.
        Layer $1$ holds one dominant code per position, alternating with the parity of $t$: odd positions encode the current token; even positions encode the pair they complete.
        Even $t$'s also hold a small code of the children, and $t\equiv3\pmod{4}$ a small code of the previous parent (see $t\,{=}\,27,31$).
        No variable deeper than a pair is encoded, and the residual's share is at most $2\%$.
    }
    \label{fig:rhm-tree-encodings-layer1}
\end{figure}

\begin{figure}[p]
    \centering
    \includegraphics[width=\linewidth]{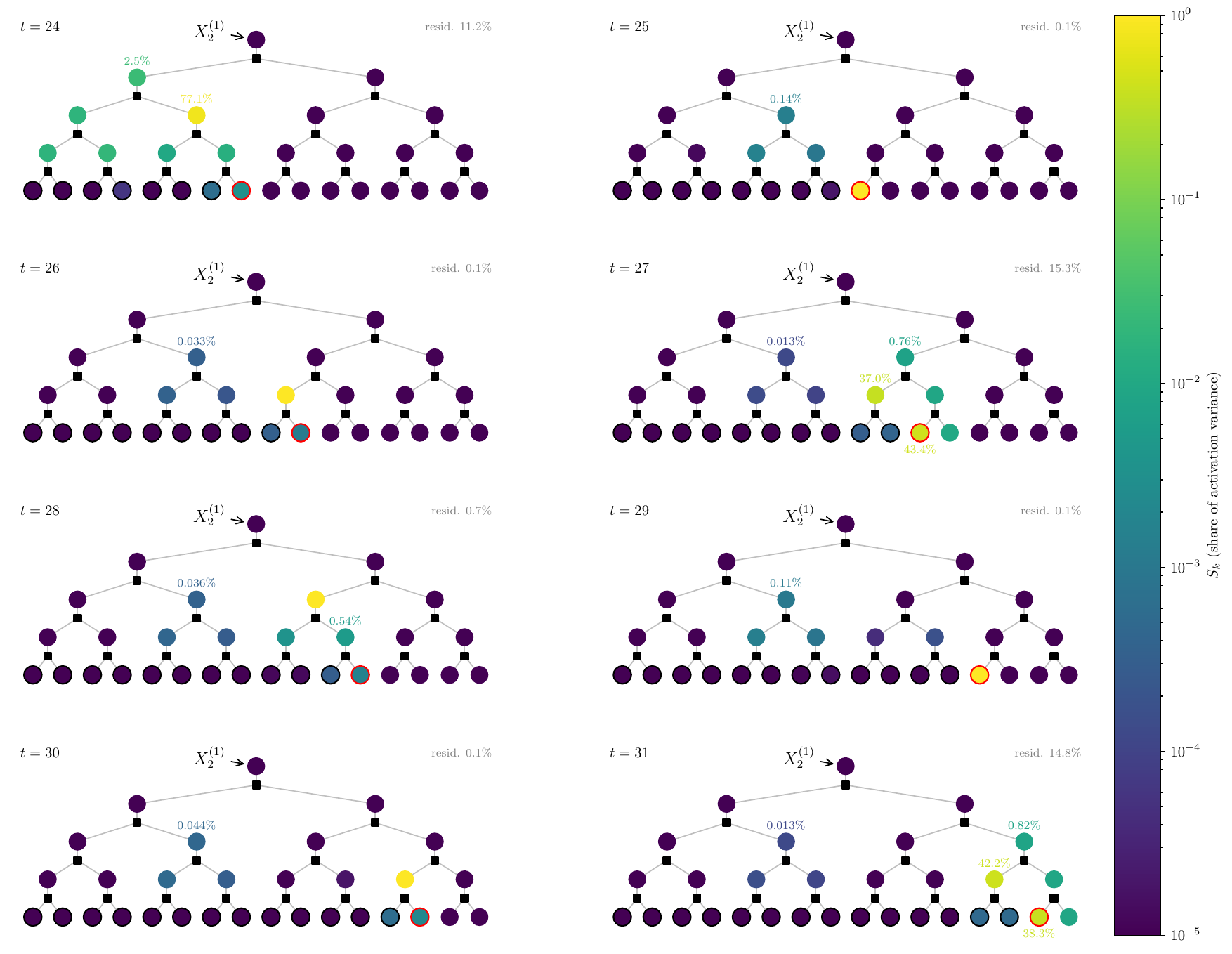}
    \caption{
        \textbf{Codes of layer $2$ on the data tree}, as in Fig.~\ref{fig:rhm-tree-encodings-layer1}.
        Layer $2$ adds the depth-$2$ variables at the positions that complete them (with $t$ a multiple of $4$, the span length of depth-$2$ subtrees).
        The code of a pair now persists over the span of its parent, as at $t\,{=}\,26,27$ and $t\,{=}\,30,31$ (code formed at $t\equiv 2\pmod{4}$, then kept at the next position next to the current token).
    }
    \label{fig:rhm-tree-encodings-layer2}
\end{figure}

\begin{figure}[p]
    \centering
    \includegraphics[width=\linewidth]{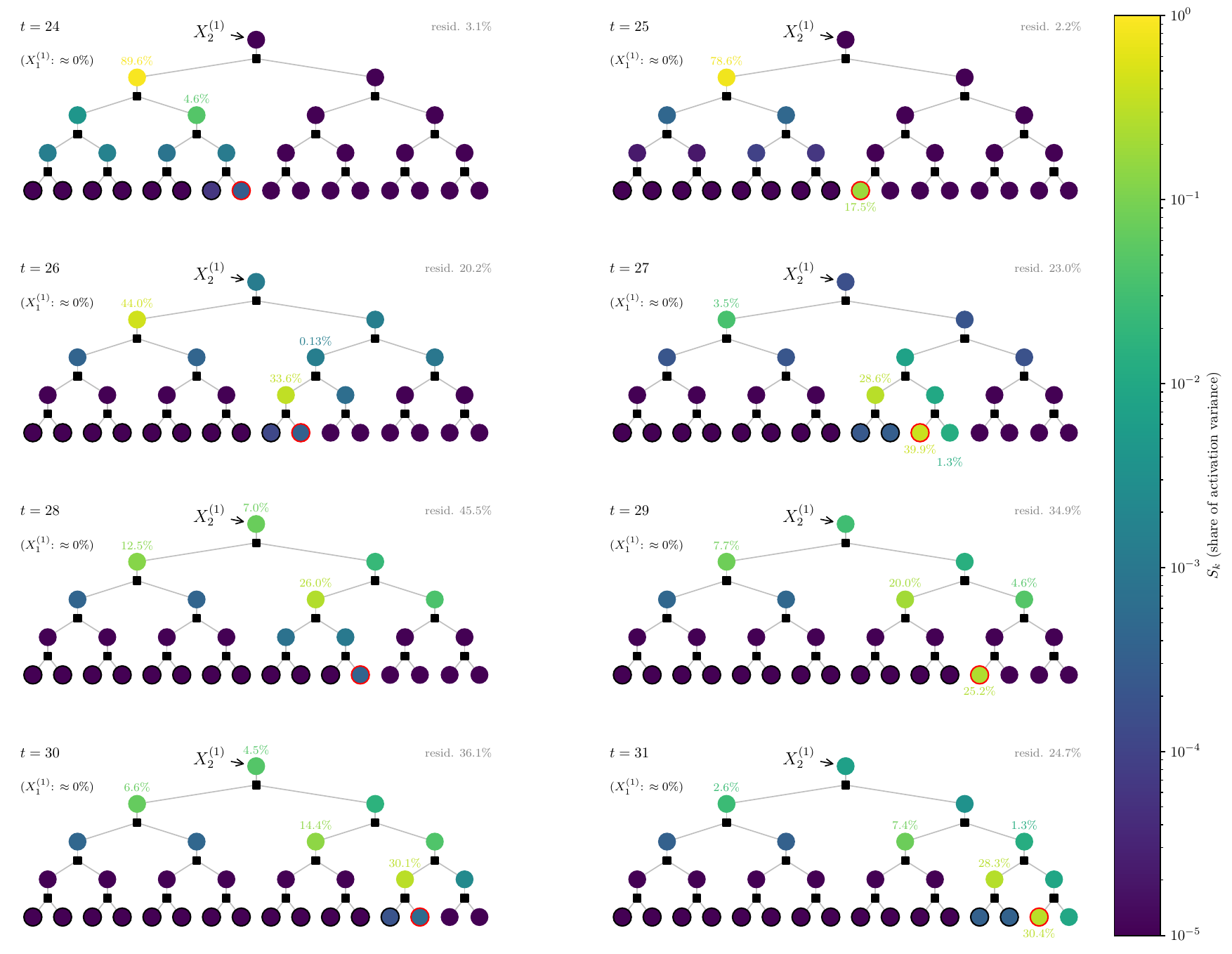}
    \caption{
        \textbf{Codes of layer $3$ on the data tree}, as in Fig.~\ref{fig:rhm-tree-encodings-layer1}; the top-left corner also reports the strength of $X^{\scriptscriptstyle(1)}_1$, the depth-$4$ variable outside the subtree.
        The depth-$3$ variable $X^{\scriptscriptstyle(2)}_3$ is formed at $t\,{=}\,24$ ($90\%$) and persists, fading, over the rest of the span of its parent ($79\%$ at $t\,{=}\,25$ down to $2.6\%$ at $t\,{=}\,31$), while the local pattern of layer $2$ repeats beneath it (pairs at $t\,{=}\,26,30$, the 4-span $X^{\scriptscriptstyle(3)}_7$ at $t\,{=}\,28$).
        From $t\,{=}\,28$ the still-open depth-$4$ variable $X^{\scriptscriptstyle(1)}_2$ receives a partial code ($7\%$ at the root of the subtree), whereas the completed $X^{\scriptscriptstyle(1)}_1$ is not encoded at any of these positions.
        The residual is small where one code dominates ($2$--$3\%$ at $t\,{=}\,24,25$) and reaches $20$--$46\%$ elsewhere.
        At $t\,{=}\,31$, the three variables whose spans contain both $X_{31}$ and the earlier positions ($X^{\scriptscriptstyle(4)}_{15}$, $X^{\scriptscriptstyle(3)}_7$, $X^{\scriptscriptstyle(2)}_3$ at $28$, $7.4$ and $2.6\%$) are the ones that carry the plateaus of the autocorrelation (Fig.~\ref{fig:rhm-trace-shares}).
    }
    \label{fig:rhm-tree-encodings-layer3}
\end{figure}

\paragraph{Encoding maps.}
Tables~\ref{tab:enc-layers12}--\ref{tab:enc-layer5} list the strengths of all codes at every position of the five layers (only positions $1$--$16$ for layers $1$ and $2$, where the second half repeats the first).
To keep the tables readable, the codes of the tokens and of the depth-$1$ and depth-$2$ variables, which come and go with the position, are listed by their role relative to the position (current token, last completed pair, and so on).
The tables of layers $1$, $2$ and $3$ are accompanied by figures, Figs.~\ref{fig:rhm-tree-encodings-layer1}, ~\ref{fig:rhm-tree-encodings-layer2}, ~\ref{fig:rhm-tree-encodings-layer3}, illustrating the distribution of latent codes on the data tree.
For the sake of clarity, we focus on the rightmost depth-$4$ subtree, descending from the second level-$1$/depth-$4$ variable $X^{\scriptscriptstyle(1)}_2$ (highlighted at the root of each subtree in the figures) and generating the last $2^{4}\,{=}\,16$ tokens of each sentence (the leaves of each subtree in the figures).
Each figure consists of $8$ subtrees, reporting the fractions of explained variance of the activations of layers $1$, $2$ and $3$ in the last $8$ positions, $t\,{=}\,24$ to $31$ (indicated in the top-left corner of each sub-figure). The fraction of variance explained by each latent variable is reported as the colour of the corresponding node (colourbar on the right). The fraction of the nonlinear residual is reported in the top-right corner of each sub-figure.
Together, tables and figures give the numbers behind the item list of Sec.~\ref{sec:rhm}.
\begin{table}[tbp]
\centering
\footnotesize
\setlength{\tabcolsep}{2.4pt}
\renewcommand{\arraystretch}{0.98}
\begin{tabular}{lrrrrrrrrrrrrrrrr}
\toprule
\multicolumn{17}{l}{\textbf{Layer 1, positions 1--16}} \\
\midrule
\multicolumn{1}{l}{$t$} & 1 & 2 & 3 & 4 & 5 & 6 & 7 & 8 & 9 & 10 & 11 & 12 & 13 & 14 & 15 & 16 \\
\multicolumn{1}{l}{\emph{closes depth}} & -- & 1 & -- & 2 & -- & 1 & -- & 3 & -- & 1 & -- & 2 & -- & 1 & -- & 4 \\
\midrule
$X_t$ & \textbf{100} & 0.5 & \textbf{91.5} & 2.3 & \textbf{100} & 0.6 & \textbf{91.6} & 5.3 & \textbf{100} & 0.6 & \textbf{93.3} & 6.1 & \textbf{100} & 0.9 & \textbf{86.9} & 3.1 \\
other $X_i$ &  & 0.2 & 1.4 & 1.2 &  & 0.2 & 1.5 & 1.0 &  & 0.2 & 1.3 & 1.1 &  & 0.2 & 2.4 & 1.2 \\
$X^{(4)}_{p}$ &  & \textbf{99.1} & 5.1 & \textbf{95.1} &  & \textbf{98.8} & 4.9 & \textbf{91.8} &  & \textbf{98.7} & 3.8 & \textbf{90.5} &  & \textbf{98.3} & 7.4 & \textbf{94.6} \\
\midrule
sum of codes & 100 & 99.8 & 98.0 & 98.7 & 100 & 99.6 & 98.0 & 98.2 & 100 & 99.5 & 98.4 & 97.7 & 100 & 99.5 & 96.8 & 99.0 \\
explained & 100 & 99.7 & 99.2 & 98.8 & 100 & 99.7 & 99.3 & 98.5 & 100 & 99.7 & 99.4 & 98.5 & 100 & 99.7 & 98.4 & 98.7 \\
residual &  & 0.3 & 0.8 & 1.2 &  & 0.3 & 0.7 & 1.5 &  & 0.3 & 0.6 & 1.5 &  & 0.3 & 1.6 & 1.3 \\
\midrule
\multicolumn{17}{l}{\textbf{Layer 2, positions 1--16}} \\
\midrule
\multicolumn{1}{l}{$t$} & 1 & 2 & 3 & 4 & 5 & 6 & 7 & 8 & 9 & 10 & 11 & 12 & 13 & 14 & 15 & 16 \\
\multicolumn{1}{l}{\emph{closes depth}} & -- & 1 & -- & 2 & -- & 1 & -- & 3 & -- & 1 & -- & 2 & -- & 1 & -- & 4 \\
\midrule
$X_t$ & \textbf{100} & 0.1 & \textbf{34.3} &  & \textbf{100} & 0.1 & \textbf{39.8} & 0.2 & \textbf{99.4} & 0.1 & \textbf{38.2} & 0.1 & \textbf{99.3} & 0.2 & \textbf{29.1} & 0.5 \\
$X_{t+1}$ &  &  & 0.8 &  &  &  & 0.8 &  &  &  & 0.9 &  &  &  & 0.9 &  \\
other $X_i$ &  &  & 0.2 &  &  & 0.1 & 0.2 & 0.1 &  &  & 0.2 &  &  & 0.1 & 0.4 & 0.3 \\
$X^{(4)}_{p}$ &  & \textbf{99.8} & \textbf{48.6} & 0.4 &  & \textbf{99.7} & \textbf{42.9} & 1.1 & 0.1 & \textbf{99.6} & \textbf{42.9} & 0.3 &  & \textbf{99.4} & \textbf{43.6} & 12.7 \\
$X^{(4)}_{p-1}$ &  &  &  & 0.3 &  &  &  & 0.7 & 0.1 &  &  & 0.2 &  &  &  & 5.4 \\
$X^{(4)}_{p+1}$ &  &  & 0.8 &  &  &  & 0.8 &  &  &  & 0.9 &  &  &  & 1.2 &  \\
other $X^{(4)}_i$ &  &  &  &  &  &  &  & 3.3 &  &  & 0.1 &  & 0.3 & 0.1 & 0.1 &  \\
$X^{(3)}_{q}$ &  &  &  & \textbf{99.1} &  &  &  & \textbf{78.5} & 0.2 &  &  & \textbf{98.7} &  &  &  & \textbf{70.3} \\
$X^{(3)}_{q-1}$ &  &  &  &  &  &  &  & 1.8 &  &  &  &  & 0.2 &  & 0.1 &  \\
$X^{(3)}_{q+1}$ &  &  & 0.7 &  &  &  & 0.7 &  &  &  & 0.7 &  &  &  & 1.9 &  \\
$X^{(2)}_1$ [1--8] &  &  &  &  &  &  &  & 2.6 &  &  &  &  &  &  &  &  \\
\midrule
sum of codes & 100 & 99.9 & 85.5 & 99.8 & 100 & 99.9 & 85.3 & 88.3 & 99.8 & 99.8 & 84.0 & 99.4 & 99.8 & 99.8 & 77.4 & 89.3 \\
explained & 100 & 99.9 & 86.7 & 99.4 & 100 & 99.9 & 86.7 & 89.0 & 99.9 & 99.9 & 85.4 & 99.5 & 99.9 & 99.9 & 80.5 & 87.8 \\
residual &  & 0.1 & 13.3 & 0.6 &  & 0.1 & 13.3 & 11.0 & 0.1 & 0.1 & 14.6 & 0.5 & 0.1 & 0.1 & 19.5 & 12.2 \\
\bottomrule
\end{tabular}
\caption{
    \textbf{Encoding maps of layers 1 and 2}, positions $1$--$16$ (positions $17$--$31$ repeat the pattern).
    Cells: percentage of the activation variance occupied by the code of each node at position $t$  ($P\,{=}\,2^{18}$), blank below $0.05\%$, bold at or above $20\%$.
    The row \emph{closes depth} gives the depth of the deepest variable whose span ends at $t$.
    Tokens and depth-$1$ and depth-$2$ variables are listed by their role relative to $t$, with $p\,{=}\,\lfloor t/2\rfloor$ and $q\,{=}\,\lfloor t/4\rfloor$ the numbers of completed pairs and 4-spans:
    $X_t$ is the current token and $X_{t+1}$ the next; $X^{\scriptscriptstyle(4)}_{p}$ is the last completed pair, $X^{\scriptscriptstyle(4)}_{p-1}$ the one before it and $X^{\scriptscriptstyle(4)}_{p+1}$ the current open pair for odd $t$, or the following one for even $t$.
    Likewise for the 4-spans $X^{\scriptscriptstyle(3)}_{q}$, $X^{\scriptscriptstyle(3)}_{q-1}$, $X^{\scriptscriptstyle(3)}_{q+1}$.
    The \emph{other} rows sum the remaining nodes of that depth. Deeper variables are listed by name, with their span in brackets.
    \emph{Sum of codes} adds the strengths of all $63$ nodes, \emph{explained} is the variance of the sum of the codes, and \emph{residual} its complement, i.e. the part of the activation that is not a sum of single-node functions.
    \textbf{Layer $1$} holds essentially one code per position, with odd positions encoding the current token ($100\%$ at positions $t\equiv 1\pmod{4}$ and $87$--$94\%$ at positions $t\equiv 3\pmod{4}$, where the variable of the pair just completed is retained at $3.5$--$7.4\%$), and even positions the depth-$1$ parent of the completed pair ($97.5$--$99.1\%$ at positions $t\equiv 2\pmod{4}$, and $89$--$95\%$ at the multiples of $4$, where the current token takes $2$--$7\%$).
    \textbf{Layer $2$} adds depth-$2$ variables at the multiples of $4$: at $98$--$99\%$ at positions $4$, $12$, $20$ and $28$, and at $70$--$79\%$ at positions $8$, $16$ and $24$, where all latent variables of the completed $8$-span linger (c.f. $t\,{=}\,24$ tree of Fig.~\ref{fig:rhm-tree-encodings-layer2}).
    Note positions $t\equiv 3\pmod{4}$, where the depth-$1$ variable of the last completed pair is kept at $37$--$49\%$ next to the current token at $29$--$43\%$.
    This is \emph{persistence}: the code of a pair survives the position after the pair closes, and is dropped at the next one, when the span of its parent closes.
    }
\label{tab:enc-layers12}
\end{table}


\begin{table}[p]
\centering
\footnotesize
\setlength{\tabcolsep}{2.4pt}
\renewcommand{\arraystretch}{0.98}
\begin{tabular}{lrrrrrrrrrrrrrrrr}
\toprule
\multicolumn{17}{l}{\textbf{Layer 3, positions 1--16}} \\
\midrule
\multicolumn{1}{l}{$t$} & 1 & 2 & 3 & 4 & 5 & 6 & 7 & 8 & 9 & 10 & 11 & 12 & 13 & 14 & 15 & 16 \\
\multicolumn{1}{l}{\emph{closes depth}} & -- & 1 & -- & 2 & -- & 1 & -- & 3 & -- & 1 & -- & 2 & -- & 1 & -- & 4 \\
\midrule
$X_t$ & \textbf{100} & 0.1 & \textbf{33.7} &  & \textbf{27.0} &  & \textbf{28.4} & 0.1 & \textbf{26.6} &  & \textbf{32.3} &  & 19.9 &  & 11.4 & 0.2 \\
$X_{t+1}$ &  &  & 1.5 &  &  &  & 1.0 &  &  &  & 1.3 &  &  &  & 0.3 &  \\
other $X_i$ &  &  & 0.2 &  &  & 0.1 & 0.2 & 0.1 &  & 0.1 & 0.2 & 0.2 & 0.2 & 0.2 & 0.3 & 0.4 \\
$X^{(4)}_{p}$ &  & \textbf{99.8} & \textbf{41.1} & 0.3 &  & \textbf{39.5} & \textbf{24.9} & 0.3 &  & \textbf{41.9} & \textbf{30.2} & 0.1 &  & \textbf{23.6} & 12.5 & 6.3 \\
$X^{(4)}_{p-1}$ &  &  &  & 0.2 &  &  &  & 0.2 &  &  &  & 0.1 &  &  &  & 3.9 \\
$X^{(4)}_{p+1}$ &  &  & 1.1 &  &  & 0.7 & 1.1 &  &  & 0.1 & 0.9 &  &  & 0.2 & 0.6 &  \\
other $X^{(4)}_i$ &  &  & 0.1 &  &  & 0.1 & 0.1 & 0.6 &  &  & 0.1 & 0.1 & 0.1 & 0.1 & 0.1 & 0.1 \\
$X^{(3)}_{q}$ &  &  &  & \textbf{99.2} & \textbf{70.2} & \textbf{36.0} & 16.3 & 9.9 & 0.1 &  &  & \textbf{30.0} & 19.8 & 11.0 & 9.9 & \textbf{33.7} \\
$X^{(3)}_{q-1}$ &  &  &  &  &  &  &  & 0.8 & 0.1 &  &  &  &  &  & 0.1 & \textbf{24.9} \\
$X^{(3)}_{q+1}$ &  &  & 0.8 &  &  & 1.3 & 0.9 &  &  & 0.1 & 0.7 & 2.9 & 5.3 & 3.1 & 2.7 &  \\
other $X^{(3)}_i$ &  &  &  &  &  &  &  &  &  & 0.1 & 0.1 & 0.1 & 0.1 & 0.1 & 0.1 &  \\
$X^{(2)}_1$ [1--8] &  &  &  &  &  & 1.0 & 0.3 & \textbf{82.5} & \textbf{71.4} & \textbf{42.0} & 6.5 & 14.1 & 8.5 & 7.1 & 9.4 &  \\
$X^{(2)}_2$ [9--16] &  &  &  &  &  &  &  &  &  & 0.1 &  & 2.2 & 1.7 & 4.0 & 3.4 & 10.9 \\
$X^{(1)}_1$ [1--16] &  &  &  &  &  &  &  &  &  & 0.1 &  & 8.1 & 3.7 & 12.4 & 10.3 &  \\
\midrule
sum of codes & 100 & 99.9 & 78.5 & 99.7 & 97.3 & 78.7 & 73.2 & 94.4 & 98.1 & 84.6 & 72.3 & 58.1 & 59.3 & 62.0 & 61.1 & 80.4 \\
explained & 100 & 99.9 & 80.0 & 99.5 & 97.5 & 79.5 & 74.4 & 95.0 & 98.2 & 84.9 & 73.7 & 59.9 & 61.7 & 61.9 & 62.1 & 72.5 \\
residual &  & 0.1 & 20.0 & 0.5 & 2.5 & 20.5 & 25.6 & 5.0 & 1.8 & 15.1 & 26.3 & 40.1 & 38.3 & 38.1 & 37.9 & 27.5 \\
\midrule
\multicolumn{17}{l}{\textbf{Layer 3, positions 17--31}} \\
\midrule
\multicolumn{1}{l}{$t$} & 17 & 18 & 19 & 20 & 21 & 22 & 23 & 24 & 25 & 26 & 27 & 28 & 29 & 30 & 31 &  \\
\multicolumn{1}{l}{\emph{closes depth}} & -- & 1 & -- & 2 & -- & 1 & -- & 3 & -- & 1 & -- & 2 & -- & 1 & -- &  \\
\midrule
$X_t$ & \textbf{92.5} & 0.1 & \textbf{37.1} &  & \textbf{27.3} &  & \textbf{32.1} &  & 17.5 &  & \textbf{39.9} &  & \textbf{25.2} & 0.1 & \textbf{30.4} &  \\
$X_{t+1}$ &  &  & 1.6 &  &  &  & 1.1 &  &  &  & 1.3 &  &  &  & 1.0 &  \\
other $X_i$ &  &  & 0.2 &  &  & 0.1 & 0.2 &  &  & 0.1 & 0.2 & 0.2 & 0.2 & 0.2 & 0.2 &  \\
$X^{(4)}_{p}$ & 1.1 & \textbf{97.3} & \textbf{36.9} & 0.2 &  & \textbf{34.8} & \textbf{25.9} & 0.1 &  & \textbf{33.6} & \textbf{28.6} & 0.1 &  & \textbf{30.1} & \textbf{28.3} &  \\
$X^{(4)}_{p-1}$ & 0.4 & 0.4 & 0.1 & 0.2 &  &  &  & 0.1 &  &  &  & 0.1 &  &  &  &  \\
$X^{(4)}_{p+1}$ &  &  & 1.1 &  &  & 0.8 & 1.1 &  &  & 0.1 & 1.0 &  &  & 0.2 & 0.9 &  \\
other $X^{(4)}_i$ &  & 0.2 & 0.1 & 0.3 & 0.3 & 0.3 & 0.1 & 0.7 & 0.5 & 0.4 & 0.1 & 0.4 & 0.3 & 0.3 & 0.1 &  \\
$X^{(3)}_{q}$ & 4.4 & 1.5 & 0.2 & \textbf{97.7} & \textbf{69.0} & \textbf{35.3} & 12.8 & 4.6 &  &  &  & \textbf{26.0} & \textbf{20.0} & 14.4 & 7.4 &  \\
$X^{(3)}_{q-1}$ &  &  &  & 1.0 & 0.8 & 0.7 & 0.1 & 0.4 &  &  &  &  &  &  &  &  \\
$X^{(3)}_{q+1}$ &  &  & 0.8 &  &  & 1.5 & 0.9 &  &  & 0.1 & 0.8 & 3.9 & 4.6 & 4.3 & 1.3 &  \\
other $X^{(3)}_i$ &  &  &  &  &  &  &  & 1.1 & 1.4 & 1.2 & 0.2 & 1.0 & 0.6 & 0.8 & 0.1 &  \\
$X^{(2)}_3$ [17--24] &  &  &  &  &  & 1.2 & 0.2 & \textbf{89.6} & \textbf{78.6} & \textbf{44.0} & 3.5 & 12.5 & 7.7 & 6.6 & 2.6 &  \\
$X^{(2)}_4$ [25--32] &  &  &  &  &  &  &  &  &  & 0.1 &  & 2.2 & 1.4 & 1.6 & 0.4 &  \\
$X^{(1)}_2$ [17--32] &  &  &  &  &  &  &  &  &  & 0.1 &  & 7.0 & 2.9 & 4.5 & 0.7 &  \\
\midrule
sum of codes & 98.4 & 99.4 & 78.0 & 99.4 & 97.4 & 74.9 & 74.5 & 96.7 & 98.1 & 79.9 & 75.5 & 53.6 & 63.1 & 63.2 & 73.3 &  \\
explained & 97.3 & 99.0 & 79.4 & 99.0 & 97.4 & 75.5 & 76.0 & 96.9 & 97.8 & 79.8 & 77.0 & 54.5 & 65.1 & 63.9 & 75.3 &  \\
residual & 2.7 & 1.0 & 20.6 & 1.0 & 2.6 & 24.5 & 24.0 & 3.1 & 2.2 & 20.2 & 23.0 & 45.5 & 34.9 & 36.1 & 24.7 &  \\
\bottomrule
\end{tabular}
\caption{
    \textbf{Encoding map of layer $3$}, all positions; conventions as in Table~\ref{tab:enc-layers12}, with the second half of the context shown in full and the depth-$3$ and depth-$4$ variables listed by name.
    Part of layer $3$ extends the pattern of layer $2$ by one level.
    The depth-$2$ variables now persist over the span of their parent: $X^{\scriptscriptstyle(3)}_1$, formed at $t\,{=}\,4$ ($99\%$), is kept at $70$, $36$ and $16\%$ for $t\,{=}\,5$--$7$ and dropped at $t\,{=}\,8$, where its parent closes.
    The depth-$3$ variables are formed at $t\,{=}\,8$ and $24$ ($83$ and $90\%$).
    In addition, layer $3$ starts doing part of the job of layer $4$.
    The depth-$3$ variables persist over the span of their parent: $X^{\scriptscriptstyle(2)}_1$ is kept at $71$ and $42\%$ for $t\,{=}\,9,10$ and at $7$--$14\%$ up to $t\,{=}\,15$, and dropped at $t\,{=}\,16$; likewise $X^{\scriptscriptstyle(2)}_3$, from $79\%$ at $t\,{=}\,25$ to $2.6\%$ at $t\,{=}\,31$.
    The depth-$4$ variables receive partial codes while their span is still open ($X^{\scriptscriptstyle(1)}_1$ at $4$--$12\%$ for $t\,{=}\,12$--$15$, $X^{\scriptscriptstyle(1)}_2$ at $1$--$7\%$ for $t\,{=}\,28$--$31$), but are not kept once complete: at $t\,{=}\,16$ the layer holds the children $X^{\scriptscriptstyle(3)}_3$ and $X^{\scriptscriptstyle(3)}_4$ ($25$ and $34\%$) rather than $X^{\scriptscriptstyle(1)}_1$.
    The residual grows with the number of coexisting codes: from at most $5\%$ where a single variable dominates, to up to ${\sim}45\%$ when variables of multiple depths are present.
    }
\label{tab:enc-layer3}
\end{table}

\begin{table}[p]
\centering
\footnotesize
\setlength{\tabcolsep}{2.4pt}
\renewcommand{\arraystretch}{0.98}
\begin{tabular}{lrrrrrrrrrrrrrrrr}
\toprule
\multicolumn{17}{l}{\textbf{Layer 4, positions 1--16}} \\
\midrule
\multicolumn{1}{l}{$t$} & 1 & 2 & 3 & 4 & 5 & 6 & 7 & 8 & 9 & 10 & 11 & 12 & 13 & 14 & 15 & 16 \\
\multicolumn{1}{l}{\emph{closes depth}} & -- & 1 & -- & 2 & -- & 1 & -- & 3 & -- & 1 & -- & 2 & -- & 1 & -- & 4 \\
\midrule
$X_t$ & \textbf{100} &  & \textbf{34.4} &  & \textbf{53.3} &  & \textbf{32.5} & 0.1 & \textbf{66.4} &  & \textbf{35.1} &  & \textbf{43.1} &  & \textbf{25.5} & 0.2 \\
$X_{t+1}$ &  &  & 5.7 &  & 0.2 &  & 4.7 &  &  &  & 5.0 &  & 1.0 &  & 1.6 &  \\
other $X_i$ &  &  & 0.2 &  & 0.1 & 0.2 & 0.2 & 0.1 &  & 0.2 & 0.2 & 0.2 & 0.2 & 0.2 & 0.3 & 0.4 \\
$X^{(4)}_{p}$ &  & \textbf{99.9} & \textbf{20.2} & 0.2 &  & \textbf{29.2} & 12.9 & 0.3 &  & \textbf{49.9} & 15.0 & 0.1 &  & \textbf{22.2} & 8.2 & 5.4 \\
$X^{(4)}_{p-1}$ &  &  &  & 0.2 &  &  &  & 0.2 &  &  &  &  &  &  &  & 3.2 \\
$X^{(4)}_{p+1}$ &  &  & 1.5 &  & 0.1 & 8.7 & 1.6 &  &  & 0.8 & 1.4 &  & 0.3 & 13.9 & 0.9 &  \\
other $X^{(4)}_i$ &  &  & 0.1 &  & 0.1 & 0.1 & 0.1 & 0.6 &  & 0.1 & 0.1 & 0.1 & 0.2 & 0.1 & 0.1 & 0.1 \\
$X^{(3)}_{q}$ &  &  &  & \textbf{99.3} & \textbf{28.8} & 13.9 & 7.9 & 9.6 &  &  &  & \textbf{24.1} & 6.2 & 4.5 & 5.5 & \textbf{29.1} \\
$X^{(3)}_{q-1}$ &  &  &  &  &  &  &  & 0.8 &  &  &  &  &  &  &  & \textbf{21.8} \\
$X^{(3)}_{q+1}$ &  &  & 1.0 &  & 0.1 & 2.0 & 0.9 &  &  & 0.3 & 0.8 & 11.0 & 4.2 & 4.8 & 2.2 &  \\
other $X^{(3)}_i$ &  &  &  &  &  & 0.1 &  &  &  & 0.2 & 0.1 & 0.1 &  & 0.1 & 0.1 &  \\
$X^{(2)}_1$ [1--8] &  &  &  &  & 0.1 & 1.2 & 0.2 & \textbf{82.9} & \textbf{26.0} & 15.1 & 3.0 & 8.7 & 2.4 & 2.4 & 4.9 & 0.3 \\
$X^{(2)}_2$ [9--16] &  &  &  &  &  &  &  &  &  & 0.2 &  & 2.0 & 0.7 & 1.7 & 2.0 & 9.7 \\
$X^{(1)}_1$ [1--16] &  &  &  &  &  &  &  &  &  & 0.1 &  & 4.2 & 0.9 & 3.6 & 5.0 & 7.5 \\
\midrule
sum of codes & 100 & 100 & 63.2 & 99.7 & 82.8 & 55.4 & 61.2 & 94.5 & 92.6 & 66.8 & 60.8 & 50.7 & 59.4 & 53.7 & 56.6 & 77.8 \\
explained & 100 & 100 & 66.7 & 99.5 & 83.7 & 58.6 & 64.0 & 95.0 & 92.7 & 68.3 & 64.0 & 55.2 & 61.4 & 57.1 & 58.9 & 71.7 \\
residual &  &  & 33.3 & 0.5 & 16.3 & 41.4 & 36.0 & 5.0 & 7.3 & 31.7 & 36.0 & 44.8 & 38.6 & 42.9 & 41.1 & 28.3 \\
\midrule
\multicolumn{17}{l}{\textbf{Layer 4, positions 17--31}} \\
\midrule
\multicolumn{1}{l}{$t$} & 17 & 18 & 19 & 20 & 21 & 22 & 23 & 24 & 25 & 26 & 27 & 28 & 29 & 30 & 31 &  \\
\multicolumn{1}{l}{\emph{closes depth}} & -- & 1 & -- & 2 & -- & 1 & -- & 3 & -- & 1 & -- & 2 & -- & 1 & -- &  \\
\midrule
$X_t$ & \textbf{87.4} &  & \textbf{36.9} &  & \textbf{53.4} &  & \textbf{35.8} &  & \textbf{48.6} &  & \textbf{42.6} &  & \textbf{47.7} &  & \textbf{36.4} &  \\
$X_{t+1}$ &  &  & 5.7 &  & 0.2 &  & 4.6 &  &  &  & 4.1 &  & 1.0 &  & 3.8 &  \\
other $X_i$ &  & 0.1 & 0.2 & 0.2 & 0.1 & 0.2 & 0.2 & 0.2 & 0.1 & 0.2 & 0.2 & 0.3 & 0.2 & 0.2 & 0.2 &  \\
$X^{(4)}_{p}$ & 0.2 & \textbf{84.8} & 17.4 & 0.1 &  & \textbf{24.8} & 14.3 & 0.1 &  & \textbf{30.5} & 15.7 &  &  & \textbf{24.2} & 16.8 &  \\
$X^{(4)}_{p-1}$ & 0.1 & 0.1 &  & 0.1 &  &  &  & 0.1 &  &  &  &  &  &  &  &  \\
$X^{(4)}_{p+1}$ &  &  & 1.6 &  & 0.1 & 9.8 & 1.6 &  &  & 2.3 & 1.4 &  & 0.3 & 17.8 & 1.4 &  \\
other $X^{(4)}_i$ &  & 0.1 & 0.1 & 0.2 & 0.2 & 0.2 & 0.1 & 0.5 & 0.2 & 0.3 & 0.1 & 0.2 & 0.3 & 0.2 & 0.1 &  \\
$X^{(3)}_{q}$ & 0.6 & 0.4 & 0.1 & \textbf{62.4} & \textbf{23.7} & 11.4 & 6.5 & 3.0 &  &  &  & 13.3 & 5.8 & 4.6 & 3.9 &  \\
$X^{(3)}_{q-1}$ &  &  &  & 0.4 & 0.2 & 0.2 &  & 0.3 &  &  &  &  &  &  &  &  \\
$X^{(3)}_{q+1}$ &  &  & 0.9 & 0.7 & 0.2 & 2.2 & 0.9 &  &  & 0.6 & 0.8 & \textbf{22.0} & 4.6 & 5.6 & 1.2 &  \\
other $X^{(3)}_i$ &  &  &  &  &  & 0.1 &  & 0.6 & 0.4 & 0.6 & 0.1 & 0.3 & 0.2 & 0.2 & 0.1 &  \\
$X^{(2)}_2$ [9--16] & 0.1 &  &  &  &  &  &  &  &  &  &  &  &  &  &  &  \\
$X^{(2)}_3$ [17--24] &  &  &  & 0.2 & 0.1 & 1.1 & 0.2 & \textbf{51.8} & 20.0 & 10.7 & 1.7 & 4.4 & 2.0 & 1.9 & 1.3 &  \\
$X^{(2)}_4$ [25--32] &  &  &  & 0.1 &  &  &  & 6.1 & 3.3 & 1.9 &  & 2.5 & 0.7 & 0.9 & 0.2 &  \\
$X^{(1)}_1$ [1--16] & 7.3 & 3.6 & 0.1 & 5.0 & 0.6 & 0.6 &  & 3.1 & 1.3 & 1.3 &  & 0.7 & 0.1 & 0.1 &  &  \\
$X^{(1)}_2$ [17--32] &  &  &  & 0.1 &  &  &  & 1.5 & 0.7 & 0.6 &  & 2.7 & 0.8 & 1.3 & 0.3 &  \\
$X^{(0)}_1$ [1--32] &  &  &  & 0.1 &  &  &  & 0.6 & 0.3 & 0.2 &  & 0.1 &  &  &  &  \\
\midrule
sum of codes & 95.8 & 89.3 & 63.1 & 69.6 & 78.8 & 50.7 & 64.4 & 68.0 & 75.0 & 49.3 & 66.7 & 46.6 & 63.6 & 57.0 & 65.7 &  \\
explained & 95.6 & 89.4 & 66.4 & 70.2 & 79.7 & 53.6 & 67.3 & 69.2 & 75.9 & 50.9 & 69.5 & 49.5 & 64.9 & 60.3 & 68.9 &  \\
residual & 4.4 & 10.6 & 33.6 & 29.8 & 20.3 & 46.4 & 32.7 & 30.8 & 24.1 & 49.1 & 30.5 & 50.5 & 35.1 & 39.7 & 31.1 &  \\
\bottomrule
\end{tabular}
\caption{
    \textbf{Encoding map of layer $4$}, all positions; conventions as in Table~\ref{tab:enc-layers12}.
    Layer $4$ keeps the persistent codes of layer $3$ over the same spans, at lower strength: e.g.\ $X^{\scriptscriptstyle(2)}_3$ goes from $52\%$ at $t\,{=}\,24$ to $1.3\%$ at $t\,{=}\,31$ ($90$ and $2.6\%$ in layer $3$).
    The completed depth-$4$ variable $X^{\scriptscriptstyle(1)}_1$ is now encoded at and after the end of its span ($7.5$, $7.3$ and $3.6\%$ at $t\,{=}\,16$--$18$, $5\%$ at $t\,{=}\,20$, $3\%$ at $t\,{=}\,24$), and the root receives a first code ($0.6\%$ at $t\,{=}\,24$).
    Predictions of unseen variables get stronger: the next pair at $9$--$18\%$ for $t\equiv 2\pmod{4}$, the next 4-span at $11$ and $22\%$ at $t\,{=}\,12,28$, and the next token at up to $6\%$ for $t\equiv 3\pmod{4}$.
    The residual is $30$--$50\%$ at most positions.
    }
\label{tab:enc-layer4}
\end{table}

\begin{table}[p]
\centering
\footnotesize
\setlength{\tabcolsep}{2.4pt}
\renewcommand{\arraystretch}{0.98}
\begin{tabular}{lrrrrrrrrrrrrrrrr}
\toprule
\multicolumn{17}{l}{\textbf{Layer 5, positions 1--16}} \\
\midrule
\multicolumn{1}{l}{$t$} & 1 & 2 & 3 & 4 & 5 & 6 & 7 & 8 & 9 & 10 & 11 & 12 & 13 & 14 & 15 & 16 \\
\multicolumn{1}{l}{\emph{closes depth}} & -- & 1 & -- & 2 & -- & 1 & -- & 3 & -- & 1 & -- & 2 & -- & 1 & -- & 4 \\
\midrule
$X_t$ & \textbf{100} &  & 13.0 &  & \textbf{56.6} &  & 6.6 & 0.1 & \textbf{72.3} &  & 13.5 &  & \textbf{32.3} &  & 10.3 & 0.4 \\
$X_{t+1}$ &  &  & 6.3 &  & 0.2 &  & 17.8 &  &  &  & 8.8 &  & 2.8 &  & 18.5 &  \\
other $X_i$ &  &  & 0.3 & 0.1 & 0.2 & 0.3 & 0.3 & 0.3 & 0.1 & 0.1 & 0.3 & 0.3 & 0.3 & 0.2 & 0.3 & 1.0 \\
$X^{(4)}_{p}$ &  & \textbf{100} & 7.0 & 0.1 &  & \textbf{21.2} & 2.1 & 0.4 &  & \textbf{69.8} & 5.4 &  &  & 9.2 & 1.7 & 6.5 \\
$X^{(4)}_{p-1}$ &  &  &  & 0.3 &  &  &  & 0.4 &  &  &  &  &  &  &  & 5.9 \\
$X^{(4)}_{p+1}$ &  &  & 2.5 &  & 0.1 & 11.0 & 2.3 &  &  & 0.3 & 1.6 &  & 0.5 & \textbf{37.0} & 5.3 &  \\
other $X^{(4)}_i$ &  &  & 0.2 &  & 0.2 & 0.1 & 0.2 & 0.6 & 0.1 & 0.1 & 0.2 & 0.2 & 0.4 & 0.1 & 0.2 & 0.1 \\
$X^{(3)}_{q}$ &  &  &  & \textbf{99.0} & 4.3 & 6.3 & 1.5 & 7.4 &  &  &  & 18.4 & 1.2 & 1.0 & 0.7 & \textbf{32.5} \\
$X^{(3)}_{q-1}$ &  &  &  &  &  &  &  & 0.9 &  &  &  &  &  &  &  & 19.7 \\
$X^{(3)}_{q+1}$ &  &  & 1.5 &  & 0.1 & 1.9 & 0.8 &  &  & 0.2 & 1.1 & 7.8 & 1.4 & 3.8 & 0.9 &  \\
other $X^{(3)}_i$ &  &  & 0.1 &  &  & 0.1 & 0.1 &  &  & 0.1 & 0.1 & 0.1 & 0.1 &  & 0.1 & 0.1 \\
$X^{(2)}_1$ [1--8] &  &  &  &  & 0.1 & 1.4 & 0.5 & \textbf{83.7} & 4.2 & 7.8 & 0.4 & 8.5 & 0.6 & 0.5 & 0.6 & 0.7 \\
$X^{(2)}_2$ [9--16] &  &  &  &  &  &  &  &  &  & 0.1 & 0.1 & 3.2 & 0.3 & 0.5 & 0.3 & 8.5 \\
$X^{(1)}_1$ [1--16] &  &  &  &  &  &  &  &  &  & 0.1 &  & 2.1 & 0.2 & 0.2 & 0.5 & 4.6 \\
\midrule
sum of codes & 100 & 100 & 31.0 & 99.5 & 62.0 & 42.4 & 32.4 & 94.0 & 76.8 & 78.4 & 31.5 & 40.7 & 40.1 & 52.6 & 39.3 & 80.1 \\
explained & 100 & 100 & 32.9 & 99.5 & 62.9 & 46.2 & 36.3 & 92.2 & 76.8 & 79.6 & 35.5 & 39.3 & 44.1 & 58.6 & 47.1 & 67.5 \\
residual &  &  & 67.1 & 0.5 & 37.1 & 53.8 & 63.7 & 7.8 & 23.2 & 20.4 & 64.5 & 60.7 & 55.9 & 41.4 & 52.9 & 32.5 \\
\midrule
\multicolumn{17}{l}{\textbf{Layer 5, positions 17--31}} \\
\midrule
\multicolumn{1}{l}{$t$} & 17 & 18 & 19 & 20 & 21 & 22 & 23 & 24 & 25 & 26 & 27 & 28 & 29 & 30 & 31 &  \\
\multicolumn{1}{l}{\emph{closes depth}} & -- & 1 & -- & 2 & -- & 1 & -- & 3 & -- & 1 & -- & 2 & -- & 1 & -- &  \\
\midrule
$X_t$ & \textbf{90.2} &  & 15.0 &  & \textbf{72.8} &  & 8.0 &  & \textbf{84.7} &  & 12.6 &  & \textbf{25.3} &  & 7.0 &  \\
$X_{t+1}$ &  &  & 7.0 &  & 0.2 &  & 19.2 &  &  &  & 13.3 &  & 5.6 &  & \textbf{30.6} &  \\
other $X_i$ &  &  & 0.3 & 0.2 & 0.1 & 0.3 & 0.3 & 0.4 & 0.1 & 0.3 & 0.3 & 0.3 & 0.3 & 0.1 & 0.2 &  \\
$X^{(4)}_{p}$ &  & \textbf{96.5} & 9.1 &  &  & \textbf{20.1} & 2.8 & 0.1 &  & \textbf{33.0} & 4.1 &  &  & 7.8 & 2.2 &  \\
$X^{(4)}_{p+1}$ &  &  & 2.5 &  & 0.1 & 13.9 & 3.0 &  &  & 3.0 & 2.2 &  & 1.0 & \textbf{51.0} & 5.3 &  \\
other $X^{(4)}_i$ &  &  & 0.2 & 0.1 & 0.1 & 0.1 & 0.2 & 0.3 &  & 0.2 & 0.2 & 0.2 & 0.6 & 0.1 & 0.1 &  \\
$X^{(3)}_{q}$ &  &  &  & \textbf{27.6} & 2.1 & 5.2 & 1.9 & 0.4 &  &  &  & 5.8 & 0.8 & 0.5 & 0.5 &  \\
$X^{(3)}_{q-1}$ &  &  &  & 0.1 &  &  &  & 0.1 &  &  &  &  &  &  &  &  \\
$X^{(3)}_{q+1}$ &  &  & 1.3 & 0.3 & 0.1 & 2.2 & 0.9 &  &  & 0.9 & 1.0 & \textbf{23.8} & 4.3 & 4.6 & 1.1 &  \\
other $X^{(3)}_i$ &  &  & 0.1 & 0.1 &  & 0.1 & 0.1 & 0.3 &  & 0.3 & 0.1 & 0.1 & 0.1 &  & 0.1 &  \\
$X^{(2)}_3$ [17--24] &  &  &  & 0.3 & 0.1 & 1.1 & 0.3 & 12.4 & 0.8 & 2.2 & 0.3 & 0.9 & 0.2 & 0.1 & 0.2 &  \\
$X^{(2)}_4$ [25--32] &  &  &  & 0.1 &  &  &  & 2.8 & 0.5 & 0.7 & 0.1 & 4.7 & 0.3 & 0.4 & 0.2 &  \\
$X^{(1)}_1$ [1--16] & 0.7 & 0.7 & 0.1 & \textbf{31.5} & 0.1 & 0.6 & 0.1 & 13.0 & 0.2 & 3.2 & 0.2 & 2.5 & 0.1 &  &  &  \\
$X^{(1)}_2$ [17--32] &  &  &  & 0.1 &  &  &  & 3.4 & 0.1 & 0.7 & 0.1 & 0.9 & 0.1 & 0.1 & 0.1 &  \\
$X^{(0)}_1$ [1--32] &  &  &  & 0.1 &  &  &  & 2.2 &  & 0.4 &  & 0.3 &  &  &  &  \\
\midrule
sum of codes & 91.0 & 97.2 & 35.7 & 60.6 & 75.8 & 43.8 & 36.9 & 35.6 & 86.5 & 44.7 & 34.5 & 39.5 & 38.8 & 64.8 & 47.8 &  \\
explained & 91.0 & 97.1 & 37.3 & 61.8 & 77.9 & 47.3 & 42.2 & 34.8 & 86.8 & 47.3 & 39.6 & 38.4 & 42.7 & 69.9 & 53.1 &  \\
residual & 9.0 & 2.9 & 62.7 & 38.2 & 22.1 & 52.7 & 57.8 & 65.2 & 13.2 & 52.7 & 60.4 & 61.6 & 57.3 & 30.1 & 46.9 &  \\
\bottomrule
\end{tabular}
\caption{
    \textbf{Encoding map of layer $5$}, all positions; conventions as in Table~\ref{tab:enc-layers12}.
    Layer $5$ turns to the output.
    Completed variables keep clean codes at a few positions ($84$--$100\%$ at $t\,{=}\,2,4,8,18$), but the persistent codes of layers $3$ and $4$ are mostly gone (at $t\,{=}\,31$: $X^{\scriptscriptstyle(4)}_{15}$, $X^{\scriptscriptstyle(3)}_7$ and $X^{\scriptscriptstyle(2)}_3$ at $2.2$, $0.5$ and $0.2\%$).
    Elsewhere, the activation is organised around what comes next: for $t\equiv 3\pmod{4}$ the next token holds $6$--$31\%$, more than the current token at $t\,{=}\,7,15,23,27,31$, and the unseen next pair reaches $37$ and $51\%$ at $t\,{=}\,14,30$, against $9$ and $8\%$ for the pair just completed.
    The completed depth-$4$ variable $X^{\scriptscriptstyle(1)}_1$ reaches $31.5\%$ at $t\,{=}\,20$, and the root its largest code, $2.2\%$, at $t\,{=}\,24$.
    The residual exceeds $30\%$ at two thirds of the positions, up to $67\%$ at $t\,{=}\,3$.
    }
\label{tab:enc-layer5}
\end{table}

\clearpage
\section{Composition of the autocorrelations: details}
\label{app:rhm-composition}

This appendix derives the split of Eq.~\ref{eq:rhm-trace-split}, uses it to break down the autocorrelations of layers $3$ and $4$, and summarises the other layers.

\paragraph{Derivation of Eq.~\ref{eq:rhm-trace-split}.}
In this appendix we call the reference position $T$ the \emph{anchor} and the position $T-\tau$ the \emph{target}.
Write the activations at both positions as in Eq.~\ref{eq:linear-representation-hypothesis}, $\phi_T = A_T\mathbf{Y} + \epsilon_T$ and $\phi_{T-\tau} = A_{T-\tau}\mathbf{Y} + \epsilon_{T-\tau}$, with the same $\mathbf{Y}$: the regression at each position uses all $63$ nodes, whether their span is complete at that position or not.
Expanding the dot product of Eq.~\ref{eq:autocorrelations} gives four terms,
\begin{align}\label{eq:trace-four-terms}
    \mathbb{E}\!\left[\phi_{T-\tau}\cdot\phi_T\right]
    = \mathrm{Tr}\!\left[A_T\,\mathrm{Cov}(\mathbf{Y})\,A_{T-\tau}^\top\right]
    + \mathbb{E}\!\left[\epsilon_{T-\tau}\cdot A_T\mathbf{Y}\right]
    + \mathbb{E}\!\left[A_{T-\tau}\mathbf{Y}\cdot\epsilon_T\right]
    + \mathbb{E}\!\left[\epsilon_{T-\tau}\cdot\epsilon_T\right],
\end{align}
and the two mixed terms vanish because each residual is uncorrelated with every node, $\mathbb{E}[\epsilon_{T-\tau}\cdot A_T\mathbf{Y}] = \mathrm{Tr}\!\left[A_T\,\mathrm{Cov}(\mathbf{Y},\epsilon_{T-\tau})\right] = 0$, which is the normal equation of least squares at position $T-\tau$.
Dividing by $\sqrt{\mathrm{Tr}\,\Sigma_T\,\mathrm{Tr}\,\Sigma_{T-\tau}}$, the normalisation of Eq.~\ref{eq:autocorrelations}, gives Eq.~\ref{eq:rhm-trace-split}.
The split holds for any pair of positions and involves no cross-position moment: the code part is a prediction of the autocorrelation from the single-position regressions alone.
Expanding $\mathrm{Cov}(\mathbf{Y})$ over its $63\times63$ blocks $\mathrm{Cov}(\mathbf{y}_X,\mathbf{y}_{X'})$ splits the code part into a sum over pairs of nodes,
\begin{align}\label{eq:same-cross}
    c^{\rm code}_\ell(\tau)
    = \sum_X \frac{\mathrm{Tr}\!\left[A_{T,X}\,\Sigma_X\,A_{T-\tau,X}^\top\right]}{\sqrt{\mathrm{Tr}\,\Sigma_T\,\mathrm{Tr}\,\Sigma_{T-\tau}}}
    + \sum_{X\neq X'} \frac{\mathrm{Tr}\!\left[A_{T,X}\,\mathrm{Cov}(\mathbf{y}_X,\mathbf{y}_{X'})\,A_{T-\tau,X'}^\top\right]}{\sqrt{\mathrm{Tr}\,\Sigma_T\,\mathrm{Tr}\,\Sigma_{T-\tau}}},
\end{align}
the same-node and the cross-node terms of the main text.
The same-node term of $X$ is the expected dot product of the code of $X$ at the anchor with the code of $X$ at the target, $\mathbb{E}[(A_{T-\tau,X}\mathbf{y}_X)\cdot(A_{T,X}\mathbf{y}_X)]$: the autocorrelation that the variable would produce on its own.
A cross-node term couples two different variables through their correlation, which in the tree runs through the rule drawn at their lowest common ancestor.
In practice the code part is computed from the regressions at $P\,{=}\,2^{18}$ and the measurement at $P\,{=}\,2^{25}$ (Appendix~\ref{app:rhm-setup}), and the residual part is their difference.
The identity is exact within the regression sample, and the sampling error of the measurement is negligible, so the residual part inherits the resolution of the regression discussed in Appendix~\ref{app:probing}; a same-node term is linear in the strength of a faint code, with the loud code at the other position acting as a matched filter, so it resolves retained codes fainter than the strengths of Tables~\ref{tab:enc-layers12}--\ref{tab:enc-layer5} can certify, provided it is stable in $P$.
Beyond lag $16$ the cross-node terms rest on the covariance of nodes in different halves of the sentence, which is at its sampling floor at $P\,{=}\,2^{18}$, and we do not use them.

\begin{figure}[h]
    \centering
    \includegraphics[width=\linewidth]{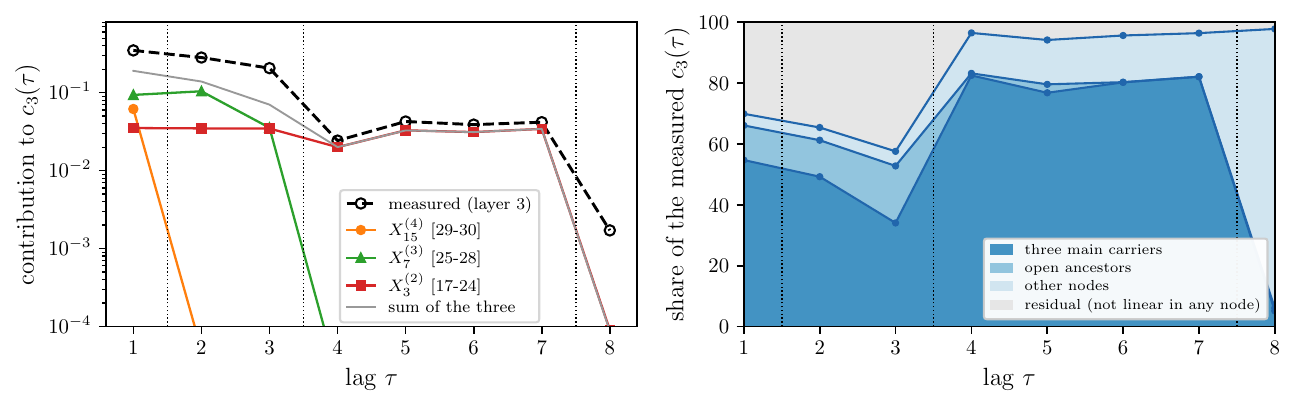}\\
    \includegraphics[width=\linewidth]{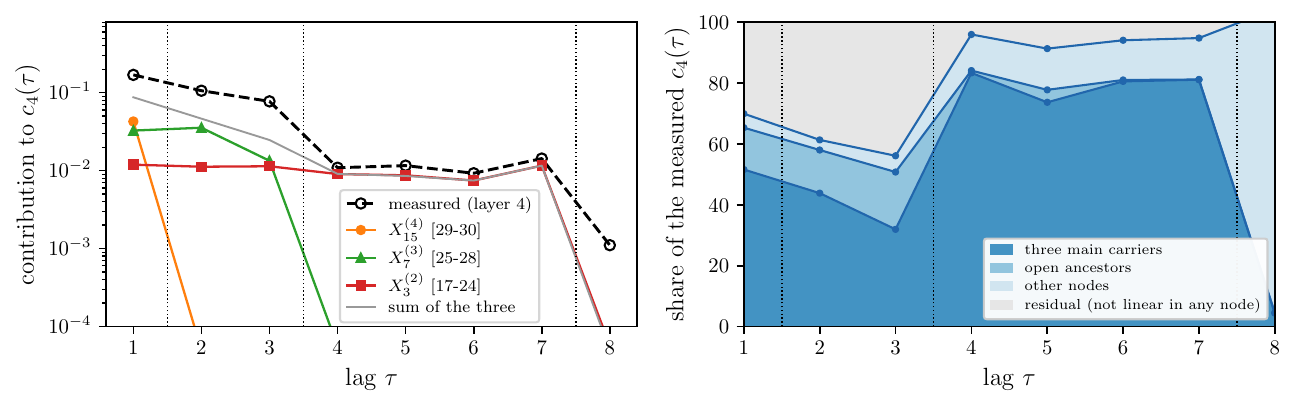}
    \caption{Composition of the autocorrelations of layer $3$ (top) and layer $4$ (bottom). \textbf{Left:} measured autocorrelation (dashed) against the same-node terms of the three carriers $X^{(4)}_{15}$, $X^{(3)}_{7}$ and $X^{(2)}_{3}$ and their sum (grey), as in Fig.~\ref{fig:rhm-trace-composition}. \textbf{Right:} share of the measured autocorrelation accounted for, from bottom to top, by the three carriers, by the open ancestors of the anchor, by all other node codes including the cross-node terms, and by the non-additive residual; the four bands sum to $100\%$. Dotted lines at the span ends of the carriers.}
    \label{fig:rhm-trace-shares}
\end{figure}

\begin{figure}[h]
    \centering
    \includegraphics[width=0.6\linewidth]{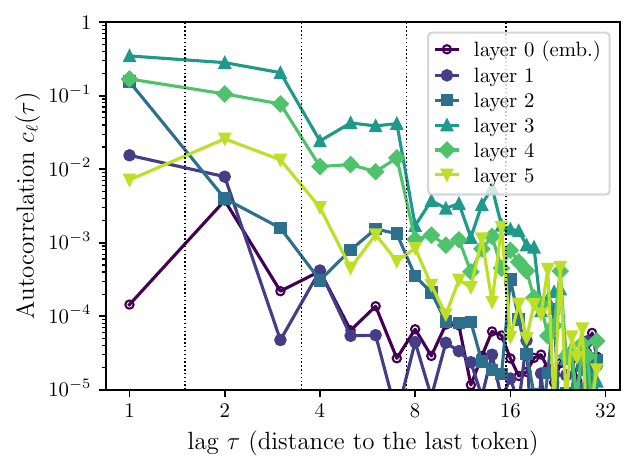}
    \caption{Autocorrelation $c_\ell(\tau)$ of all six layers of the transformer trained on the RHM, as in Fig.~\ref{fig:main}a with layer $5$ added: absolute values on a log scale, dotted lines at lags $1.5$, $3.5$, $7.5$ and $15.5$. The embedding is drawn with hollow markers. Layer $4$ repeats the staircase of layer $3$ at a lower amplitude; layer $5$ is negative at lag $1$ (absolute value shown), peaks at lag $2$, ten times below layer $3$, and lies within a few noise floors at most lags beyond $8$.}
    \label{fig:rhm-all-layers}
\end{figure}

\paragraph{Layers $3$ and $4$.}
The strengths of the codes at the anchor determine how much each of them contributes to the autocorrelation, through the same-node terms of Eq.~\ref{eq:same-cross}.
Fig.~\ref{fig:rhm-trace-shares} shows this for layers $3$ and $4$, the two layers whose autocorrelation is a staircase.
The left panels repeat the right panel of Fig.~\ref{fig:rhm-trace-composition}; the right panels split the measured autocorrelation at lags $1$--$8$ into four shares that sum to $100\%$: the same-node terms of the three carriers, those of the open ancestors of the anchor ($X^{(4)}_{16}$, $X^{(3)}_8$, $X^{(2)}_4$, $X^{(1)}_2$ and the root, whose spans contain the unseen $X_{32}$), all other terms of $c^{\rm code}_\ell$, i.e.\ the same-node terms of the other nodes and all the cross-node terms, and the residual $c^{\rm res}_\ell$.
The two layers have the same composition.
The carriers give $34$--$55\%$ (layer $3$) and $32$--$52\%$ (layer $4$) at lags $1$--$3$, and $77$--$83\%$ and $74$--$83\%$ at lags $4$--$7$, where $X^{(2)}_3$ acts alone; the open ancestors add $11$--$19\%$ and $14$--$19\%$ at lags $1$--$3$; the residual is $30$--$42\%$ and $30$--$44\%$ at lags $1$--$3$, and $4$--$6\%$ and $4$--$9\%$ at lags $4$--$7$; the cross-node terms stay within $1$--$9\%$ up to lag $7$ (Table~\ref{tab:trace-shares}).
The amplitude is what differs: the carriers hold $28$, $7.4$ and $2.6\%$ of the anchor variance in layer $3$ and $17$, $3.9$ and $1.3\%$ in layer $4$ (Tables~\ref{tab:enc-layer3} and~\ref{tab:enc-layer4}), and the autocorrelation of layer $4$ is $2$--$4$ times smaller over lags $1$--$7$ (Fig.~\ref{fig:rhm-all-layers}), with the same steps at the span ends (a factor $7$ between lags $3$ and $4$ and $13$ between lags $7$ and $8$, against $8.5$ and $24$ in layer $3$).
What layer $4$ adds to the encoding of layer $3$ leaves little trace in the autocorrelation.
The louder prediction codes contribute little: the pair $X^{(4)}_{16}$, encoded at $18\%$ by the target of lag $1$ but at $1.4\%$ by the anchor, gives $4\%$ of the lag-$1$ value.
The completed depth-$4$ variable $X^{(1)}_{1}$, whose window would cover lags $8$--$15$, is not retained at the anchor above the resolution of the regression ($0.007\%$, at the noise floor of Appendix~\ref{app:probing}), so it contributes nothing there.
At lag $8$, beyond the span of $X^{(2)}_3$, the autocorrelation drops by more than a factor ten and the carriers give only $5\%$ of it; from there on, the cross-node terms grow to $21$--$35\%$ and change sign from lag to lag.

\paragraph{Other layers.}
Fig.~\ref{fig:rhm-all-layers} shows the autocorrelation of all six layers, the embedding and the five blocks (Fig.~\ref{fig:main}a is its restriction to layers $0$--$4$), and Table~\ref{tab:trace-shares} the composition of layers $1$--$4$ at lags $1$--$8$.
Layer $1$ is above the noise only at lags $1$--$2$, where the codes account for $96$--$102\%$ of the measurement, mostly through same-node terms of faint retentions overlapping with loud codes: at lag $1$, for instance, the anchor retains the pair $X^{(4)}_{15}$ at $3.7\%$ of its variance, while the target, position $30$, holds it at $97.5\%$.
Layer $2$ is the one layer whose autocorrelation is carried by cross-node terms beyond lag $1$: at lags $2$ and $3$, where the measured value is $40$--$100$ times smaller than at lag $1$, they give $100$ and $56\%$ of it, coupling the anchor's code of $X^{(4)}_{15}$ with the codes of its child $X_{29}$ (lag $2$) and of $X^{(3)}_{7}$, the sibling of its parent (lag $3$).
Layer $5$ is organised around the output.
Its variance is concentrated in a few directions (the effective rank $r$ of Appendix~\ref{app:rhm-setup} is $6$--$25$ across lags, against $250$--$460$ for layers $1$--$4$), its loud codes are predictions whose content changes from position to position (Table~\ref{tab:enc-layer5}), and the additive decomposition is no longer quantitative: the code part is $2.6$ times the measurement at lag $1$ and $0.2$--$0.4$ times it at lags $2$--$4$, the residual dominating. For this reason, layer $5$ is left out of Table~\ref{tab:trace-shares}.

\begin{table}[tbp]
\centering
\footnotesize
\setlength{\tabcolsep}{4pt}
\begin{tabular}{llrrrrrrrr}
\toprule
layer & share of $c_\ell(\tau)$ & $\tau\,{=}\,1$ & $\tau\,{=}\,2$ & $\tau\,{=}\,3$ & $\tau\,{=}\,4$ & $\tau\,{=}\,5$ & $\tau\,{=}\,6$ & $\tau\,{=}\,7$ & $\tau\,{=}\,8$ \\
\midrule
\multirow{3}{*}{1} & same-node & 80 & 72 & -- & -- & -- & -- & -- & -- \\
 & cross-node & 16 & 30 & -- & -- & -- & -- & -- & -- \\
 & residual & 4 & -2 & -- & -- & -- & -- & -- & -- \\
\midrule
\multirow{3}{*}{2} & same-node & 93 & -5 & 32 & -- & 82 & 65 & 64 & -- \\
 & cross-node & 7 & 100 & 56 & -- & 8 & 21 & 35 & -- \\
 & residual & 0 & 6 & 12 & -- & 10 & 14 & 1 & -- \\
\midrule
\multirow{3}{*}{3} & same-node & 67 & 62 & 54 & 92 & 86 & 88 & 88 & 76 \\
 & cross-node & 3 & 3 & 3 & 5 & 8 & 8 & 9 & 21 \\
 & residual & 30 & 35 & 42 & 4 & 6 & 4 & 4 & 2 \\
\midrule
\multirow{3}{*}{4} & same-node & 66 & 59 & 52 & 95 & 86 & 91 & 88 & 70 \\
 & cross-node & 4 & 2 & 4 & 1 & 5 & 3 & 7 & 35 \\
 & residual & 30 & 39 & 44 & 4 & 9 & 6 & 5 & -4 \\
\bottomrule
\end{tabular}
\caption{\textbf{Composition of the autocorrelations of layers 1--4 at lags 1--8.} Same-node, cross-node and residual parts of Eq.~\ref{eq:rhm-trace-split}, in percent of the measured $c_\ell(\tau)$ (the three rows of each layer sum to $100$). Cells are left blank where the measured value lies within 20 noise floors (Appendix~\ref{app:rhm-setup}), as at lag $4$ of layer $2$, where the terms cancel. The code part is computed from the regression at $P\,{=}\,2^{18}$ and the measurement at $P\,{=}\,2^{25}$.}
\label{tab:trace-shares}
\end{table}

\section{Development over training: details}
\label{app:rhm-training}

\begin{figure}[h]
    \centering
    \includegraphics[width=0.5\linewidth]{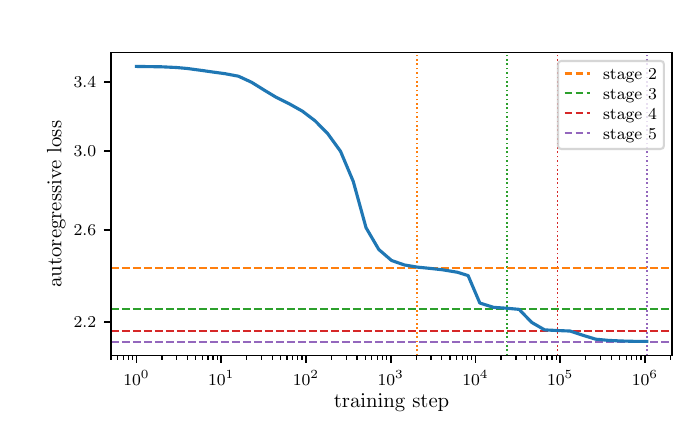}
    \caption{Autoregressive validation loss (mean over the $32$ positions) against the training step, with the stage values $(\log v + (2^d-1)\log m)/2^d$ for $d\,{=}\,2,\dots,5$ (dashed) and the checkpoints of Fig.~\ref{fig:rhm-dynamics} (dotted).}
    \label{fig:rhm-training-ar}
\end{figure}

\paragraph{Checkpoints and stages.}
The checkpoints of Fig.~\ref{fig:rhm-dynamics} are the training steps $2048$, $23552$, $93184$ and $1048576$, which sit on the plateaus of stages $2$ to $5$: their last-token losses are $0.990$, $0.392$, $0.202$ and $0.156$ nats against the stage values $0.945$, $0.387$, $0.201$ and $0.156$, and their autoregressive losses $2.431$, $2.256$, $2.166$ and $2.124$ nats per token against $2.426$, $2.253$, $2.166$ and $2.123$ (Fig.~\ref{fig:rhm-training-ar}).
The first stage, $d\,{=}\,1$, is crossed within the first few hundred steps without a visible plateau.
At the last checkpoint the loss is at its optimal value at every position, including the positions of the second half of the sentence whose prediction depends on the first half through the depth-$4$ variable; this cross-half information is acquired between the checkpoints of stages $4$ and $5$.

\paragraph{Layer 4.}
At the checkpoint on stage $3$, lags $1$ and $2$ are at their final values while lag $3$ is at a third of it; lag $3$ completes at stage $4$, together with the plateau at lags $4$--$7$, which appears at $2$--$3.5$ times its final height and settles at stage $5$; the plateau at lags $8$--$15$ is negative at stage $4$ and becomes positive at stage $5$ (the panel shows absolute values).
Each plateau thus appears one stage after the loss shows that its carrier has been learned: the plateau of the depth-$2$ carrier $X^{(3)}_7$ at stage $3$, that of the depth-$3$ carrier $X^{(2)}_3$ at stage $4$.
This suggests that a variable is carried across positions once its parent, which is assembled from it, is being learned; testing it requires the decomposition of Eq.~\ref{eq:linear-representation-hypothesis} at the intermediate checkpoints.

\begin{figure}[h]
    \centering
    \includegraphics[width=0.32\linewidth]{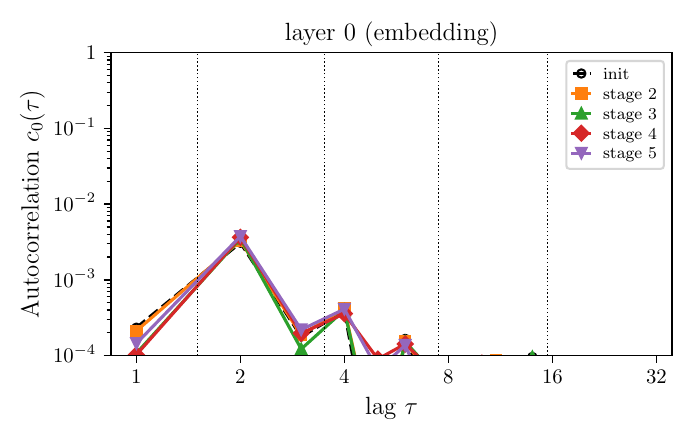}
    \includegraphics[width=0.32\linewidth]{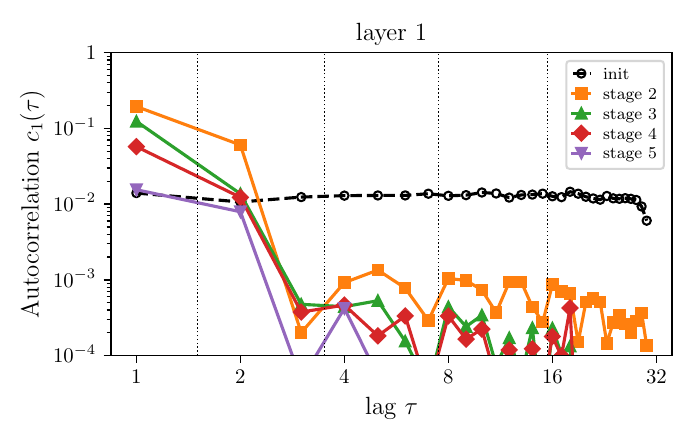}
    \includegraphics[width=0.32\linewidth]{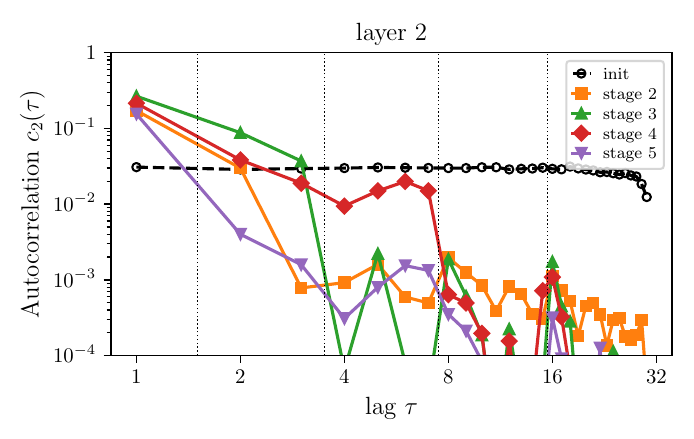}\\
    \includegraphics[width=0.32\linewidth]{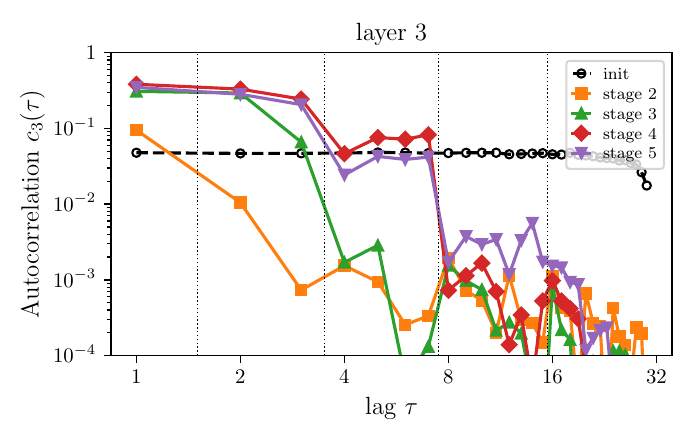}
    \includegraphics[width=0.32\linewidth]{figures/trace-training_layer4_n1-30.pdf}
    \caption{Absolute value of the autocorrelation of layers $0$ to $4$ at initialisation and at the four checkpoints of Fig.~\ref{fig:rhm-dynamics}, same conventions as Fig.~\ref{fig:rhm-dynamics}b.}
    \label{fig:rhm-training-layers}
\end{figure}

\paragraph{The other layers.}
Fig.~\ref{fig:rhm-training-layers} shows the same measurement for layers $0$ to $4$.
The embedding (layer $0$) does not change over training, as its autocorrelation is the token-level one pushed through the embedding vectors.
Layer $2$ holds the plateau at lags $4$--$7$ at stage $4$ and has handed it over to layer $3$ by the end of training, and the autocorrelation of layer $1$ at lag $1$ decreases monotonically from the first checkpoint on, as the layer specializes to the tokens and the depth-$1$ variables.
Layer $3$ follows layer $4$ step by step at a higher amplitude.

\paragraph{Initialisation.}
At initialisation every block has a positive, lag-independent autocorrelation, growing with depth from $1.3\times 10^{-2}$ (layer $1$) to $6.6\times 10^{-2}$ (layer $4$), three orders of magnitude above the noise floor, while the embedding has none.
The mechanism is causal attention with nearly uniform weights: each block adds to the residual stream the average of the value-projected inputs over the context, so that the reference position and any target share the target's whole prefix.
This predicts an autocorrelation independent of the lag up to a factor $t'/(t'+1)$ at target position $t'$, a level growing linearly with the number of blocks, and a positive sign, all of which are observed; the absolute level is reproduced only within a factor $1.2$--$2.5$, the fluctuations of the attention weights around uniform being neglected.
Training removes this component: the first checkpoint retains a residual tail of about $10^{-3}$ at every lag, and the trained layer $1$ is at the noise floor beyond lag $5$.

\section{Autocorrelations of the other models}
\label{app:autocorrelations-other-models}

\begin{figure}[h]
    \centering
    \includegraphics[width=0.9\linewidth]{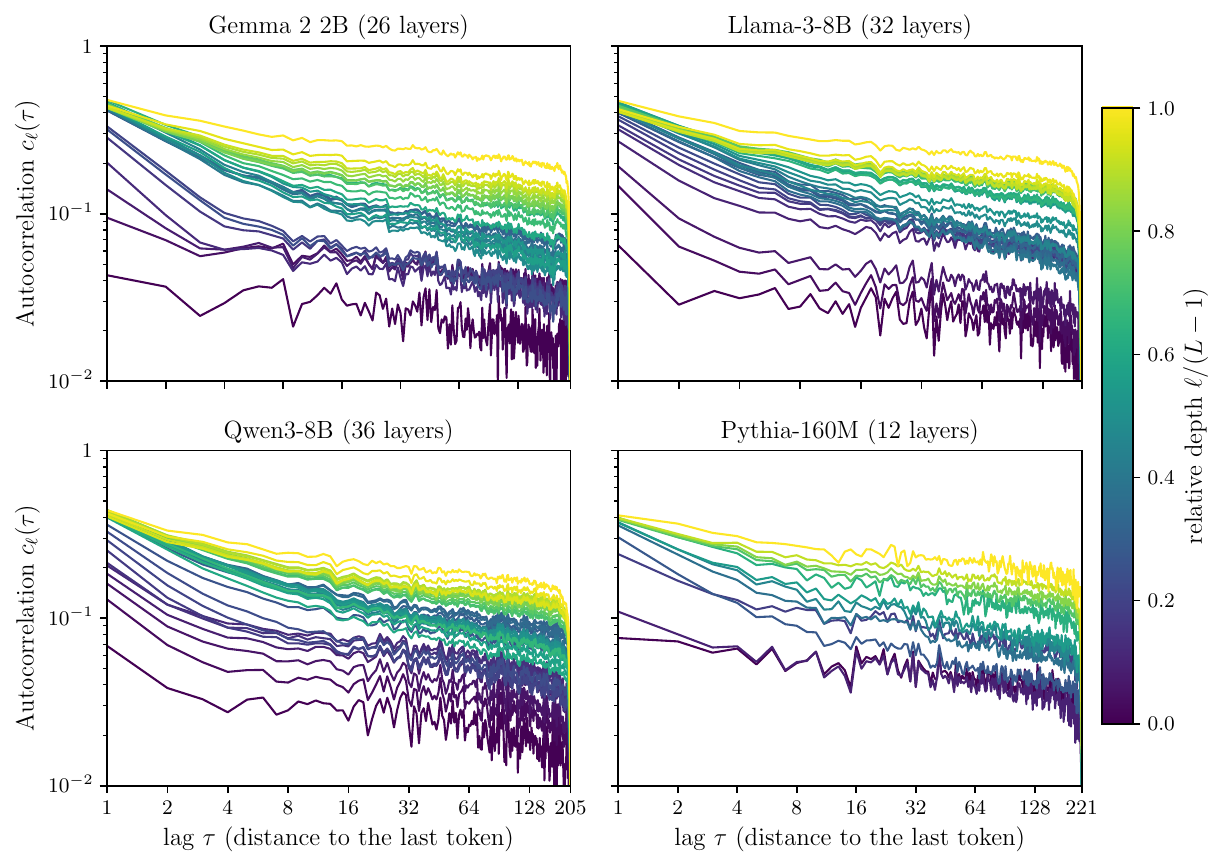}
    \caption{Autocorrelation $c_\ell(\tau)$ against the lag $\tau$ for every layer of Gemma 2 2B, Llama-3-8B, Qwen3-8B and Pythia-160M on The Pile, one curve per layer colored by relative depth $\ell/(L-1)$ (a shared color scale, as the layer counts differ), same vertical range in all panels. The corresponding panel for Gemma 2 9B is Fig.~\ref{fig:main}b.}
    \label{fig:autocorrelations-other-models}
\end{figure}

Fig.~\ref{fig:main}b shows the autocorrelation $c_\ell(\tau)$ of every layer of Gemma 2 9B on The Pile.
Fig.~\ref{fig:autocorrelations-other-models} shows the same measurement for the other four models.
Each curve is computed exactly as in Fig.~\ref{fig:main}b.
We take $500$ samples of the corpus, cut them to the same length, and correlate the activation of the last token of each sample with the activation $\tau$ tokens earlier.
Before doing so we subtract, at each position, the mean activation over the $500$ samples, and we divide by the norms so that the result is a cosine.
In every autocorrelation of this paper, Fig.~\ref{fig:main}b included, the activations of each layer of each sample are first clipped to the $5$th and $95$th percentiles of their values, taken over all positions and coordinates of that layer and sample, to limit the weight of a few very large coordinates.
Appendix~\ref{app:clip-control} repeats the measurement without clipping.

In every model, deeper layers have a larger autocorrelation at any given lag.
The shallowest layers decay within a few tokens.
The deepest layers sit on a flat tail that spans the whole context; Appendix~\ref{app:subtracting-mean-chi} shows that this tail is an offset shared by all the tokens of a sample, not a slow decay.
The order of the layers is clean in the Gemma and Llama models and noisier in Pythia-160M, whose curves cross in places.
At the largest lags the token $\tau$ positions before the reference is the first token of the sample, whose activation is atypical (Appendix~\ref{sec:SAE-explained-variance}), so we do not read anything into the drop of the curves there.

\section{Sequence-wide component of the autocorrelations and frequency control}
\label{app:subtracting-mean-chi}

\begin{figure}[h]
    \centering
    \includegraphics[width=\linewidth]{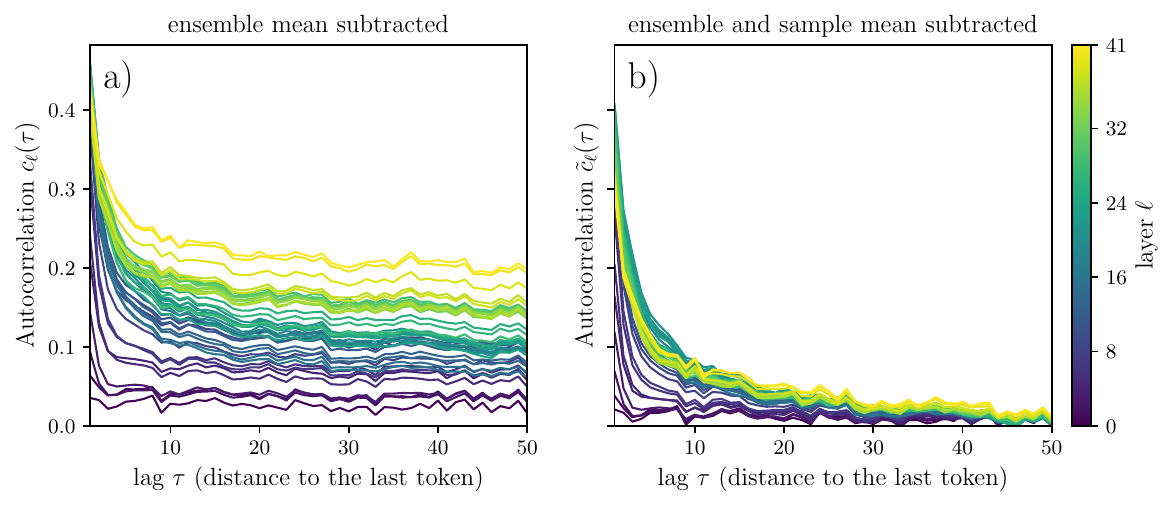}
    \caption{Autocorrelation of every layer of Gemma 2 9B on The Pile over the first 50 lags, on linear axes, with the position-wise mean over samples subtracted as in Eq.~\ref{eq:autocorrelations} (a) and with the mean token of each sample subtracted as well (b). One curve per layer, same vertical range in both panels, with the vertical axis starting at zero for readability.}
    \label{fig:mean-token-subtraction}
\end{figure}

The autocorrelations of natural-language representations can contain two conceptually distinct contributions: a sequence-wide component, shared across the positions of a sample, and a position-dependent component that changes along the context.
To separate them, let \(m=T^{-1}\sum_t\phi_t\) denote the mean representation of a sequence and \(\delta_t=\phi_t-m\). 
In this appendix we recompute the autocorrelation using only \(\delta_t\), and we call it $\tilde{c}_{\ell}(\tau)$.
This removes the sequence-wide component \(m\) from every position, while retaining fluctuations around it.
The two components need not be viewed as signal and artifact: the sequence mean may itself encode information that remains relevant across the full context, while \(\delta_t\) isolates representational structure that changes with position.
Figure~\ref{fig:mean-token-subtraction} shows that the slowly varying tail has a substantial sequence-wide contribution.
Even after centering of the activations, correlations approach a non-zero plateau whose magnitude increases with depth.
Removing \(m\) eliminates this plateau.
Thus, the increase of the very-long-range tail in the deepest layers reflects information shared broadly across the sequence, whereas the position-dependent component saturates earlier with depth.

\begin{figure}[h!]
    \centering
    \includegraphics[width=\linewidth]{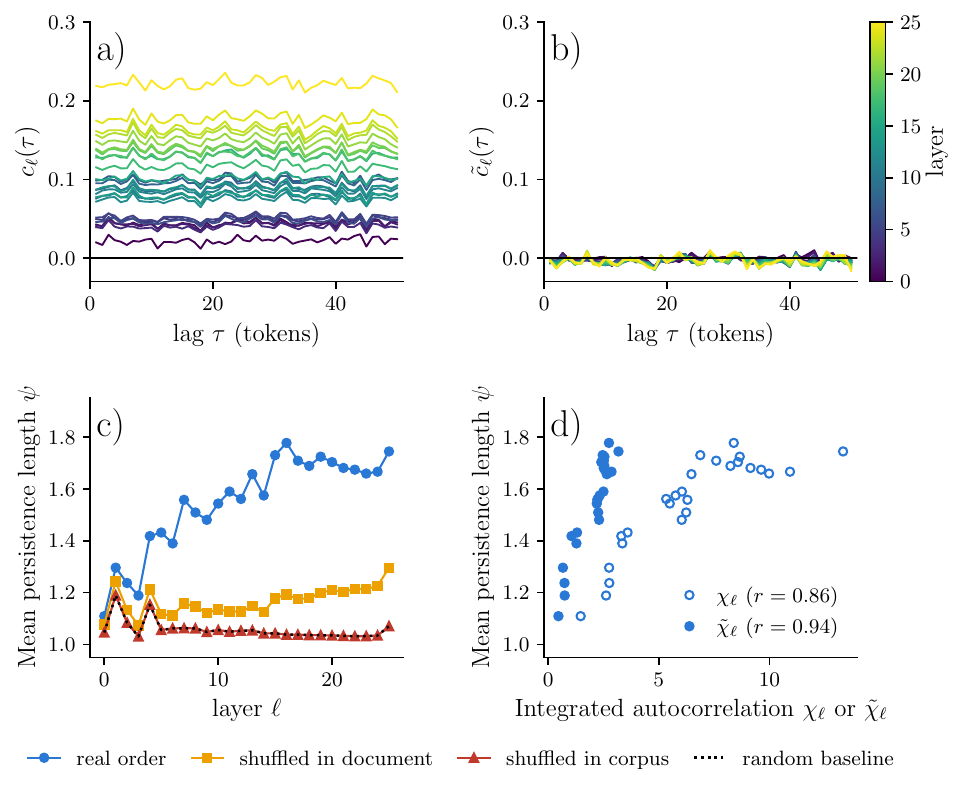}
    \caption{Shuffle control, Gemma 2 2B on The Pile. (a) Autocorrelation $c_\ell(\tau)$ of every layer after permuting at random the token positions of each sequence. (b) The same with the mean $m$ of each sequence subtracted, $\tilde{c}_\ell(\tau)$. (c) Mean persistence $\psi$ of the $1000$ most active latents of each layer (strict episodes) in the real order, with the positions shuffled within each sequence, and with the tokens shuffled across the whole corpus; dotted line: expectation for latents active independently at each token with their corpus frequency. (d) Mean persistence against the integrated autocorrelation, $\chi_\ell$ (hollow) and $\tilde{\chi}_\ell$ (filled), both measured in the real order, one point per layer; $r$ is the Pearson correlation over the $26$ layers.}
    \label{fig:frequency-control}
\end{figure}

\paragraph{Shuffle control.}
\label{app:frequency-control}
In terms of SAE latents, the sequence-wide component $m$ is set by the activation frequencies: it depends on how often each latent is active in a sequence, and not on when, while the order of the tokens enters only through $\delta_t$.
We test this reading by permuting at random the token positions of each sequence, for the activations $\phi_t$ and for the SAE codes computed from them.
The permutation keeps, for every latent, the number of tokens of each sequence on which it is active, hence its activation frequency and $m$, and destroys any other temporal pattern.
Fig.~\ref{fig:frequency-control} shows the outcome for Gemma 2 2B on The Pile.
After the shuffle, the autocorrelation no longer decays with the lag, but it sits on a plateau that grows with depth (Fig.~\ref{fig:frequency-control}a).
Indeed, for a sequence with activations $\phi_1,\dots,\phi_T$, the product between two distinct positions, averaged over the permutations, is
\begin{equation}
    \frac{1}{T(T-1)}\sum_{s\neq t}\phi_s\cdot\phi_t = \|m\|^2 - \frac{1}{T-1}\,\frac{1}{T}\sum_{t}\|\delta_t\|^2,
    \label{eq:shuffled-autocorrelation}
\end{equation}
whatever the lag between them, and the same average for the fluctuations $\delta_t$ is the second term alone.
Therefore, up to a correction of order $1/T$, below $10^{-2}$ here, the autocorrelation of the shuffled activations equals $\mathbb{E}\left[\|m\|^2\right]$ at every lag, whereas $\tilde{c}_\ell(\tau)$ vanishes, as Fig.~\ref{fig:frequency-control}b confirms.
The plateau of $c_\ell(\tau)$ is thus all that the activation frequencies alone produce, and subtracting $m$ removes it.

\paragraph{Persistence of the shuffled sequences.}
The persistence $\psi$ could also reflect how often the SAE latents are active, rather than when: a latent that is active on a fraction $f$ of the tokens, independently at each token, has runs of mean length $1/(1-f)$, which grows with $f$.
We measure this contribution under shuffling in Fig.~\ref{fig:frequency-control}c for the $1000$ most active latents of each layer (strict episodes, Appendix~\ref{app:SAE-layer-dependencies}).
In the real order (blue), $\psi$ grows from $1.1$ at layer $0$ to $1.8$.
With the positions shuffled within each sequence (yellow), it stays between $1.1$ and $1.3$.
With the tokens shuffled across the whole corpus (red), it coincides with the expectation for latents that are active independently at each token with their corpus frequency $f$, which does not grow with depth.
The mean activation frequency of the selected latents does not grow with depth either, staying within $0.03$--$0.07$ at every layer.
Therefore, the growth of the persistence with depth is a property of the temporal order of the activations, not of their frequency.

\begin{figure}[h]
    \centering
    \includegraphics[width=0.5\linewidth]{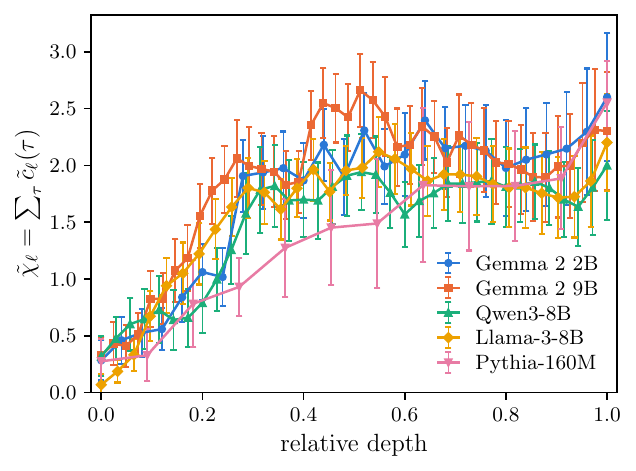}
    \caption{Layer dependence of the integrated autocorrelation with the mean token of each sample removed, $\tilde{\chi}_\ell$, for the five models, as Fig.~\ref{fig:layer-dependences-NLP}a for $\chi_\ell$. Markers are the average of The Pile and WikiText, error bars half the difference between the two corpora.}
    \label{fig:mean-token--susceptibility}
\end{figure}

\paragraph{Position-dependent correlations and persistence.}
Importantly, removing the sequence-wide component does not eliminate the relationship with persistence.
The integrated autocorrelation \(\tilde\chi_\ell = \sum_\tau \tilde{c}_\ell(\tau)\) grows through the first half of the network and subsequently saturates, more closely resembling the depth profile of SAE feature persistence (see Fig.~\ref{fig:mean-token--susceptibility}). Correspondingly, the correlation between the two quantities remains high and increases after mean-token removal in most model–corpus pairs (Table~\ref{tab:docmean-correlations}).
For Gemma 2 2B on The Pile, over all $26$ layers, the Pearson correlation with the mean persistence goes from $r\,{=}\,0.86$ for $\chi_\ell$ to $r\,{=}\,0.94$ for $\tilde{\chi}_\ell$ (Fig.~\ref{fig:frequency-control}d).

\section{Layer dependence of SAE-latent persistence}
\label{app:SAE-layer-dependencies}

\begin{figure}[h]
    \centering
    \includegraphics[width=0.95\linewidth]{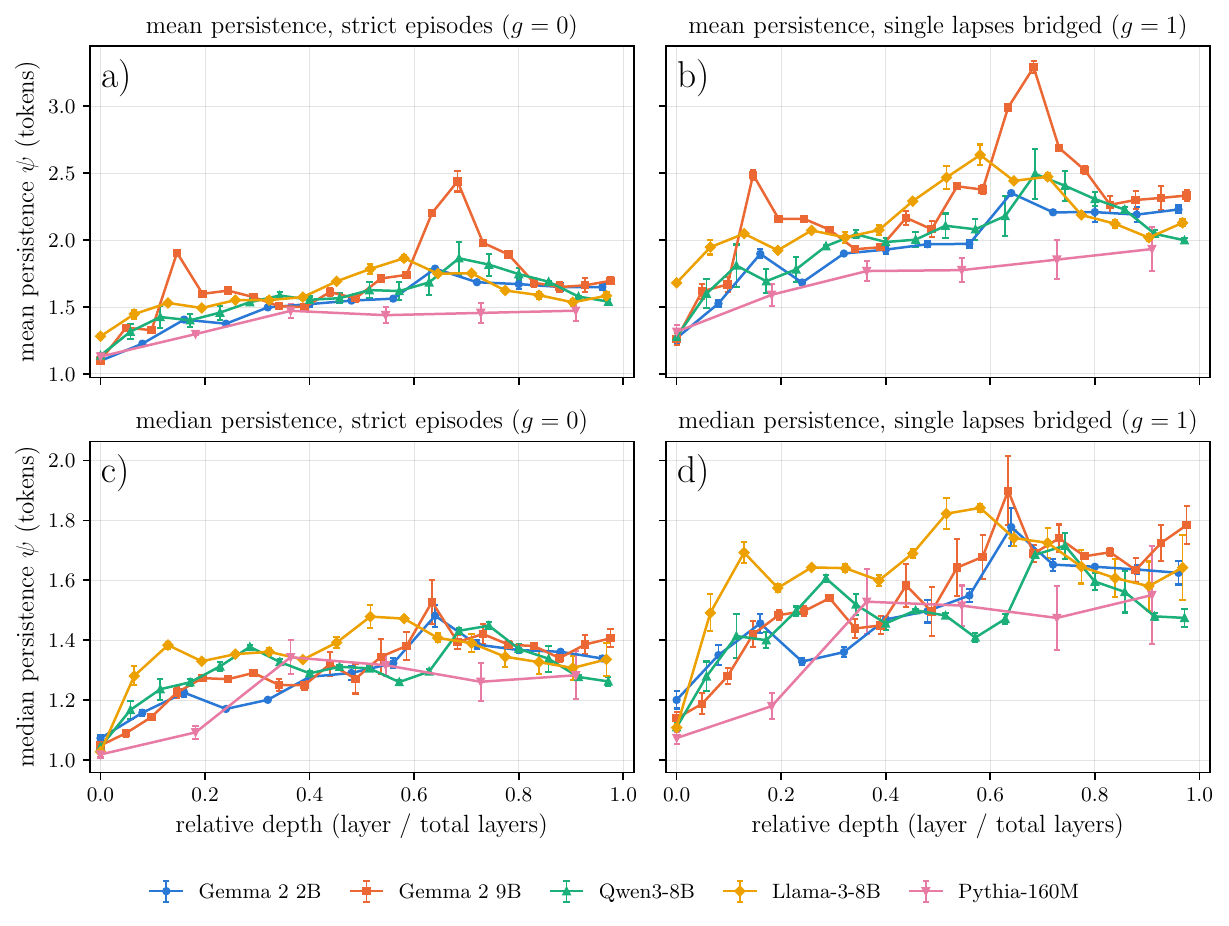}
    \caption{SAE-latent persistence $\psi$ as a function of relative depth for the five models: mean (a, b) and median (c, d) over the 1000 most active latents of each layer, for strict firing episodes ($g\,{=}\,0$; a, c) and with single inactive tokens bridged ($g\,{=}\,1$; b, d). Markers are the average of The Pile and WikiText, error bars half the difference between the two corpora.}
    \label{fig:SAE-layer-dependencies}
\end{figure}

The persistence statistics of Figs.~\ref{fig:layer-dependences-NLP}b, \ref{fig:layer-dependences-NLP}c, \ref{fig:SAE-layer-dependencies} and \ref{fig:SAE-distributions-all-models} are computed on $2722$ samples of each corpus, cut to a common length of $150$ to $225$ tokens depending on the tokenizer.
At each layer we keep the $1000$ latents that are active on the largest number of tokens, because rarer latents have too few firing episodes for a stable average.
The same rule is applied at every layer, so the set of latents changes from layer to layer.
Each latent gets one number, its persistence averaged over its firing episodes, and a layer is summarized by the mean or the median of these numbers over its $1000$ latents.
The sweep covers every second layer of each model.

\paragraph{Gap tolerance.}
Sec.~\ref{sec:nlp} ends a firing episode at the first token on which the feature is inactive.
An SAE feature can, however, switch off for one token and switch back on at the next.
A JumpReLU SAE sets to zero any code below a fixed threshold, and a top-$k$ SAE keeps only the $k$ largest codes of each token, so a feature that is barely active can drop out for a single token.
If such one-token lapses are noise in the readout rather than the feature really stopping and restarting, its persistence is underestimated.

We therefore implemented a gap-tolerant definition to investigate this phenomenon, where two firing episodes of the same SAE feature separated by a gap $g$ are merged together.
$g\,{=}\,0$ gives back original setup without any gap tolerance.
Figs.~\ref{fig:layer-dependences-NLP}b and \ref{fig:layer-dependences-NLP}c use $g\,{=}\,1$, for which the persistencies are obviously larger.
Fig.~\ref{fig:SAE-layer-dependencies} shows that the layer profiles of the SAE latent persistencies do not change qualitatively by the usage of $g=1$.

One-token lapses are not rare.
For a typical feature (the median over the $1000$ latents of a layer), $5$--$11\%$ of its strict episodes are followed by exactly one inactive token and then another episode of the same feature; over individual layers the range is $1$--$15\%$.\footnote{The fraction of such pairs follows, feature by feature, from the mean persistence at $g\,{=}\,0$ and $g\,{=}\,1$ as $(\psi^{(1)}-\psi^{(0)})/(1+\psi^{(1)})$, since each bridged gap adds one token to the span and removes one episode.}
Because the bridged episodes are long, bridging them raises the mean persistence of a layer by $10$--$44\%$ depending on the layer, or $21$--$35\%$ when averaged over the layers of a model.

\paragraph{Mean and median across depth.}
Fig.~\ref{fig:SAE-layer-dependencies} completes Fig.~\ref{fig:layer-dependences-NLP}b with the mean (top row) and the median (bottom row) of the per-latent persistence, for strict episodes ($g\,{=}\,0$, left) and with single lapses bridged ($g\,{=}\,1$, right).
Each marker is the average of the two corpora and the error bar is half their difference.
Persistence grows with depth in every model, most steeply over the first half of the network, and flattens or declines slightly over the last quarter.
Bridging single lapses raises the persistence by $10$--$44\%$ depending on the layer, but does not change the order of the layers.
The mean is more sensitive than the median to the few latents with very long persistence (Fig.~\ref{fig:SAE-distributions-all-models}), and this is what produces the bump in the mean of Gemma 2 9B at about two thirds of its depth, which the median does not show.

\paragraph{Correlation with the integrated autocorrelation.}
Table~\ref{tab:docmean-correlations} asks whether removing the mean token of each sample (Appendix~\ref{app:subtracting-mean-chi}) makes the integrated autocorrelation follow the persistence more closely.
For each model and corpus we take the layers of the sweep and compute the Pearson correlation across them between the persistence of a layer, mean or median over its $1000$ latents with strict episodes, and either $\chi_\ell$ or $\tilde{\chi}_\ell$ of the same layer. In most cases except for Gemma2-9b, removing the mean token augments the correlation coefficients. 

\begin{table}[h]
    \centering
    \small
    \setlength{\tabcolsep}{5pt}
\begin{tabular}{ll|cc|cc}
\hline
 & & \multicolumn{2}{c|}{mean persistence} & \multicolumn{2}{c}{median persistence} \\
Model & corpus & $\chi$ & $\tilde{\chi}$ & $\chi$ & $\tilde{\chi}$ \\
\hline
Gemma 2 2B & The Pile & 0.90 & 0.95 & 0.87 & 0.87 \\
Gemma 2 2B & WikiText & 0.92 & 0.95 & 0.85 & 0.88 \\
Gemma 2 9B & The Pile & 0.60 & 0.54 & 0.93 & 0.78 \\
Gemma 2 9B & WikiText & 0.59 & 0.53 & 0.87 & 0.84 \\
Pythia-160M & The Pile & 0.81 & 0.96 & 0.69 & 0.88 \\
Pythia-160M & WikiText & 0.68 & 0.84 & 0.50 & 0.72 \\
Llama-3-8B & The Pile & 0.71 & 0.83 & 0.68 & 0.79 \\
Llama-3-8B & WikiText & 0.71 & 0.80 & 0.53 & 0.74 \\
Qwen3-8B & The Pile & 0.72 & 0.79 & 0.62 & 0.67 \\
Qwen3-8B & WikiText & 0.89 & 0.91 & 0.78 & 0.78 \\
\hline
\end{tabular}

    \caption{Each entry is the Pearson correlation, across the swept layers of one model and corpus, between the persistence of a layer (mean or median over its $1000$ most active latents, strict episodes) and the integrated autocorrelation of the same layer, standard ($\chi$) or with the mean token of each sample removed ($\tilde{\chi}$). Vertical lines separate the $\chi$/$\tilde{\chi}$ pairs.}
    \label{tab:docmean-correlations}
\end{table}

\section{Distributions of SAE-latent persistence for every model}
\label{app:SAE-distributions-all-models}

\begin{figure}[p]
    \centering
    \includegraphics[width=\linewidth]{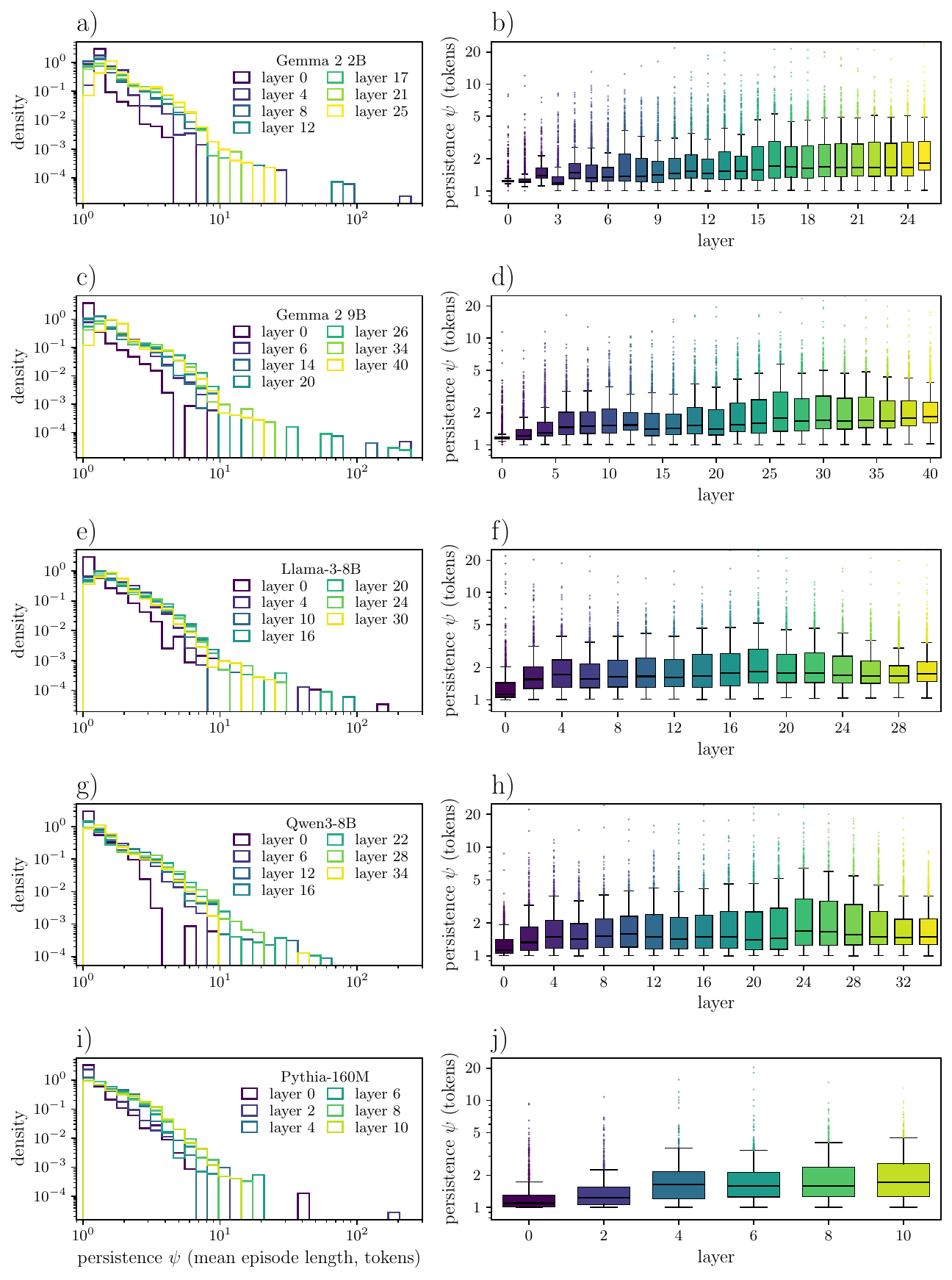}
    \caption{Distribution of the per-latent persistence $\psi$ on The Pile with single inactive tokens bridged ($g\,{=}\,1$), over the 1000 most active latents of each layer of Gemma 2 2B (a, b), Gemma 2 9B (c, d), Llama-3-8B (e, f), Qwen3-8B (g, h) and Pythia-160M (i, j). Left column: density on log-log axes at seven layers spread over the depth of each model (layers in the legend). Right column: box plots at every layer of the sweep (median line, interquartile box, whiskers at $1.5$ interquartile ranges, individual dots beyond), with a logarithmic vertical axis cut at 25 tokens. Layers are colored by relative depth on the same scale in both columns.}
    \label{fig:SAE-distributions-all-models}
\end{figure}

Each latent has one number, its persistence averaged over its firing episodes.
Fig.~\ref{fig:SAE-distributions-all-models} shows how this number is spread over the $1000$ latents of a layer, for the five models on The Pile with single lapses bridged ($g\,{=}\,1$, see Appendix~\ref{app:SAE-layer-dependencies}).
We show The Pile only.
The left column overlays the density of the persistence, on log-log axes, at seven layers spread over the depth of the model.
The right column shows a box plot for every layer of the sweep, all $26$ layers for Gemma 2 2B, with the same colors.
At the first layer almost every latent fires for a single token, and the distribution is concentrated just above $\psi\,{=}\,1$.
With depth, the bulk of the distribution moves to larger $\psi$ and the tail extends to tens of tokens, so both the median and the spread across latents grow.
Most of the shift happens over the first half of the network; after that the boxes of consecutive layers overlap.
The vertical axis of the box plots is cut at $25$ tokens, and fewer than $1\%$ of the latents of any layer lie above the cut.
The latents beyond the whiskers, drawn as dots, are present at every depth beyond the first few layers.
They pull up the mean persistence of a layer but not its median.

\section{SAE explained variance}
\label{sec:SAE-explained-variance}

Here we report how much of each layer's activation the SAE reconstructs.
For every layer $\ell$ we compute
\begin{equation}
    \mathrm{EV}_\ell = 1 - \frac{\sum_t \| \phi_t - \hat{\phi}_t \|^2}{\sum_t \| \phi_t \|^2},
    \label{eq:sae-explained-variance}
\end{equation}
where $\phi_t$ is the raw activation at token position $t$, $\hat{\phi}_t = \sum_i z_i(t)\, W_{\rm dec}(i) + b_{\rm dec}$ its SAE reconstruction from the codes $z_i(t)$, and the sums run over all positions of $300$ samples of each corpus, cut to a common length.
We call $\mathrm{EV}_\ell$ the explained variance, as is customary, but the activations in Eq.~\ref{eq:sae-explained-variance} are not centered, so it is the fraction of the squared norm of the activations that the reconstruction recovers.

\paragraph{Massive activations.}
A few tokens have activations whose norm is ten to a hundred times larger than any other token's~\citep{sun2024massive}.
The first token of every sample is one of them.
In Qwen3-8B, from the middle of the network on, there is sometimes a second such token per sample; on The Pile it is usually a digit token, on WikiText the first content token after a short opening one.
The SAEs used here do not reconstruct such activations well, and since Eq.~\ref{eq:sae-explained-variance} sums squared errors, fewer than one token in a thousand would otherwise dominate it; at layers $6$--$15$ of Qwen3-8B these tokens hold essentially all of the reconstruction error.
We therefore exclude, at every layer and for every model, the tokens whose squared norm exceeds $20$ times the median squared norm of the layer, after discarding the beginning-of-sequence position for the models whose tokenizer prepends one.
The two groups of tokens are separated by a wide margin, so the exact value of the threshold does not matter.
The rule removes at most $0.4\%$ of the tokens for four of the five models and at most $0.9\%$ for Pythia-160M.
It changes $\mathrm{EV}_\ell$ by less than $0.03$ for every model but Qwen3-8B, where the excluded tokens held most of the error.

\begin{figure}[h]
    \centering
    \includegraphics[width=0.5\linewidth]{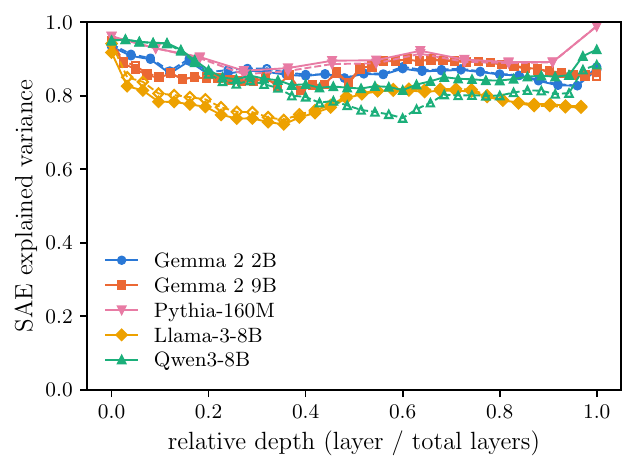}
    \caption{SAE explained variance, Eq.~\ref{eq:sae-explained-variance}, as a function of relative depth for the five models, on The Pile (filled markers, solid lines) and WikiText (hollow markers, dashed lines), 300 samples per corpus, massive-activation tokens excluded as described in the text.}
    \label{fig:sae-explained-variance}
\end{figure}

Fig.~\ref{fig:sae-explained-variance} shows $\mathrm{EV}_\ell$ across depth.
The two corpora give nearly the same values.
For every model, $\mathrm{EV}_\ell$ stays above $0.7$ at every layer, and above $0.8$ at most of them; the lowest values are those of the Llama-3-8B SAEs, and of the Qwen3-8B SAEs in the second half of the network.
The reconstruction is therefore not perfect, but it is good enough for the SAE latents to stand in for the activations in all our experiments.
The part of the activations that the SAEs miss is also one reason why Table~\ref{tab:causal-ablation} does not reach zero even when every active latent is removed.

\section{Autocorrelations without clipping the activations}
\label{app:clip-control}

\begin{figure}[h]
    \centering
    \includegraphics[width=\linewidth]{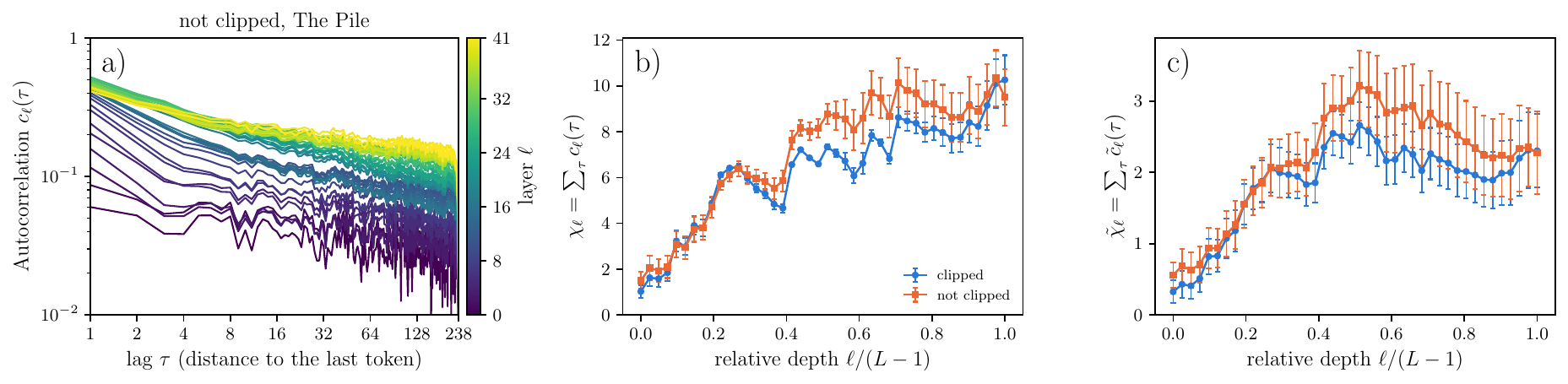}
    \caption{Control without clipping, Gemma 2 9B. (a) Autocorrelation $c_\ell(\tau)$ of every layer on The Pile computed on the raw activations, drawn as Fig.~\ref{fig:main}b. (b) Integrated correlation $\chi_\ell$ and (c) its version with the mean token of each sample removed, $\tilde{\chi}_\ell$, against relative depth, on clipped (circles) and raw (squares) activations. Markers are the average of The Pile and WikiText, error bars half the difference between the two corpora.}
    \label{fig:clip-control}
\end{figure}

The autocorrelations of Figs.~\ref{fig:main}b and \ref{fig:autocorrelations-other-models}, and every $\chi_\ell$ and $\tilde{\chi}_\ell$ built from them, are computed on clipped activations (Appendix~\ref{app:autocorrelations-other-models}).
To check that the clipping does not shape the results, we repeat the measurement for Gemma 2 9B on both corpora with the raw activations, everything else unchanged.
Fig.~\ref{fig:clip-control} shows the outcome.
The autocorrelation curves without clipping (a) are the same as in Fig.~\ref{fig:main}b: the layers order by depth in the same way and decay at the same rate.
The integrated correlation is slightly larger without clipping in the second half of the network (b), where the raw activations have a heavier tail, but its growth with depth and the position of its maximum are unchanged.
With the mean token of each sample removed (c), both versions grow over the first half of the network and midly decrease in the second half of the network.
The conclusions of the main text and of Appendix~\ref{app:subtracting-mean-chi} therefore do not depend on the clipping.

\section{Norm removed by the SAE ablations}
\label{app:ablation-removed-norm}

\begin{figure}[h]
    \centering
    \includegraphics[width=0.5\linewidth]{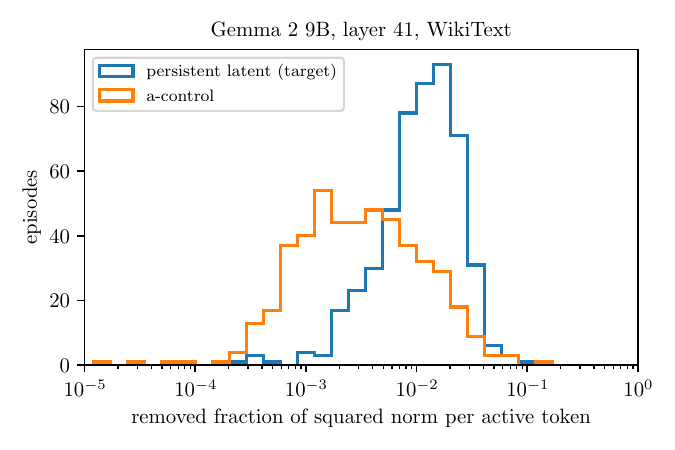}
    \caption{Removed fraction of squared norm, Eq.~\ref{eq:removed-norm-fraction}, averaged over the ablated positions at which the latent is active, for the persistent latent and the a-control of Table~\ref{tab:causal-ablation}. Gemma 2 9B, layer 41, WikiText, 500 episodes; activations centred by the layer mean, as in the ablation experiment.}
    \label{fig:removed-norm-fraction}
\end{figure}

The ablation of Eq.~\ref{eq:ablation-SAE} removes, at position $t$, the fraction
\begin{equation}
r_t(i) = \frac{(\phi_t, W_{\rm dec}(i))^2}{\|\phi_t\|^2\,\|W_{\rm dec}(i)\|^2}
\label{eq:removed-norm-fraction}
\end{equation}
of the squared norm of the activation.
For every episode of Table~\ref{tab:causal-ablation} we average $r_t$ over the ablated positions $t \in [s, e-1]$ at which latent $i$ is active, for the persistent latent (all of them, by construction) and for the a-control.
Fig.~\ref{fig:removed-norm-fraction} shows the two distributions for Gemma 2 9B on WikiText.
Where both latents are active, the a-control removes a median $0.29\%$ of the squared norm against $1.2\%$ for the persistent latent, roughly a factor 4, and the persistent latent removes more in $80\%$ of the episodes.

Temporal persistence and projection magnitude are, in principle, independent properties: a feature may remain active across many tokens while accounting for only a small fraction of the activation norm, or have a large projection over a short span. Interestingly, our results show that these two properties are coupled in our experimental setting, with persistent features also capturing a larger fraction of the activation norm when active.

\end{document}